\pdfoutput=1

\documentclass{article}

\usepackage[nonatbib,preprint]{neurips_2020}
\usepackage{url}
\usepackage{booktabs}
\usepackage{amsfonts}
\usepackage{nicefrac}
\usepackage{microtype}
\usepackage{xspace}
\usepackage[pdftex]{graphicx}
\usepackage{subfigure}
\usepackage[pdftex]{color}
\usepackage{cite}
\usepackage{amsmath}

\newcommand{\method}{CCAC\xspace}
\newcommand{\basicModel}{CCAC\xspace}
\newcommand{\advanceModel}{CCAC-S\xspace}
\newcommand{\transferMethod}{CCAC-T\xspace}

\newcommand{\ouralg}{CCAC\xspace}
\newcommand{\ouralgtwo}{CCAC-S\xspace}
\newcommand{\ouralgthree}{CCAC-T\xspace}

\usepackage{ifthen}
\newboolean{showcomments}
\setboolean{showcomments}{true}
\newcommand{\note}[1]{\ifthenelse{\boolean{showcomments}}
{\textcolor{red}{#1}}{}}
\newcommand{\zhihui}[1]{{\color{black}{#1}}}

\title{Calibrating Deep Neural Network Classifiers on Out-of-Distribution Datasets}

\author{
 Zhihui Shao\thanks{E-mail: zshao006@ucr.edu}\\
  UC Riverside\\
   \And
  Jianyi Yang\thanks{E-mail: jyang239@ucr.edu}\\
  UC Riverside \\
   \And
  Shaolei Ren\thanks{E-mail: sren@ece.ucr.edu}\\
 UC Riverside\\
}

\begin{document}

\maketitle

\begin{abstract}
To increase the trustworthiness of deep neural network (DNN) classifiers,
an accurate prediction {confidence} that represents the true likelihood of correctness is crucial. Towards this end, many post-hoc calibration methods have been proposed to leverage a lightweight model to map the target DNN's output layer into a calibrated confidence. Nonetheless,
on an out-of-distribution (OOD) dataset in practice, the target DNN can often mis-classify samples with a high confidence, creating significant
challenges for the existing calibration methods to produce an accurate confidence. In this paper, we propose a new post-hoc confidence calibration method, called \method (Confidence Calibration with an Auxiliary Class), for DNN classifiers on OOD datasets. The key novelty of \ouralg is  an auxiliary class in the calibration model which separates mis-classified samples from correctly classified ones, thus effectively mitigating the target DNN's being confidently wrong. We also propose a simplified version of \ouralg to reduce free parameters and facilitate transfer to a new unseen dataset. Our experiments on different DNN models, datasets and applications show that \ouralg can consistently outperform
the prior post-hoc calibration methods.

\end{abstract}

\section{Introduction}

In recent years, classifiers based
on deep neural networks (DNNs) have been increasingly
applied to a wide variety of applications, including
safety-critical applications \cite{imagenet_lifeifei_2009_CVPR,dnn_speech_recognition_han_2014,dnn_nlp_deng_2018} such as autonomous/assisted driving \cite{dnn_autonomous_driving_hayder_IV_2017} and medical imaging \cite{dnn_medical_diagnose_MRI_sander_2019,dnn_healthcare_caruana_2015_KDD}.
For trustworthiness, a quantitative \emph{confidence}
level should also be provided  along with predicted labels
of DNN classifiers, representing the actual likelihood of correctness
of  the corresponding classification result during inference \cite{calibration_guo_2017_ICML,dirichlet_calibration_kull_2019_NIPS,scaling_binning_calibration_kumar_2019_NIPS,DNN_Uncertainty_Estimation_Summary_NIPS_2019_ovadia,DNN_Uncertainty_PriorNetworks_NIPS_2018_malinin2018predictive}.
Taking assisted driving as an example, a car may slow down and request human intervention in the event of a low-confidence prediction, ensuring a higher level of trustworthiness and safety.

Typically, for a DNN classifier with $K$ classes, the last layer contains $K$ neurons that correspond
to a $K$-dimensional array of prediction probabilities (or a $K$-dimensional logit array, which can be converted into probabilities via softmax function),
with the predicted class being the one that has the maximum prediction probability
 \cite{DNN_Book_Goodfellow-et-al-2016}. Straightforwardly, a prediction confidence can be directly obtained as the softmax probability for the predicted class \cite{DNN_Uncertainty_Baseline_OOD_ICLR_2017}. Nonetheless, oftentimes, this naive un-calibrated confidence may not reveal the true prediction confidence, especially for modern complex DNN models. The mismatch between
 the softmax probability and the true prediction confidence can be attributed
  to  over-fitting, batch normalization layers, among others \cite{calibration_guo_2017_ICML}.

Even worse, the actual target test dataset on which a  DNN
is applied in practice  is often \emph{out-of-distribution} (OOD) compared to
the original dataset used to train the target DNN
\cite{DNN_Uncertainty_Estimation_Summary_NIPS_2019_ovadia,explain_adversarial_goodfellow_2014,DNN_Uncertainty_Baseline_OOD_ICLR_2017}.
For example, in image classification, the actual images may be rotated and shifted, come from
completely different domains with unseen labels, and/or even be
modified by adversaries,
 which we broadly refer to as \emph{OOD} samples in this paper.
 Importantly, recent studies have shown that the commonly-encountered OOD samples can
cause a considerably large accuracy degradation and lead
to over-confident predictions, thus
presenting a significant challenge to the confidence of DNN prediction results
and raising trust concerns \cite{DNN_Uncertainty_Estimation_Summary_NIPS_2019_ovadia}.
While proactively detecting OOD samples to single them out may
help prevent accuracy degradation
\cite{Verification_OOD_Detection_ODIN_Srikant_UIUC_ICLR_2018,DNN_Uncertainty_PriorNetworks_NIPS_2018_malinin2018predictive,early_output_OOD_detection_2019,deep_verifier_OOD_2019}, not all OOD samples
are mis-classified and hence a quantitative prediction confidence is still desired \cite{DNN_Uncertainty_Estimation_Summary_NIPS_2019_ovadia}.

To obtain a prediction confidence that well represents the true likelihood
of correctness, the target DNN model can be re-trained with a modified structure or training algorithm, such as Monte-Carlo dropout, SWAG, and deep ensembles \cite{deep_ensembles_charles_2017_NIPS,MCdropout_bayesian_gal_2016_ICML,swa_average_wight_gordon_2018},
which are substitutes of otherwise computationally expensive
Bayesian DNNs \cite{DNN_Calibration_BayesianDNN_NIPS_2017_7141}.
Nonetheless, re-trained DNNs may not guarantee an accurate prediction confidence
on an actual test dataset with OOD samples, and it can be expensive
or even impossible to re-train them each time the test data distribution changes.
Alternatively,  post-hoc confidence calibration
\cite{platter_scaling_1999,calibration_guo_2017_ICML,dirichlet_calibration_kull_2019_NIPS}
has been widely
studied, which learns a lightweight model/network to
map the target
DNN's output layer into a calibrated confidence score without re-training the DNN
or modifying its architecture.
Nonetheless, the calibrated
confidence can be far off from the true likelihood of correct prediction for OOD samples in practical test datasets
when the target DNN is often confidently wrong
\cite{DNN_Uncertainty_Estimation_Summary_NIPS_2019_ovadia}.

\textbf{Our contribution.} We study post-hoc confidence calibration 
for a  DNN classifier when the actual test dataset can
be possibly OOD (compared to the dataset used to train the target DNN).
 We propose a new post-hoc confidence calibration method, called \method (Confidence Calibration with an Auxiliary Class), which
building on top of a lightweight (e.g., 2-layer) neural network,
 only needs the target DNN's logit layer as input and maps it to a new
calibrated softmax probability which can be used to indicate the prediction confidence.
The key novelty of \ouralg is the introduction of an auxiliary class which represents
mis-classified samples  and separates them from correctly classified
ones in the actual test dataset, thus
effectively mitigating the target DNN's being confidently wrong (i.e., assigning a high confidence
for mis-classified samples). 
To reduce the number of free parameters in \ouralg, we propose
a further simplified
calibration model, called \ouralgtwo (\ouralg-Simplified) as illustrated in Fig.~\ref{fig:calib2_model}, which
uses a single temperature parameter to map the target DNN's logit for
correct classes
in addition to leveraging a small neural network for the mis-classification class.
More importantly, 
given a new unseen OOD test dataset in practice,
we can transfer \ouralg to the test dataset by using only
a small number of newly labeled
samples.

To evaluate the calibration performance, we conduct experiments for different DNNs on OOD datasets with both image and document classification applications (e.g., VGG16
on CIFAF-100 for image classification). The results show that our approach can consistently outperform
recent post-hoc calibration methods in terms of
the widely-used metrics of expected calibration error and Brier score,
supporting the introduction of an auxiliary class to represent
mis-classified samples to achieve a better confidence on OOD datasets.

\section{Related Work}\label{sec:related_work}

\textbf{Confidence estimation.} Most conventional DNN
models do not provide
accurate prediction uncertaintities/confidences
\cite{DNN_Uncertainty_Estimation_Summary_NIPS_2019_ovadia}.
While Bayesian DNN is an effective approach for uncertainty estimation,
it requires significant modifications
to the training procedure and suffers from a high computational
cost for inference \cite{DNN_Calibration_BayesianDNN_NIPS_2017_7141}.
Consequently, novel DNN models and training algorithms, such
as
SWAG model \cite{swa_average_wight_gordon_2018}, deep ensemble models \cite{deep_ensembles_charles_2017_NIPS}  and Monte-Carlo dropout 
\cite{MCdropout_bayesian_gal_2016_ICML}, have also been proposed to re-train
DNNs as an approximate Bayesian approach.  Nonetheless, these methods require a large training dataset and apply well each time the test
data distribution changes. 
 They are complementary to and can be applied
in combination with post-hoc calibration methods (e.g.,
applying temperature scaling for each of the DNN models in an ensemble).
Additionally, without
model re-training, \cite{model_uncertain_suchi_ICAIS_2019} proposes to re-sample weights in a DNN model as a Bayesian estimation,
and \cite{DNN_Calibration_Classifier_Uncertainty_google_nips_2018}
estimates prediction uncertainty using a trust score based on a ratio of the distance
between a sample and its closest class to the distance
between this sample and its predicted class.
These methods also require the original training dataset
and hence can only work well for in-distribution datasets.

\textbf{Post-hoc confidence calibration.} Without modifying an already-trained
target DNN model, post-hoc confidence calibration can provide
an accurate estimate of the prediction uncertainty by learning
a mapping function from the model's logit or probability output to a
new probability that better represents the actual confidence. Typically, post-hoc calibration models,  such as platter scaling \cite{platter_scaling_1999}, temperature scaling, vector scaling and matrix scaling \cite{calibration_guo_2017_ICML},
require a set of parameters that can be learnt via minimizing a certain loss
function (e.g., negative log-likelihood or NLL) over a training/validation dataset.
More recently, \cite{dirichlet_calibration_kull_2019_NIPS} proposes a sophisticated scaling method with Dirichlet functions and NLL loss for post-hoc calibration.  Non-parametric calibration methods also exist, including binning methods \cite{bining_calibration_charles_ICML_2001} and isotonic regression method \zhihui{\cite{isotonic_regression_charles_2012_KDD}}. Moreover, some mixture methods are also proposed, including Bayesian binning \cite{bayesian_binning_milos_AAAI_2015} and scaling binning \cite{scaling_binning_calibration_kumar_2019_NIPS}. However, the existing
post-hoc calibration methods are mainly developed for in-distribution datasets
and, as empirically shown in
  a recent study \cite{DNN_Uncertainty_Estimation_Summary_NIPS_2019_ovadia},
cannot calibrate the prediction confidence well for OOD datasets. Our proposed calibration method falls into post-hoc calibration
but can provide a good calibration performance on an actual test dataset that typically includes
OOD samples.

\textbf{OOD/adversarial/misclassification detection.}
A simple baseline method using softmax probability (plus
some hidden layers as an extension) for misclassification/OOD
detection is considered in \cite{DNN_Uncertainty_Baseline_OOD_ICLR_2017}.
Following up, \cite{Verification_OOD_Detection_ODIN_Srikant_UIUC_ICLR_2018} detects OOD samples using temperature scaling and input perturbations,  \cite{Verification_SimpleFramework_OOD_Generative_Gaussian_NIPS_2018_7947}
studies OOD detection by learning a generative model based on hidden layers
of a white-box DNN,
\cite{early_output_OOD_detection_2019} extracts features from inner layers for OOD detection in white-box DNNs,  \cite{deep_verifier_OOD_2019}
trains a deep verifier (generative neural network) based on raw input data for OOD and misclassification detection,
 \cite{DNN_Uncertainty_PriorNetworks_NIPS_2018_malinin2018predictive} proposes
a prior network to better separate OOD/adversarial samples from in-distribution/benign
samples, and \cite{Verification_OOD_SelfSupervised_SeparateClass_TAMU_ZhengyangWang_AAAI_2020}
uses self-supervised learning to jointly detect OOD and train the target DNN model.
Another related line of research is learning with rejection \cite{Learning_Boosting_with_Abstention_NIPS_2016_LearningNIPS2016_6336,Learning_with_Rejection_ALT_2016_10.1007/978-3-319-46379-7_5,DNN_SelectiveClassification_NIPS_2017_7073}, where
the classifier is trained along with a rejection score function such
that the classifier can selectively provide predictions based
on the corresponding rejection score.
While these methods can single out OOD/adversarial samples (and sometimes
mis-classified samples), they cannot provide a quantitative prediction confidence on OOD/adversarial
samples, which may still be correctly classified by the target DNN.

\section{Confidence Calibration With an Auxiliary Class}

\subsection{Problem Formulation}

We consider a DNN classifier $\mathcal{F}_\Theta(\mathbf{x})$ with $K$ classes, where $\mathbf{x}$ is the input data with a true label $y$ and $\Theta$
represents the model parameters learned from labeled training samples 
in dataset $\mathcal{D}^T$. The model parameters $\Theta$ are not required nor
modified by post-hoc confidence calibration methods \cite{calibration_guo_2017_ICML}.

The actual test data $(\mathbf{x}, y)$ is drawn from a 
dataset $\mathcal{D}$, whose data
distribution may differ from that of the dataset $\mathcal{D}^T$ used
for training the target DNN.
 Given an input $\mathbf{x}$, the target DNN's logit output is denoted by $\mathbf{z}\in\mathbb{R}^K$, where we suppress the dependence of
$\mathbf{z}(\mathbf{x})$
 on $\mathbf{x}$
for notational convenience.
The corresponding output probability is $\mathbf{p}(\mathbf{z})=\sigma_{SM}(\mathbf{z})\in\mathbb{R}^K$, where $\sigma_{SM}$ is the softmax function.
 The predicted label $\hat{y}$ is decided
 as  $\hat{y}=\arg\max_{k\in\{1,2...K\}}\{{p}_k(\mathbf{z})\}$.
The classification is correct when $\hat{y}=y$ and wrong otherwise,
with $\mathrm{Pr}(\hat{y}=y)=\max_{k\in\{1,2...K\}}\{{p}_k(\mathbf{z})\}$
 being the (un-calibrated) classification confidence. Ideally,
 the confidence $\mathrm{Pr}(\hat{y}=y)$ should reflect the true probability
 of correct classification \cite{calibration_guo_2017_ICML}. For example, given $R$ samples
 samples each with a prediction confidence of $q$, the correctly
 predicated samples should be $R\cdot q$. Nonetheless, the un-calibrated prediction confidence of a DNN classifier
can differ significantly from the true confidence, especially
on OOD datasets \cite{DNN_Uncertainty_Estimation_Summary_NIPS_2019_ovadia}.

The goal of post-hoc confidence calibration is to map
the logit  $\mathbf{z}$ (or the softmax probability $\mathbf{p}(\mathbf{z})=\sigma_{SM}(\mathbf{z})$) of the target DNN
into a calibrated confidence that better represents the true confidence.
\subsection{Confidence Calibration}

\begin{figure*}[!t]
	\centering
	\subfigure[]{\includegraphics[trim=0cm 0cm 0cm 0cm,clip,  width=0.195\textwidth]{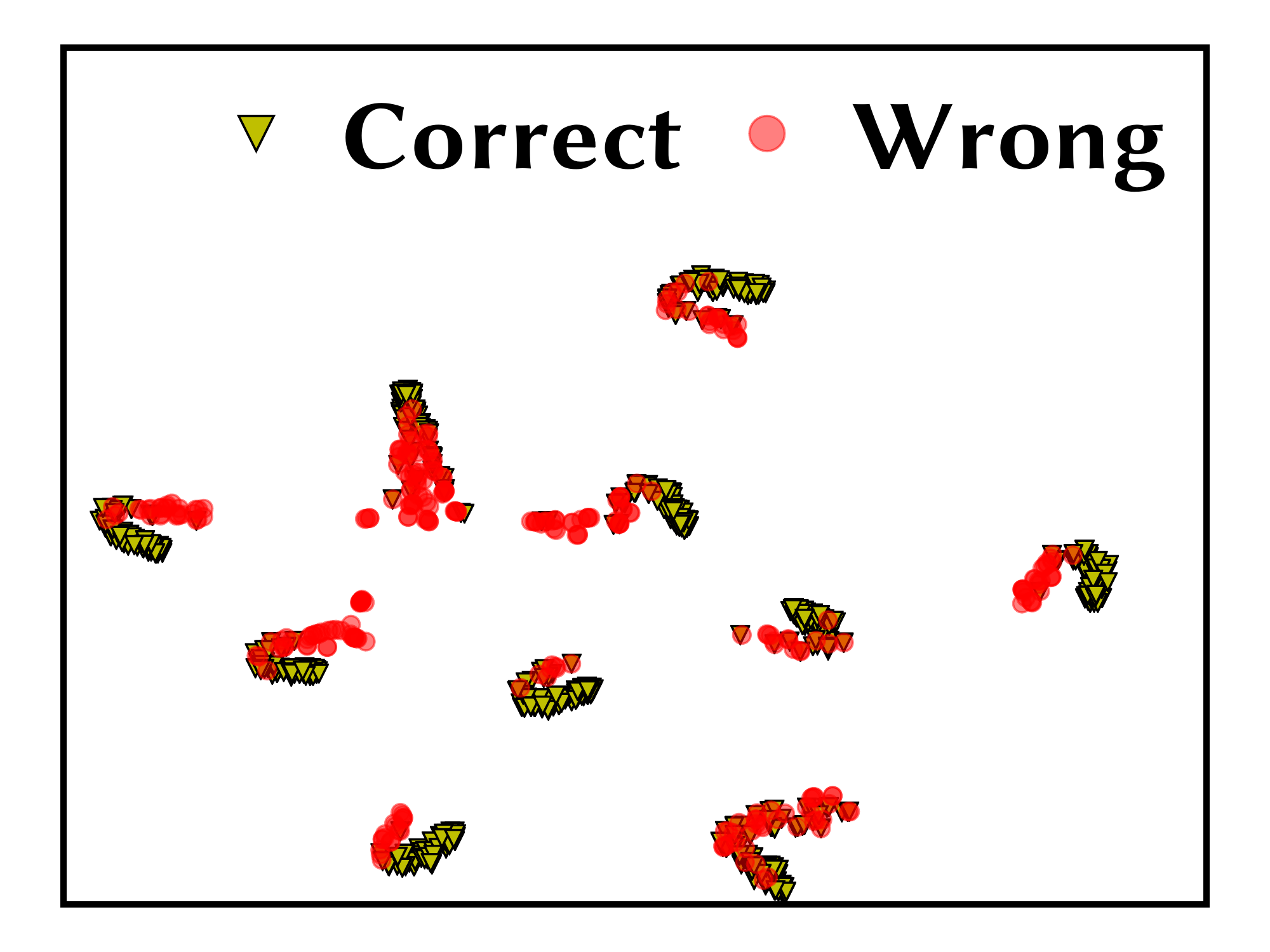}\label{fig:cifar10_tsne_OOD_MP}}
		\subfigure[]{\includegraphics[trim=0cm 0cm 0cm 0cm,clip,  width=0.195\textwidth]{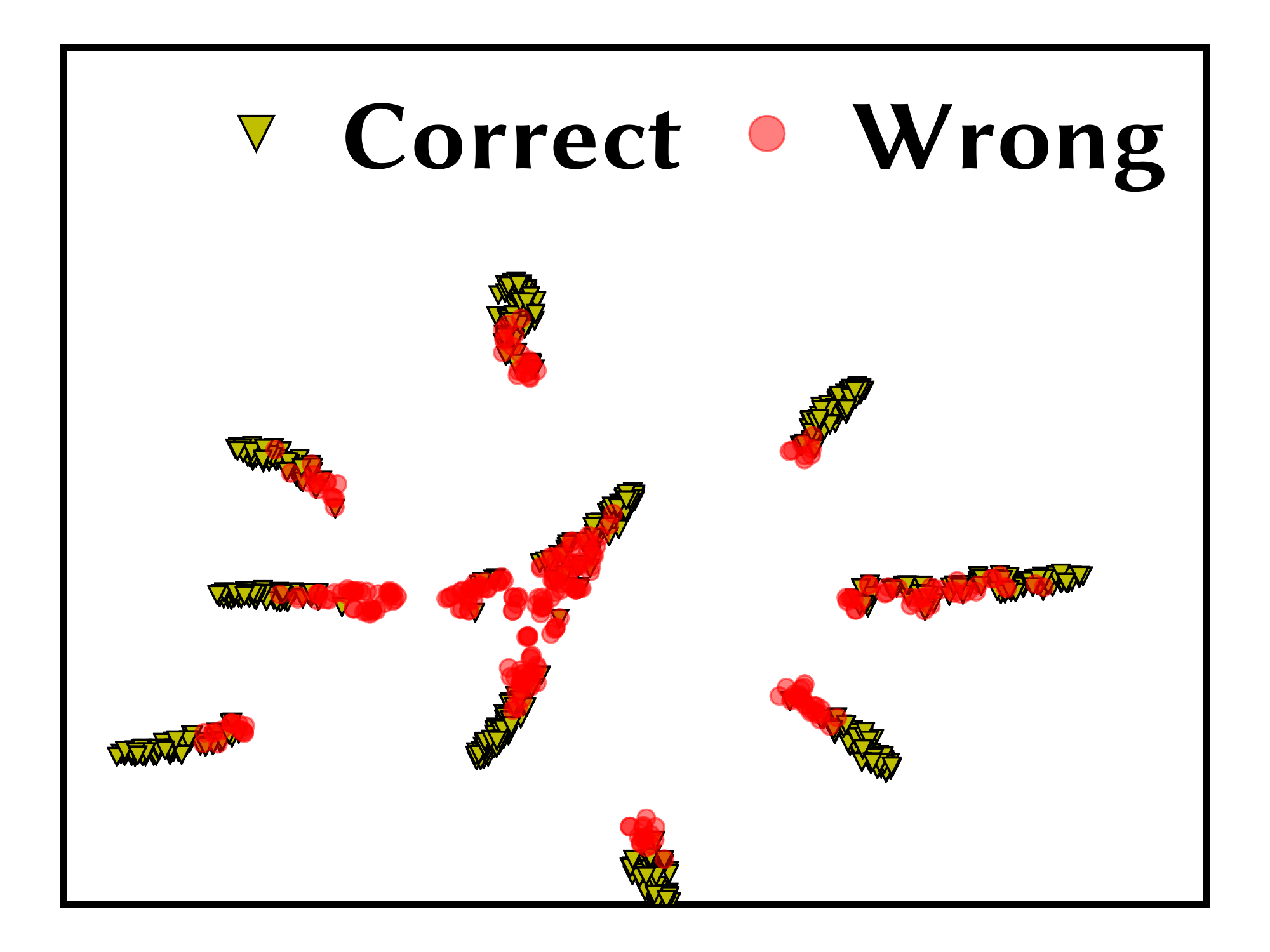}\label{fig:cifar10_tsne_OOD_TS}}	
	\subfigure[]{\includegraphics[trim=0cm 0cm 0cm 0cm,clip,  width=0.195\textwidth]{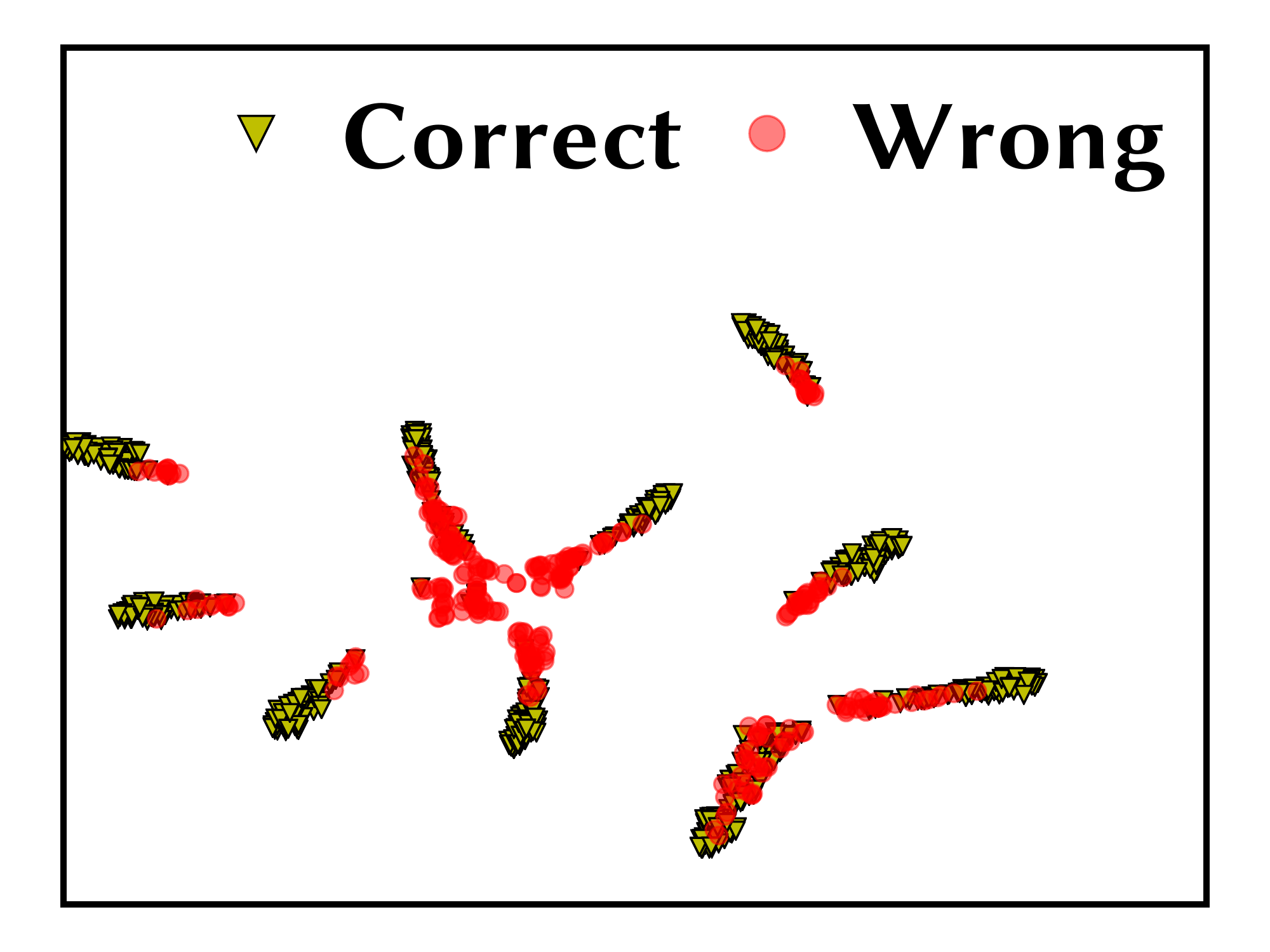}\label{fig:cifar10_tsne_OOD_SB}}
	\subfigure[]{\includegraphics[trim=0cm 0cm 0cm 0cm,clip,  width=0.195\textwidth]{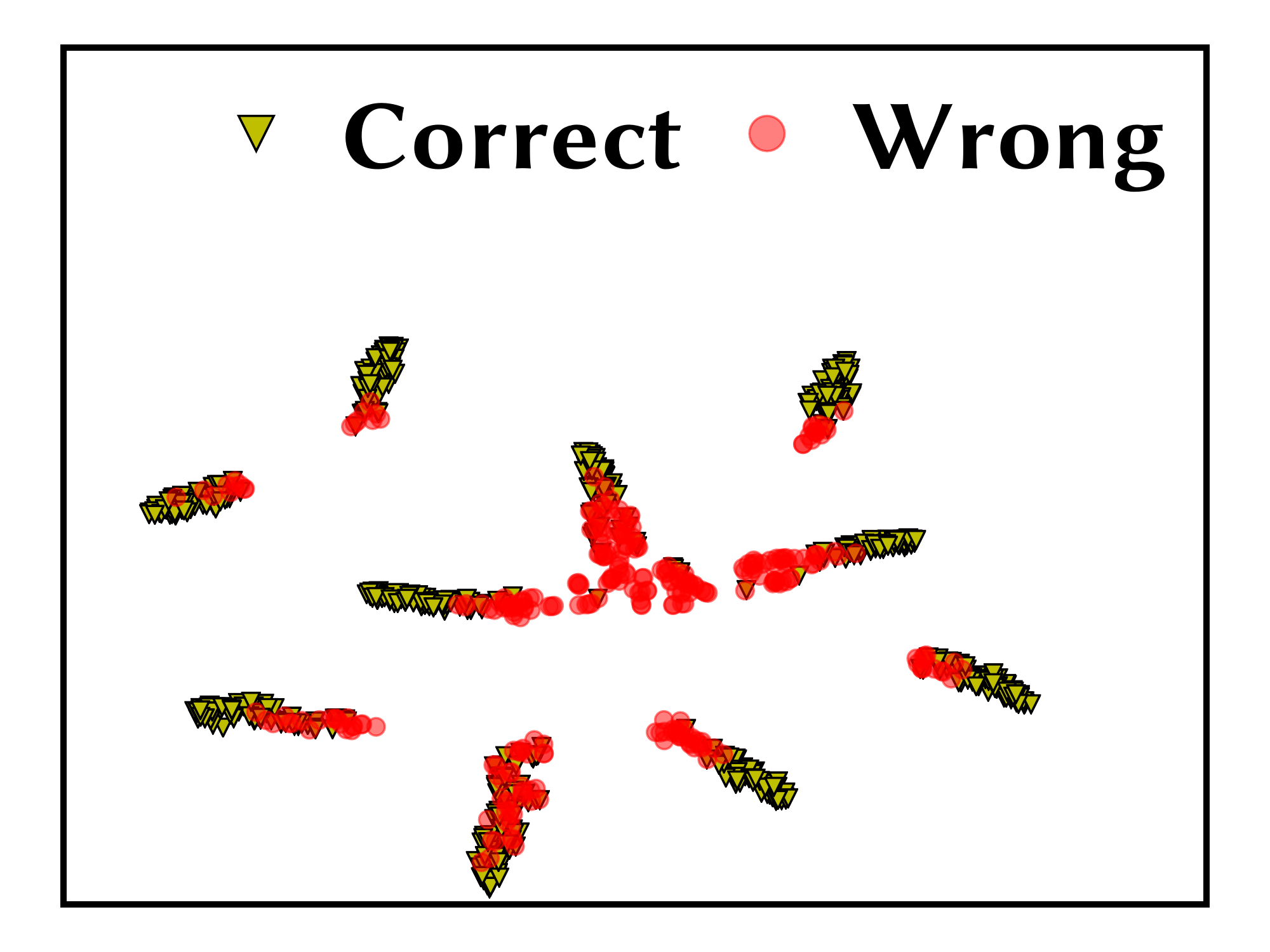}\label{fig:cifar10_tsne_OOD_Dirichlet}}
	\subfigure[]{\includegraphics[trim=0cm 5.9cm 0cm 5.9cm,clip,  width=0.195\textwidth]{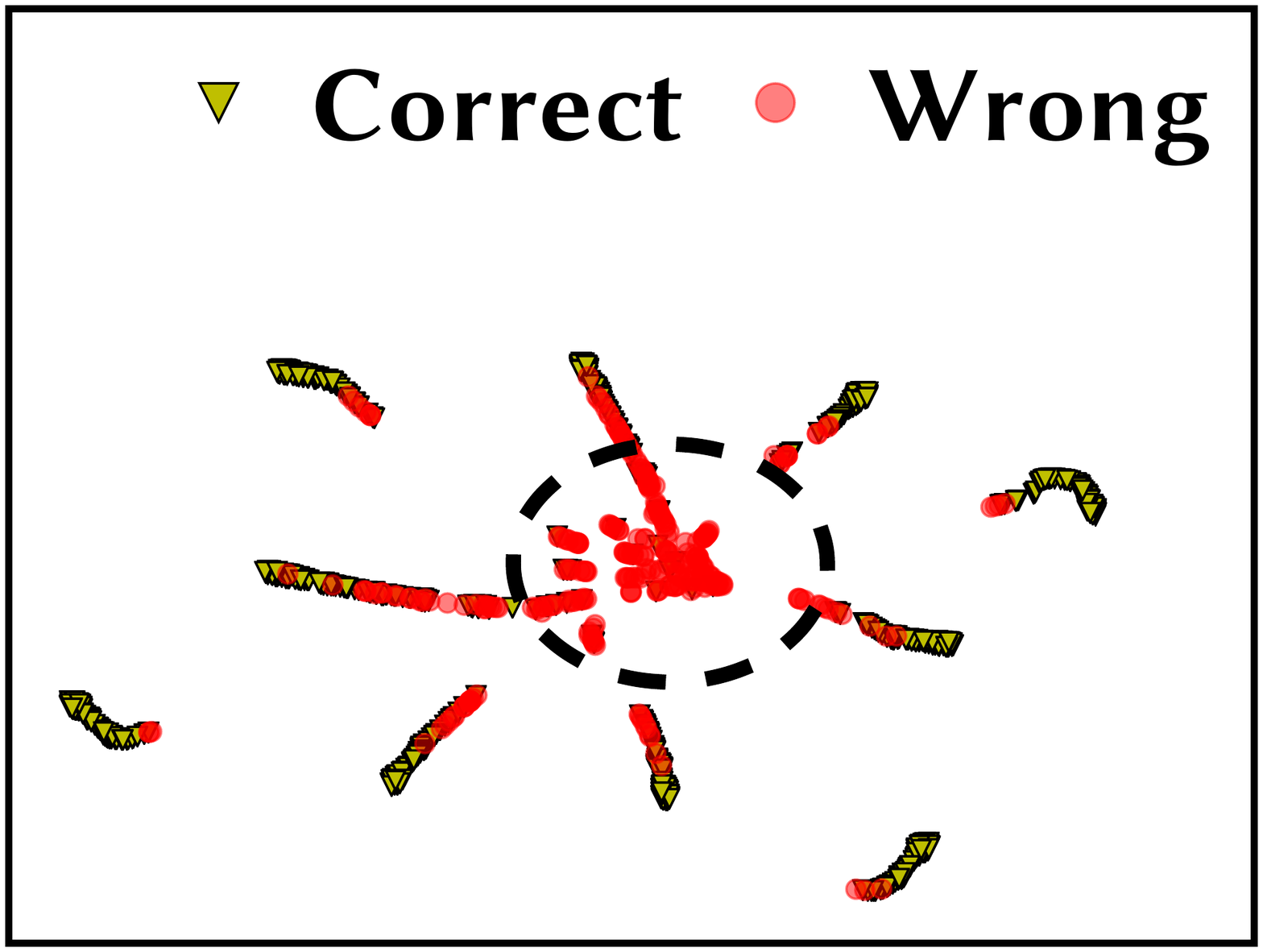}\label{fig:cifar10_tsne_OOD_Calib1_print}}
	\vspace{-0.3cm}
	\caption{t-SNE of calibrated confidence
 for {VGG16} on CIFAR-10 OOD dataset. (a) Un-calibrated. (b) Temperature scaling. (c) Scaling-binning. (d) Dirichlet calibration. (e) \ouralg.} \label{fig:cifar10_OOD_tsne}
	\vspace{-0.3cm}
\end{figure*}

\begin{figure*}[!t]
	\centering
	\subfigure[]{\includegraphics[trim=0cm 0cm 0cm 0cm,clip,  width=0.19\textwidth]{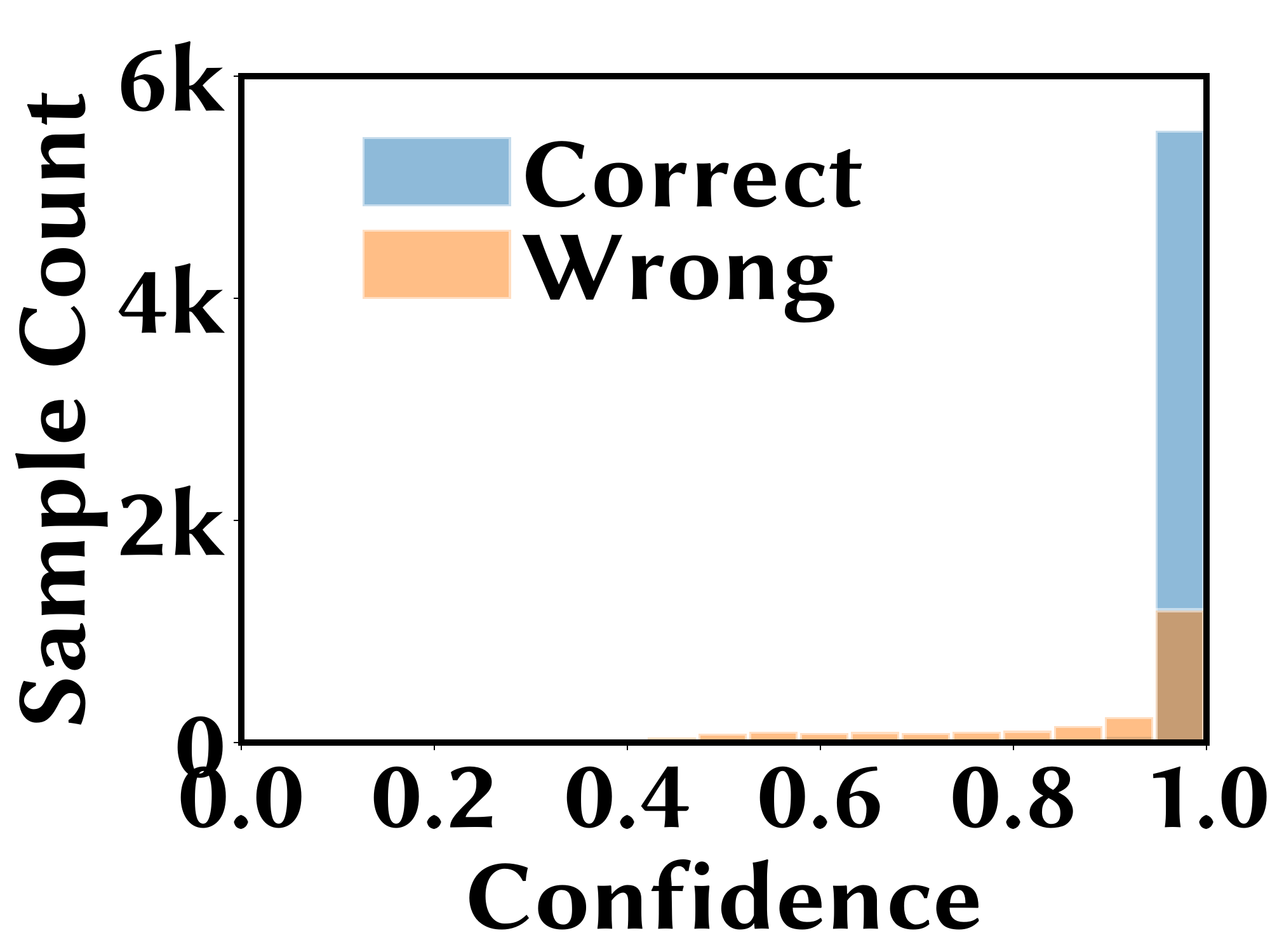}}
	\subfigure[]{\includegraphics[trim=0cm 0cm 0cm 0cm,clip,  width=0.19\textwidth]{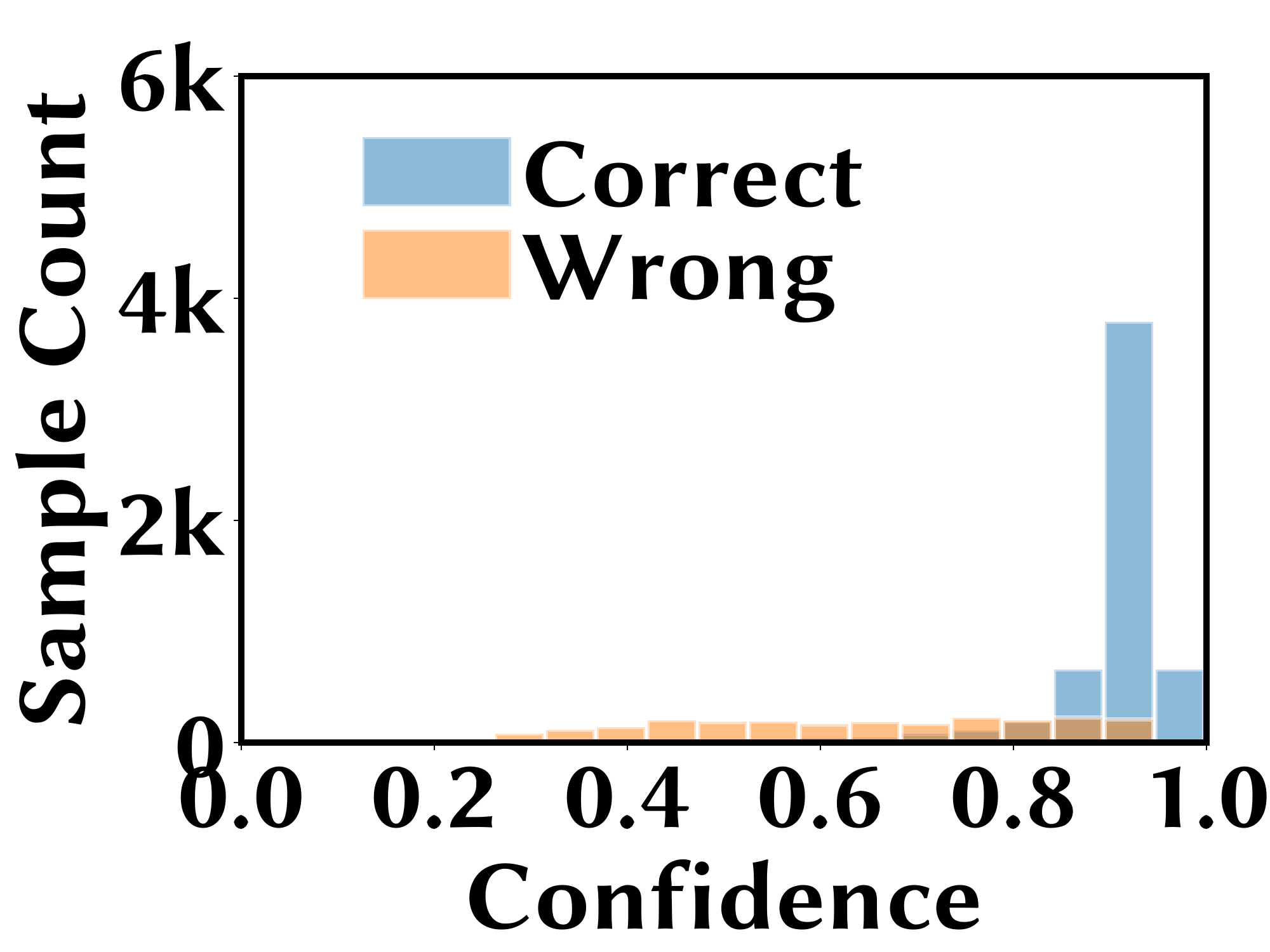}}	
	\subfigure[]{\includegraphics[trim=0cm 0cm 0cm 0cm,clip,  width=0.19\textwidth]{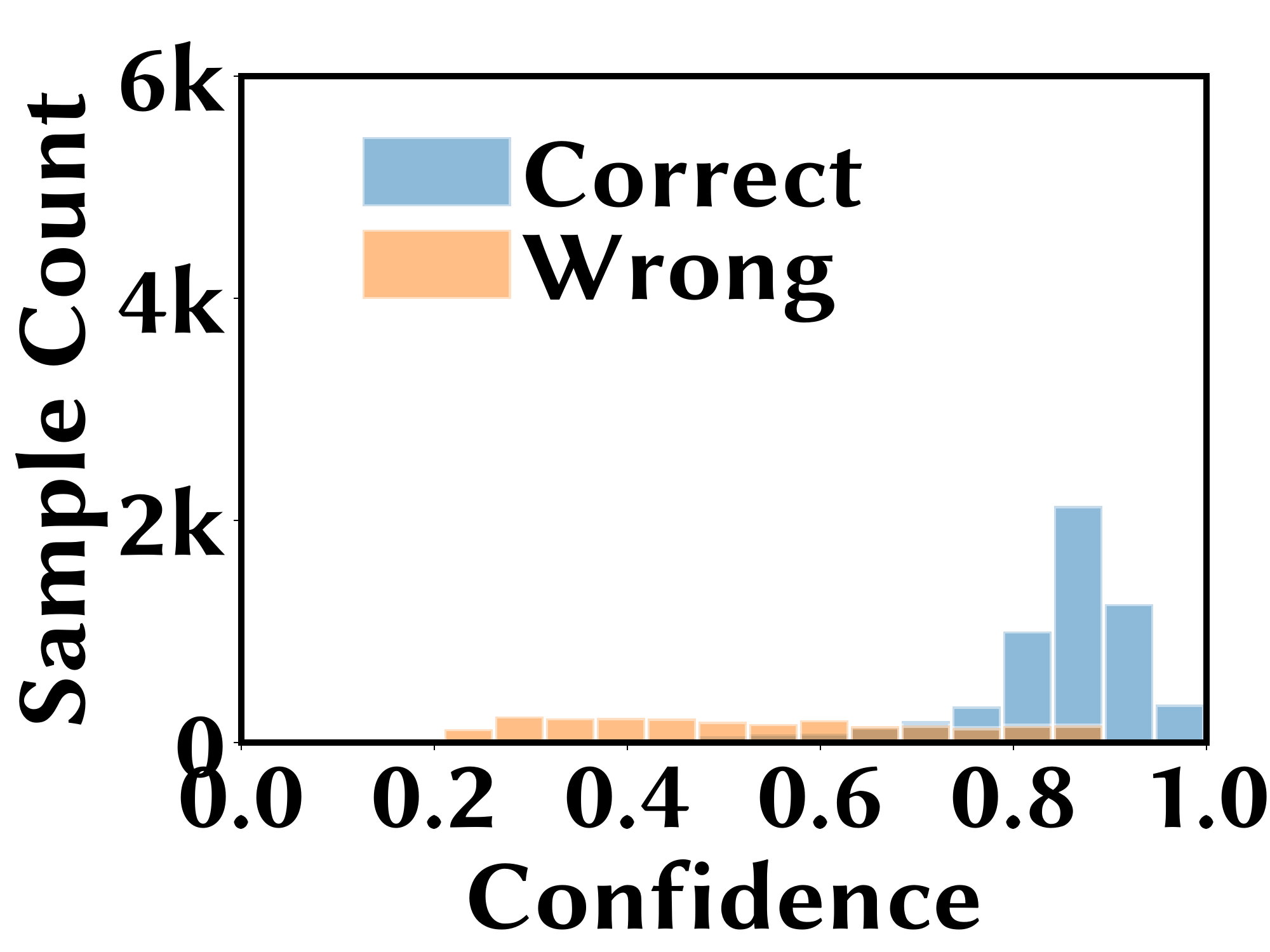}}
	\subfigure[]{\includegraphics[trim=0cm 0cm 0cm 0cm,clip,  width=0.19\textwidth]{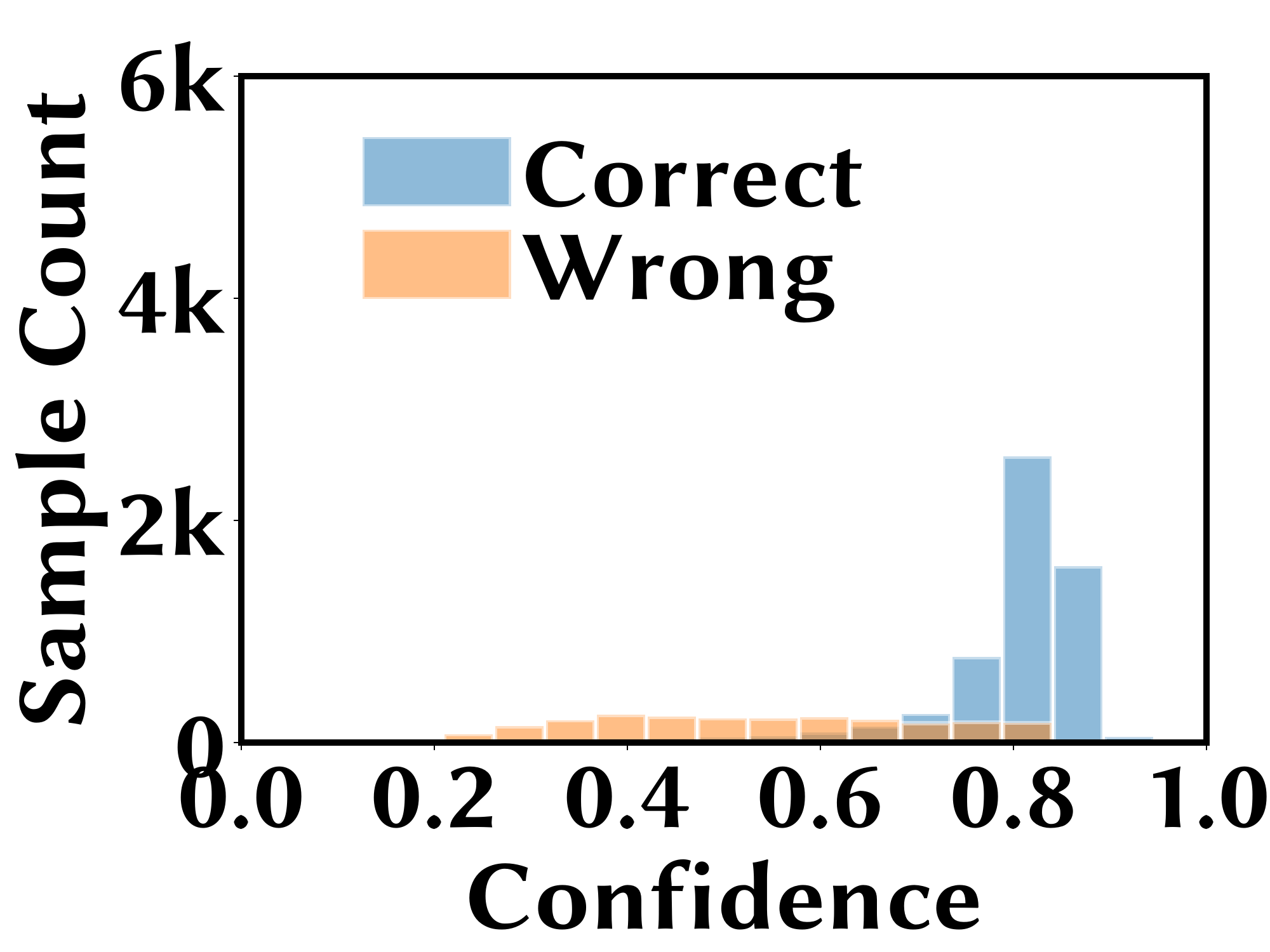}}
	\subfigure[]{\includegraphics[trim=0cm 0cm 0cm 0cm,clip,  width=0.19\textwidth]{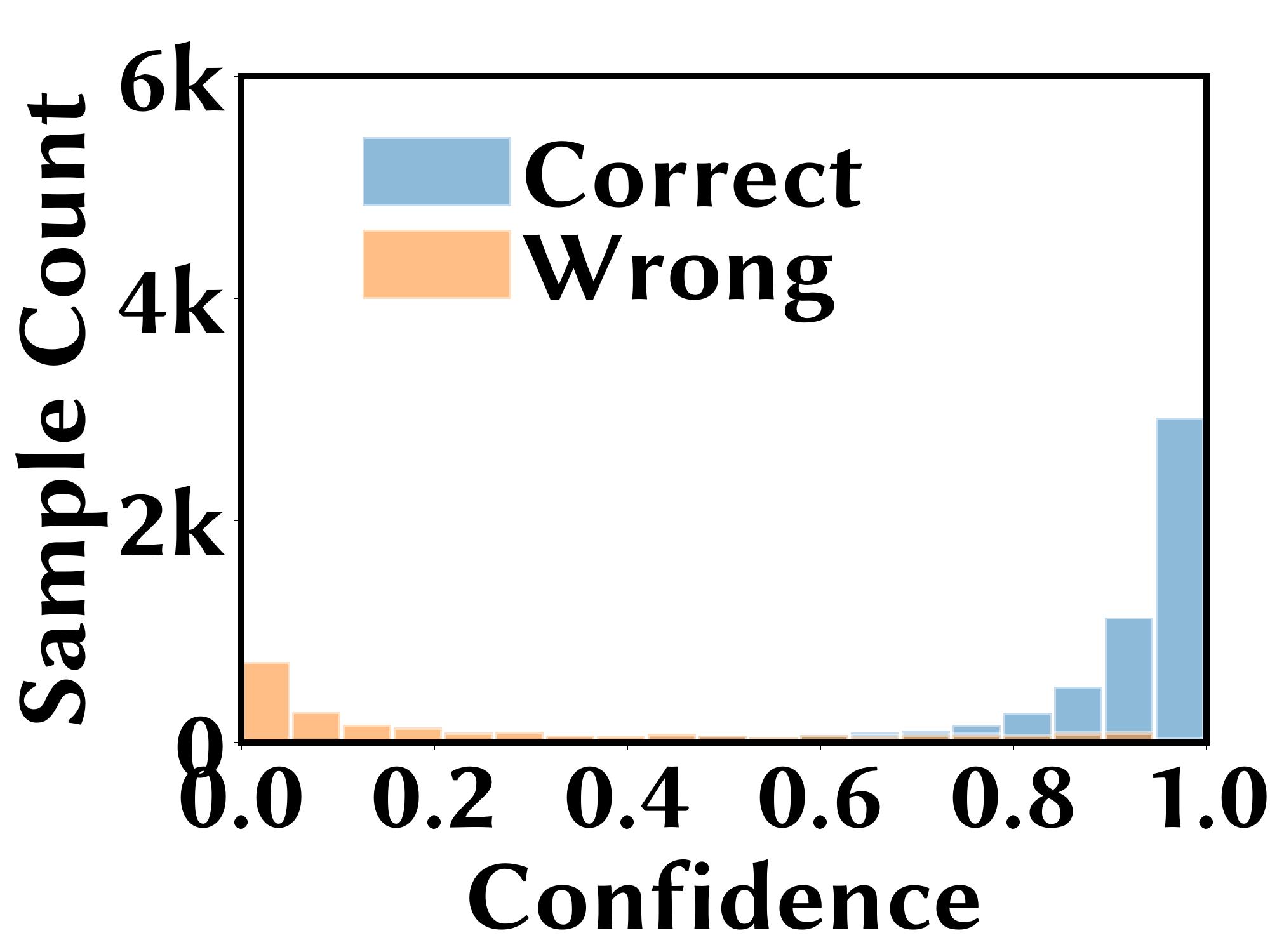}\label{fig:Cifar10_OOD_Calib1_histogram}}
	\vspace{-0.3cm}
	\caption{Histograms of prediction confidences for VGG16 on CIFAR-10 OOD dataset. (a) Un-calibrated. (b) Temperature scaling. (c) Scaling-binning. (d) Dirichlet calibration. (e) \ouralg.}\label{fig:cifar10_OOD_histogram}
	\vspace{-0.3cm}
\end{figure*}

Before describing \ouralg, we first review the confidences calibrated by
 some recent methods ---
temperature scaling \cite{calibration_guo_2017_ICML}, scaling-binning \cite{scaling_binning_calibration_kumar_2019_NIPS}, and Dirichlet calibration \cite{dirichlet_calibration_kull_2019_NIPS} --- whose details are further described in \zhihui{Section~\ref{sec:benchmarks}}.
Concretely, for presentation clarity, we
randomly select
\zhihui{50} samples from each class
and show in Fig.~\ref{fig:cifar10_OOD_tsne}
the t-SNE visualization \cite{tSNE_virtualization_geoffrey_2008} of
both un-calibrated and calibrated softmax probabilities on CIFAR-10 OOD dataset (whose details are available in our experiment results).
We notice from Figs.~\ref{fig:cifar10_tsne_OOD_MP}--\ref{fig:cifar10_tsne_OOD_Dirichlet} that while the softmax probabilities of different classes are well separated, 
wrong samples are still largely mixed together with
correct samples for each class. This is not surprising, since  mis-classified samples may not be corrected by confidence calibration alone (e.g., calibration using temperature scaling
still preserves the original predicted label \cite{calibration_guo_2017_ICML}).
By contrast,  due
to the introduction of an auxiliary class in \ouralg,
we see from Fig.~\ref{fig:cifar10_tsne_OOD_Calib1_print}
that mis-classified samples tend to be more separated from correctly classified ones
as a single cluster.
This point is also reaffirmed in Fig.~\ref{fig:cifar10_OOD_histogram},
where we show the histogram of calibrated confidences.
It can be seen that many mis-classfied samples still have a high calibrated confidence
by using the existing calibration methods, whereas
they tend to have a much lower calibrated confidence by using \ouralg.

\begin{figure*}[!t]
	\centering
	\subfigure[\ouralg]{\includegraphics[trim=0cm -0.5cm 0cm -0.5cm,clip,  height=0.2\textwidth]{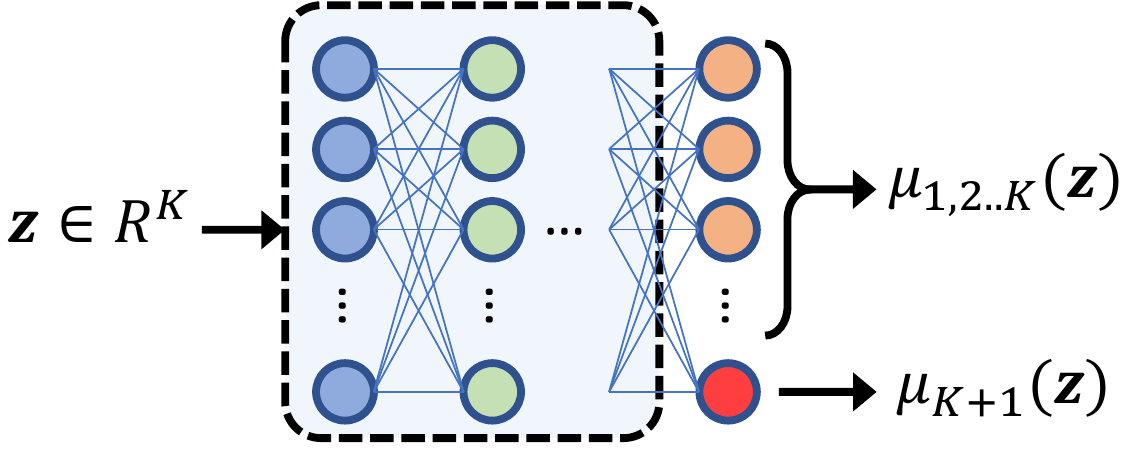}\label{fig:calib1_model}}
	\subfigure[\ouralgtwo]{\includegraphics[trim=-1cm 0cm 0cm 0cm,clip,  height=0.2\textwidth]{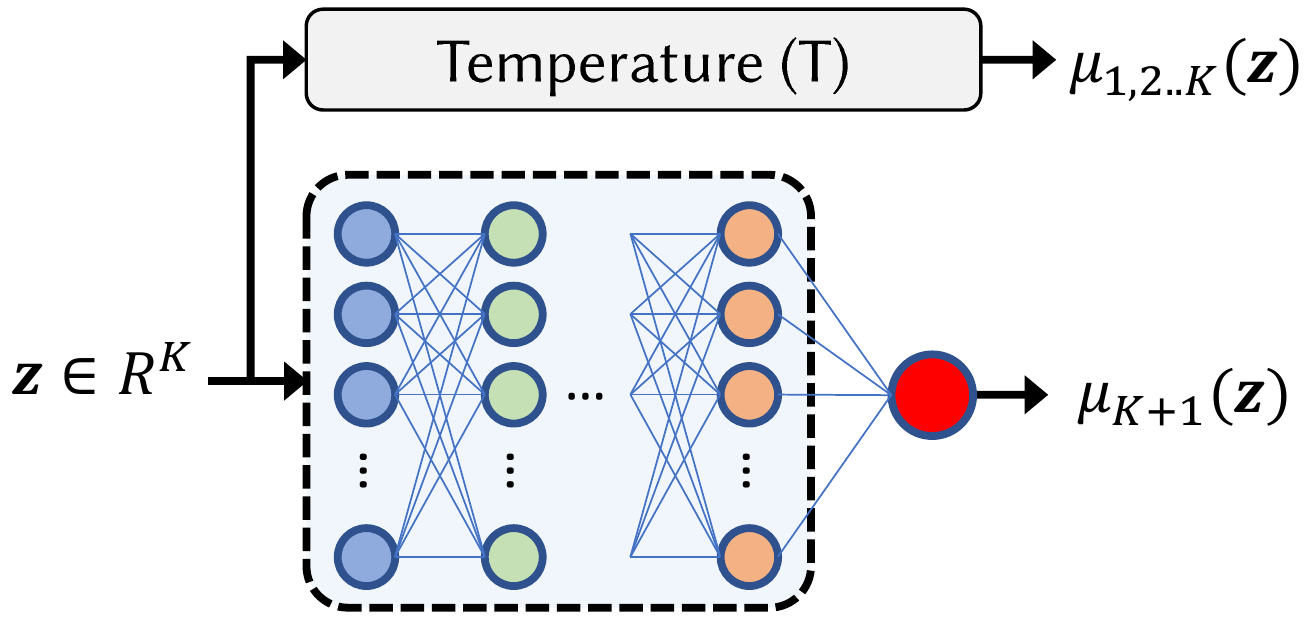}\label{fig:calib2_model}}
	\vspace{-0.3cm}
	\caption{Overview of our confidence calibration models. }\label{fig:models}
	\vspace{-0.3cm}
\end{figure*}

\textbf{Overview of \ouralg}. From the above analysis, we see that assigning
a high confidence to mis-classified samples  is a major
cause for poor calibration on OOD datasets \cite{DNN_Uncertainty_Estimation_Summary_NIPS_2019_ovadia}.
To avoid being confidently wrong, we cannot identify true labels
for those mis-classified samples by only using calibration,
since post-hoc calibration
is not designed for this purpose. 
Instead, we
propose \ouralg (Confidence Calibration with an Auxiliary Class), which separates mis-classified samples from correctly classified ones,
such that mis-classified samples can be assigned with a low confidence.
To accomplish this, we add an \emph{auxiliary class} to represent
samples that are mis-classified by the target DNN, and train a model
to map the logit $\mathbf{z}\in\mathbb{R}^K$ (produced by the target DNN) to a calibrated probability $\mu(\mathbf{z})\in\mathbb{R}^{K+1}$.
Specifically, in \ouralg,
$\mu_{k}(\mathbf{z})$
and $\mu_{K+1}(\mathbf{z})$ represent the likelihoods of class
$k$ (for $k=1,\cdots, K$) and mis-classification, respectively,
with
$\sum_{k=1}^{K+1}\mu_k(\mathbf{z})=1$.
Then,
by combining the calibrated probabilities for both the predicted label
and mis-classification, we obtain the calibrated confidence. 
Our design of \ouralg is also illustrated in Fig.~\ref{fig:models}.

\textbf{Loss function for \ouralg}.
Given the target DNN's $K$-dimensional logit $\mathbf{z}$,
our calibration model
maps
  $\mathbf{z}$ to a $(K+1)$-dimensional calibrated softmax probability $\mu(\mathbf{z})$.
In \ouralg, we extend the original $K$-class label $y$
to a $(K+1)$-class label $w$,
which includes the same set of $K$ original classes and an additional auxiliary class $K+1$ that represents
 samples mis-classified by the target DNN.
Specifically,
if a sample $(\mathbf{x},y)$ belonging to class $k$ is correctly classified
by the target DNN,
its label
$w$ in \ouralg is still class $k$ (i.e., $w=y$);
otherwise, its label $w$ becomes ``class $(K+1)$'' (i.e., class of mis-classified samples).
We use one-hot encoding $\mathbf{w}=[w_1,\cdots,w_{K+1}]$ to represent the label in \ouralg, i.e., $w_k=1$ if the sample belongs to class $k$ and $w_k=0$ otherwise, for $k=1,\cdots,K+1$.  We
consider the following  cross-entropy loss:
\begin{equation}\label{eqn:calibration_loss}
\mathcal{L} = -\sum_{k=1}^{K+1}w_k \log\left({\mu}_{k}\right),
\end{equation}
where $\mu_k$ is the calibrated softmax probability $\mu(\mathbf{z})$
for a sample belonging to class $k$ in \ouralg.

In Eqn.~\eqref{eqn:calibration_loss}, some correctly classified
samples can also have a large probability $\mu_{K+1}$  of being considered
as mis-classified.
To prevent correctly classified samples from having a large  $\mu_{K+1}$,
we can modify the loss by adding a regularization term
$-(1-w_{K+1})\log(1-{\mu}_{K+1})$ into the loss function.
For a correctly classified sample, we have $w_k=1$ for its true class
and meanwhile $w_{K+1}=0$. Thus,
the added regularization $-(1-w_{K+1})\log(1-{\mu}_{K+1})$ is
only effective for correctly classified samples
and aims at pushing them
 away from class $K+1$ (i.e., reducing $\mu_{K+1}$).
Additionally, we can add another weight for the auxiliary class $K+1$.
Thus,  we consider a modified loss function as follows:
\begin{equation}\label{eqn:calibration_loss_modified}
\mathcal{L} = -\sum_{k=1}^{K}w_k \log\left({\mu}_{k}\right)-\lambda_1(1-w_{K+1})\log(1-{\mu}_{K+1})-\lambda_2w_{K+1}\log\left({\mu}_{K+1}\right),
\end{equation}
where $\lambda_1\geq0$ and $\lambda_2\geq0$ are tunable hyperparameters.

\begin{figure*}[!t]
	\centering
	\subfigure[Un-calibrated]{\includegraphics[trim=0cm 0cm 0cm 0cm,clip,  width=0.24\textwidth]{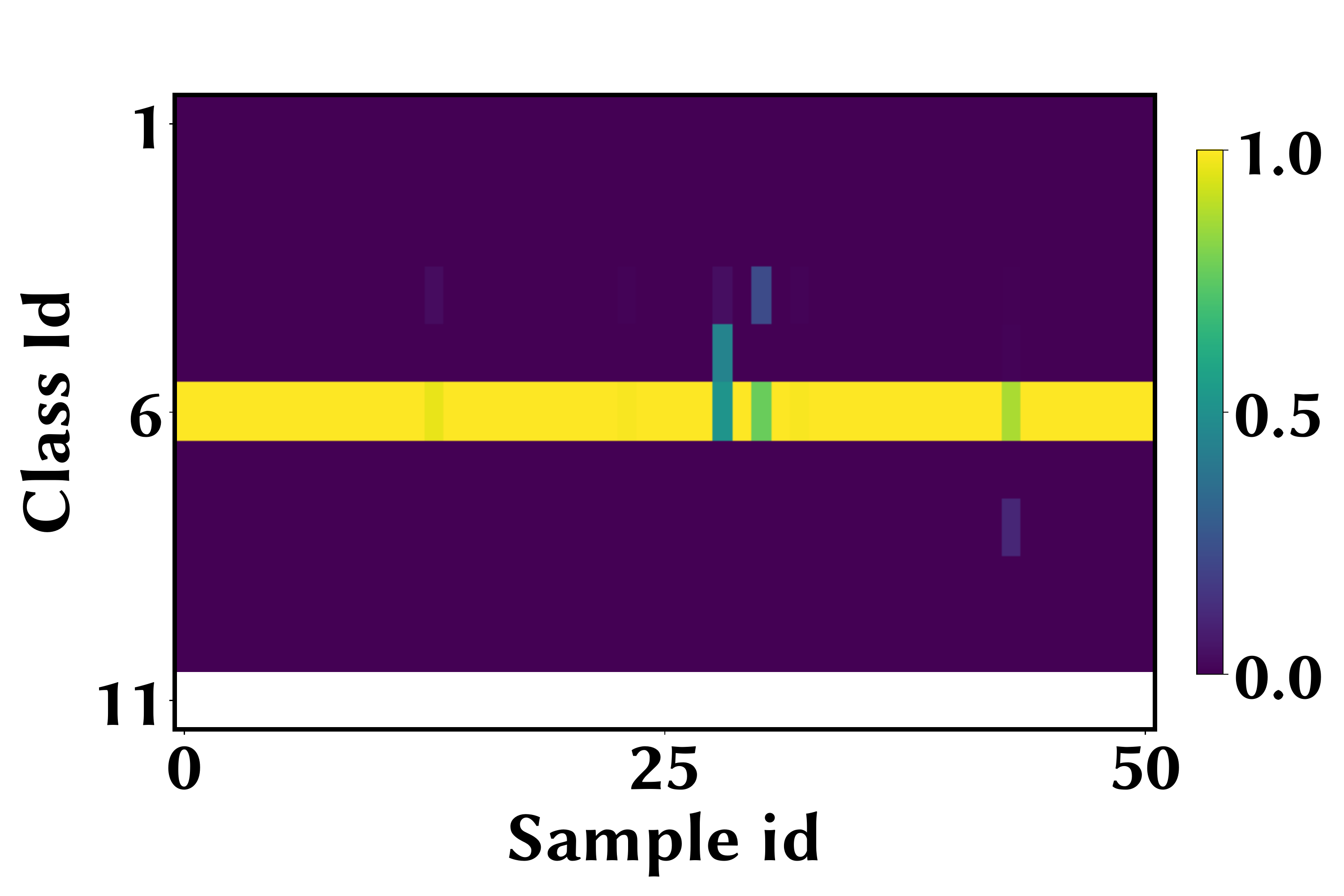}\label{fig:Cifar10_OOD_Calib1_heatmap_True_before}}	
	\subfigure[Calibrated by \ouralg]{\includegraphics[trim=0cm 0cm 0cm 0cm,clip,  width=0.24\textwidth]{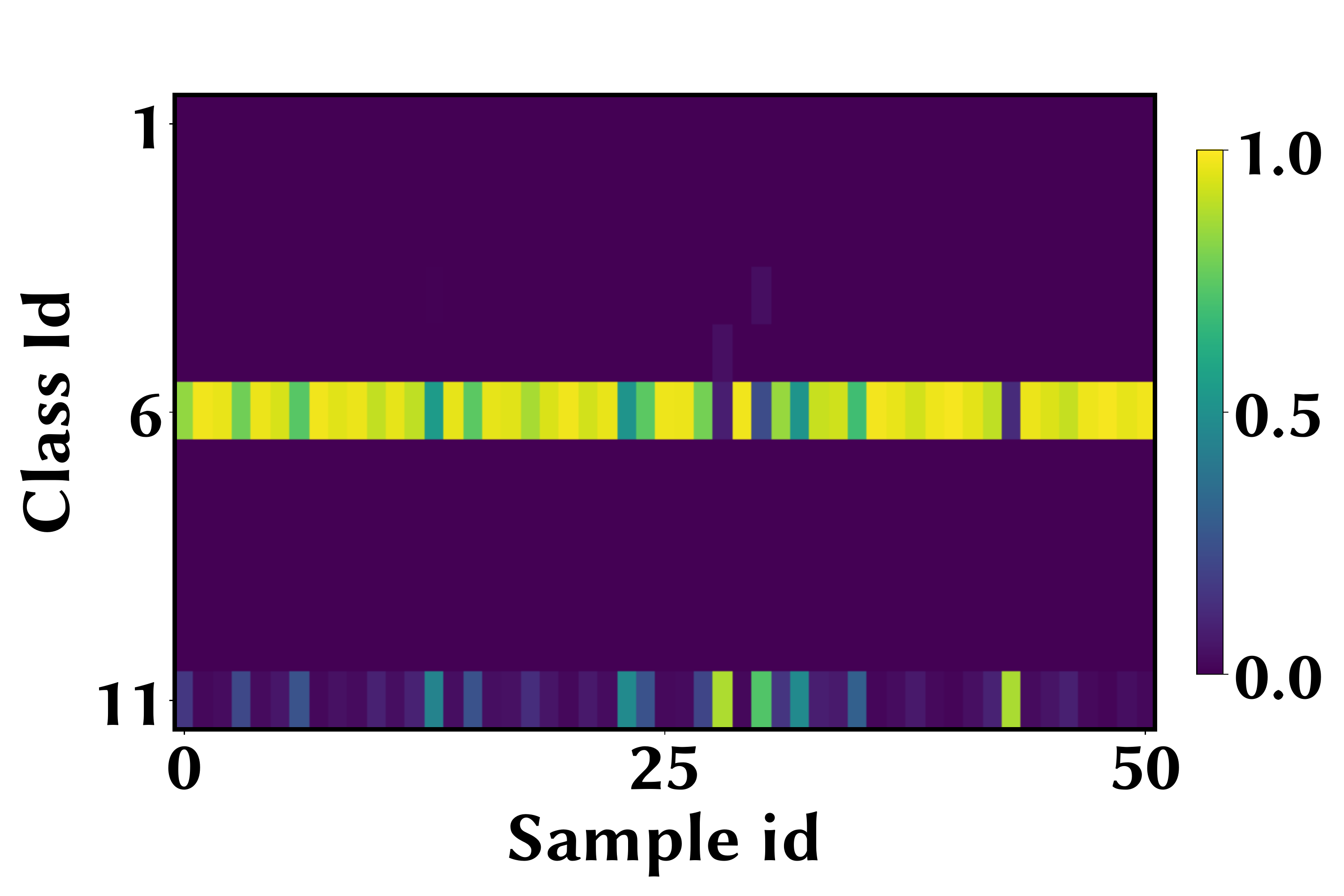}\label{fig:Cifar10_OOD_Calib1_heatmap_True_after}}	
	\subfigure[Un-calibrated]{\includegraphics[trim=0cm 0cm 0cm 0cm,clip,  width=0.24\textwidth]{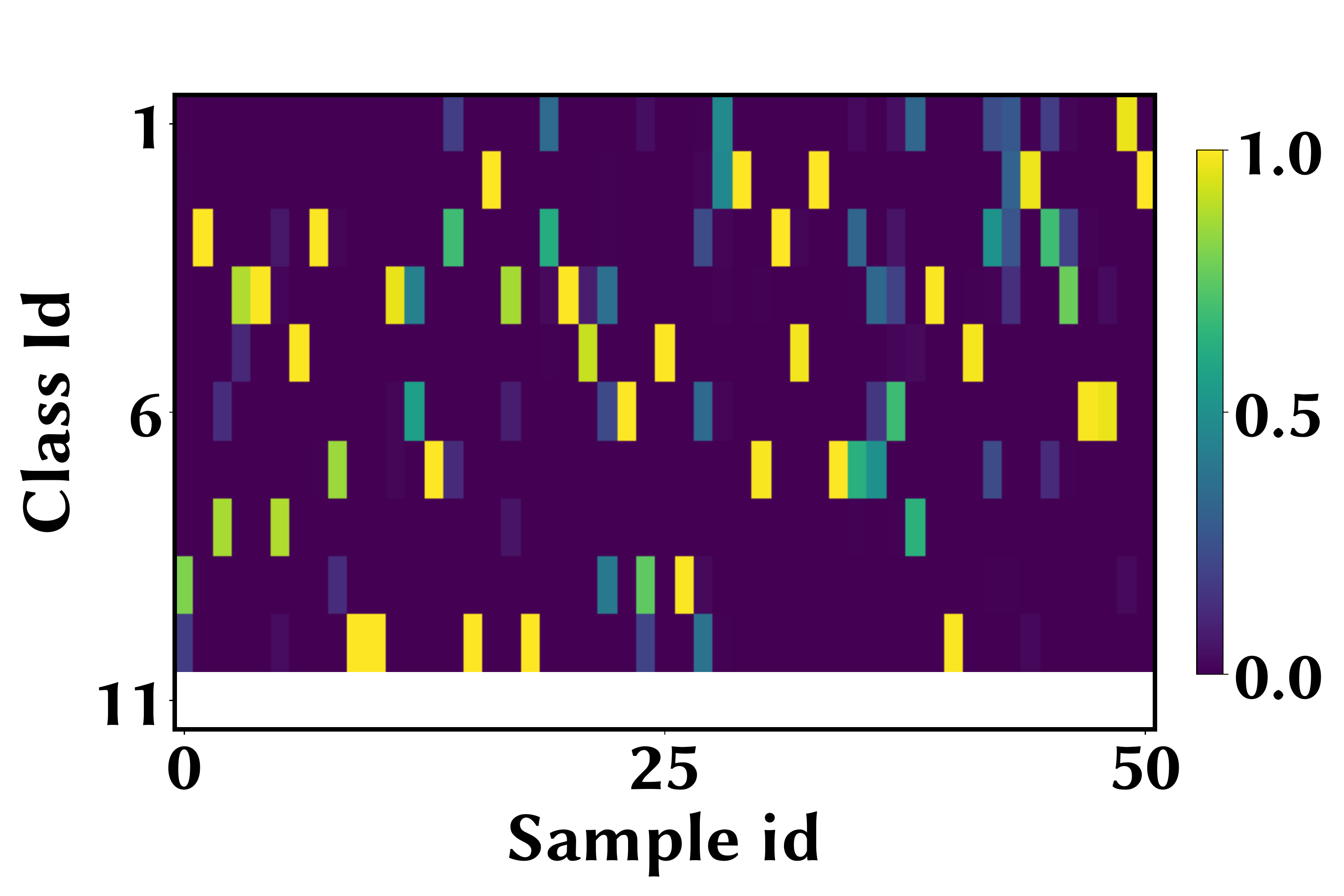}\label{fig:Cifar10_OOD_Calib1_heatmap_False_before}}	
	\subfigure[Calibrated by \ouralg]{\includegraphics[trim=0cm 0cm 0cm 0cm,clip,  width=0.24\textwidth]{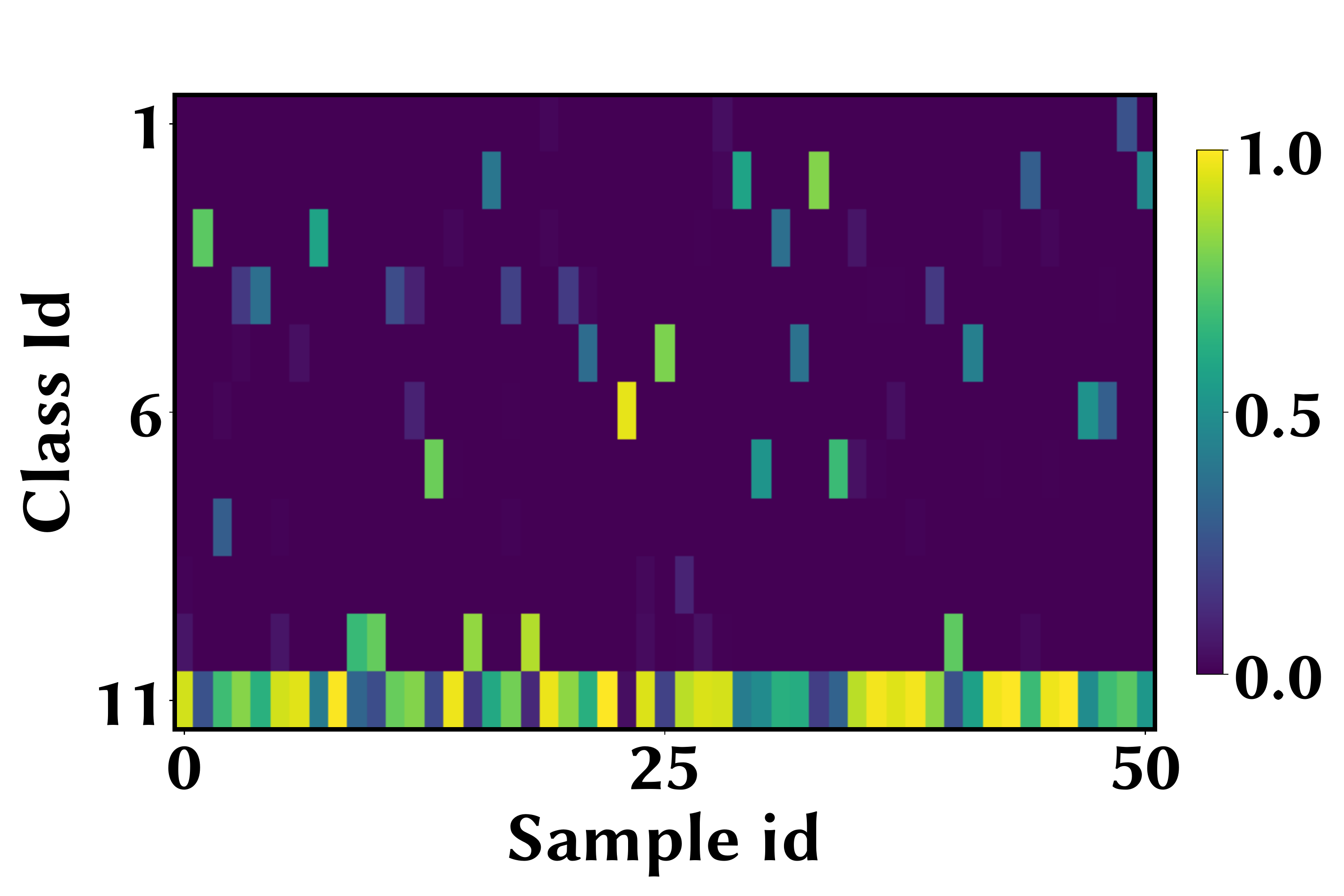}\label{fig:Cifar10_OOD_Calib1_heatmap_False_after}}
	\vspace{-0.3cm}
	\caption{Softmax probabilities for VGG16 on CIFAR-10 OOD dataset. (a)(b)
Results for 50 correctly classified samples in class 6. (c)(d) Results
for 50 mis-classified samples.
}\label{fig:heatmap_demo}
	\vspace{-0.3cm}
\end{figure*}

Next, we highlight the importance of our introduced auxiliary class
by showing in Fig.~\ref{fig:heatmap_demo} the uncalibrated/calibrated probabilities for 50 correctly classified samples (belonging
to one class)
and 50 mis-classified samples. We see that after calibration by \ouralg,
the correctly classified samples can largely keep their high probabilities
for the true class. On the other hand, \ouralg can significantly
reduce the probabilities of false classes for mis-classified samples,
pushing mis-classified samples to have a high $\mu_{K+1}$ and effectively mitigating ``confidently wrong''.

\textbf{Calibrated confidence}. If all
samples mis-classified by the target DNN are perfectly
identified and assigned
with $\mu_{K+1}=1$, then we could directly use $\mu_k$ as the calibrated
confidence. But, this is not practically possible. Here, without
modifying the original prediction label,
we propose to combine $\mu_{K+1}$ together with the calibrated softmax probability $\mu_{\hat{y}}$
for the label $\hat{y}\in\{1,\cdots,K\}$ predicted  by the target DNN. Specifically, we consider
two different types of geometric means for the calibrated
confidence:
\begin{equation}\label{eqn:confidence_calculate1}
confidence = 1-\sqrt{(1-\mu_{\hat{y}})\mu_{K+1}} \text{\;\;\;\; and \;\;\;\;}
confidence = \sqrt{\mu_{\hat{y}}(1-\mu_{K+1})}.
\end{equation}
The interpretation for $confidence = 1-\sqrt{(1-\mu_{\hat{y}})\mu_{K+1}}$
is as follows: given a sample classified
by the target DNN as class $\hat{y}$,
$1-\mu_{\hat{y}}$ and $\mu_{K+1}$ can both represent the probability
of mis-classification, $\sqrt{(1-\mu_{\hat{y}})\mu_{K+1}}$
is the geometric mean, and hence $1-\sqrt{(1-\mu_{\hat{y}})\mu_{K+1}}$
can be used as a calibrated confidence. The interpretation for $confidence = \sqrt{\mu_{\hat{y}}(1-\mu_{K+1})}$ is also similar.

\textbf{Training of \ouralg}.
As shown in Fig.~\ref{fig:calib1_model}, we can use a neural \zhihui{network}
 for minimizing the loss. 
Specifically, with labeled samples
from the actual test dataset to which the target DNN is applied,
we adopt standard training algorithms to construct the neural network.
Suppose that \ouralg uses $L$ fully connected layers each with $K$ nodes.
Then, given $K$ put nodes and $K+1$ output nodes, \ouralg needs to learn
$\sim O(L K^2)$ weight parameters.
Additionally, we also use a small validation dataset for
tuning hyperparameters  and deciding which  of two calibrated confidences
in Eqn.~\eqref{eqn:confidence_calculate1} to use.

\textbf{Simplified \ouralg and transfer}. When applying the target DNN to
a new test dataset, labeling samples to learn $\sim O(L K^2)$ weight parameters
in \ouralg may be too expensive. Thus, to reduce
the number of labeled samples needed, we propose a simplified
calibration model, called \ouralgtwo (\ouralg-Simplified), which
is illustrated
in Fig.~\ref{fig:calib2_model}. Specifically, the logit
input $\mathbf{z}$ is scaled using a temperature parameter $T$
 for the first $K$ classes in \ouralgtwo,
and the logit for the $(K+1)$-th class is obtained using a neural
network. Then,
 the $K+1$ logits are merged together, based on which
 the calibrated softmax probability $\mu$ is derived.
Note that for training both \ouralg and \ouralgtwo, we
assign ``class $(K+1)$'' to samples mis-classified
by the target DNN, while keeping the labels of those correct samples unchanged.

Another advantage of \ouralgtwo is its easy transferability. Specifically,
we can first train \ouralgtwo on a given labeled dataset. Then,
when applying the target DNN to a new unseen dataset,
we freeze all the parameters in \ouralgtwo, except for the temperature
parameter $T$ (for the first $K$ classes) and the last
fully connected layer in the neural network (for the $(K+1)$-th class).
Suppose that the penultimate layer in the neural network has $K$ nodes.
We only need to learn a total of $K+2$ weight parameters (temperature
plus $K+1$ parameters for the last layer), which requires significantly 
fewer samples than learning weights for \ouralg and \ouralgtwo from scratch.
Thus, for a new unseen dataset, we only need to label a small set of samples
from the target test dataset.

\section{Experiments}

We conduct experiments for different DNNs on various datasets with both
 image and document classification applications.
The results show that our approach can consistently outperform
recent post-hoc calibration methods in terms of
the calibration performance.
\subsection{Methodologies}

We train  both \basicModel and \advanceModel  in \textit{Tensorflow} \cite{tensorflow_2016} with \textit{Keras} layers \cite{keras_2015}.
All experiments are executed within Jupiter Notebook under the Anaconda environment. 

\subsubsection{Baseline Approaches}\label{sec:benchmarks}

Our proposed method belongs to post-hoc confidence calibration. Thus,
for fair comparison, we
 consider the following state-of-the-art post-hoc calibration approaches as baselines.

\textbf{Max probability (MP): }The prediction confidence  by MP is not calibrated and directly calculated as $MP=\max_{k\in\{1,2...K\}}\{{p}_k\}$,
 where $p_k$ is the target DNN's softmax probability for class $k$ \cite{DNN_Uncertainty_Baseline_OOD_ICLR_2017}.

\textbf{Temperature scaling (TS): } TS provides calibrates confidence by learning a single multiplicative factor $T$ (a.k.a. scaling temperature) on the target DNN's logit output $\textbf{z}$ \cite{calibration_guo_2017_ICML}. The calibrated confidence ${p}_T$ can be presented as ${p}_{T}=\max\sigma_{SM}(\textbf{z}/T)$. We
learn the temperature $T$ in TS
by minimizing a negative log-likelihood (NLL) loss \cite{calibration_guo_2017_ICML}.

\textbf{Scaling-binning (SB): } The SB calibrator combines two popular post-hoc calibration methods: platter scaling and histogram binning
\cite{scaling_binning_calibration_kumar_2019_NIPS}. Like in TS,
we also use
the same dataset utilized for training \ouralg to
train the calibration function for each bin in SB.
\textbf{Dirichlet calibration:} The recently proposed Dirichlet calibration \cite{dirichlet_calibration_kull_2019_NIPS} extends Beta calibration \cite{beta_calibration_kull_2017} to $K$-class classifiers. Specifically, a calibration model is learned to map $K$-dimensional the un-calibrated softmax probability $\textbf{p}$ into a well-calibrated confidence/probability ${\mu}(\textbf{p})$. The mapping function includes $K$ Dirichlet functions with $O(K^2)$ parameters learnt
by minimizing the NLL loss.

\subsubsection{Evaluation Metrics}

\textbf{Confidence calibration metrics:} We consider
the commonly-used calibration metrics --- expected calibration error (\textbf{ECE}) and Brier Score (\textbf{BS})
\cite{dirichlet_calibration_kull_2019_NIPS,calibration_guo_2017_ICML}.
Specifically, the test samples are firstly grouped into $M$ equal-width bins according to their (calibrated) confidence scores. Within the bin $B_m, m\in\{1,2...M\}$, the average prediction accuracy $Acc(B_m)$ and the average confidence $Conf(B_m)$ can be empirically
calculated as $Acc(B_m) = \frac{1}{|B_m|}\sum_{i\in B_m}{\textbf{1}(\hat{y_i}=y_i)}$
and $Conf(B_m) = \frac{1}{|B_m|}\sum_{i\in B_m}{\hat{c_i}}$, respectively.
Then,
the ECE value can be calculated as the weighted-average of the absolute difference between $Acc(B_m)$ and $Conf(B_m)$, i.e., $ECE = \sum_{m1}^{M}\frac{|B_m|}{N} |Acc(B_m)-Conf(B_m)|$, where $N$ is the total number of test samples and
$|B_m|$ represents the sample counts in bin $B_m$ \cite{calibration_guo_2017_ICML}.
In our experiments, we use $M=20$ bins for reliability diagrams and ECE calculation.
The BS estimates the mean squared error between correctness of prediction and confidence score \cite{brier_score_books_1950}, and can be calculated
as $BS(\mathcal{D})=\frac{1}{|\mathcal{D}|}\sum_{i\in \mathcal{D}}(\textbf{1}(\hat{y_i}=y_i)-\hat{c_i})^2$, where
 $\hat{c_i}$ is the confidence for an input $\textbf{x}_i$, $\mathcal{D}$ represents the entire test dataset and $\textbf{1}(\hat{y_i}=y_i)$ indicates if the classification
 for input $\textbf{x}_i$ is correct or not.
For both ECE and BS metrics, the lower value,  the better.

\textbf{Misclassification detection metrics:} While
our main purpose is confidence calibration, a byproduct
of our method is the better detection of mis-classified samples based on
a threshold of the calibrated confidence.
We consider \textbf{AUROC} (Area Under the Receiver Operating Characteristic) and \textbf{AUPR} (Area Under the Precision-Recall Curve)  as metrics for mis-classification  detection, since they are independent of the thresholds.
A higher AUROC/AUPR implies a better distinguishability between correctly
classified and mis-classified samples. In addition,
we also use the precision at 90\% recall (denoted as $\mathbf{p{.9}}$) as an evaluation metric, which represents the precision when we aim at detecting 90\% of the mis-classified samples.
Note
that a mis-classified sample is treated as ``positive'' in mis-classification detection when calculating AUROC, AUPR and $\mathbf{p{.9}}$.

\subsection{Results for Image Classification}

We consider 10-class, 100-class and 1000-class image classifications.

\subsubsection{10-class Image Classification  with VGG16 DNN}\label{sec:10class_vgg}

\textbf{Target DNN model. } The target DNN model is a modified VGG16 for tiny images (CIFAR). The target model includes batch normalization layers and dropout layers to improve classification performance. It is trained in \textit{TensorFlow} on CIFAR-10 training dataset (50k images). The pre-trained weights of target DNN model are downloaded from \cite{vgg16_weights_github_2018}.

\textbf{Datasets.} We evaluate the calibration performance on four datasets generated from CIFAR-10 and CIFAR-100 \cite{cifar10_100_dataset_2009}, including two augmented datasets (D1 and D2), one out-of-distribution dataset (OOD) and one adversarial dataset (AD). We first randomly select 30k samples from CIFAR-10 training dataset and 10k samples from CIFAR-10 testing dataset, and then perform augmentation operations on the selected samples. For D1, the augment operations and parameters include rotation within $[-20,20]$ degrees, vertical/horizontal shift within $[-0.25,0.25]$, zoom with $0.4$, and horizontal flip. For D2, the augmentation parameters are: rotation within $[-40,40]$ degrees, vertical/horizontal shift within $[-0.4,0.4]$, zoom with $0.5$, and horizontal flip. The OOD dataset also includes 40k samples, with 12k OOD samples randomly selected from CIFAR-100 training dataset and 28k samples from CIFAR-10 training dataset. As for 12k OOD samples from CIFAR-100, the true labels are mapped to CIFAR-10's class labels. Specifically, we consider OOD samples with label ``pick-up truck'' as ``truck'' in CIFAR-10 and OOD samples with label ``automobile'' as ``bus'' in CIFAF-10. The other OOD samples are treated with ``NULL'' label, indicating not belonging to any of the 10 classes in CIFAR-10. For the AD dataset, the 40k samples include 20k normal samples and  20k adversarial samples. The 20k normal samples contain 10k images from the CIFAR-10 training dataset and 10k images from the CIFAR-10 testing dataset, while
the 20k adversarial samples are generated based on 20k randomly selected
samples from the CIFAR-10 training dataset with DeepFool-Attack using \textit{foolbox} package \cite{adversarial_toolbox_foolbox_metthias_ICML_2017}. The inference accuracies on the four test datasets are 80\% (D1), 62\% (D2), 71\% (OOD), and 47\% (AD). Additionally, the 40k samples (in D1, D2, OOD or AD) are randomly split into a 30k dataset for training, a 2k validation dataset for hyperparameter tuning, and a 8k testing dataset for performance evaluation.

\textbf{Baselines. } For each dataset, the baselines of TS, SB, and Dirichlet calibration are trained on the respective 30k training dataset. 
For Dirichlet calibration, the regularization hyperparameters are tuned with a minimal ECE on the 2k validation dataset. The 30k training and 2k validation datasets are the same as those used for
training \ouralg and \ouralgtwo.

\textbf{Our method. } The neural network of \ouralg is implemented with 2 hidden layers, including 50 hidden neurons in first hidden layer and 20 hidden nodes in the second layer.
The input layer contains 10 nodes and the output layer contains 11 nodes (including 10 classes and one ``mis-classification'' class). \ouralg is trained over 1000 epochs using
the Adam optimizer (learning rate $10^{-3}$). The loss function hyperparameters ($\lambda_1$ and $\lambda_2$) and the confidence calculation method are selected with a minimal ECE on the validation dataset. For \ouralgtwo, the neural network is implemented with the same structure as \basicModel, but the output layer contains only 1 node, representing the ``mis-classification'' class. The temperature scaling layer is implemented with a self-defined customer Keras layer with one learnable weight $T$. The training settings of \ouralgtwo are the same as those of \basicModel. For the transferred model \transferMethod, we first pre-train \advanceModel on dataset D1. Then, when applied to another target dataset, a set of 320 samples are randomly selected from the respective 30k training dataset for model transfer. In addition, we select another 200 samples from the 2k validation dataset  to tune corresponding hyperparameters and confidence calculation method in \ouralgthree.

\begin{table}[!t]
	\centering
	
	\setlength{\tabcolsep}{1.5pt}
	\parbox{.48\linewidth}{
		\centering
		\caption{VGG16 on CIFAR-10 D1}
		\label{table:cifar10_in_ECE_D1}
		\begin{tabular}{lccccc}
			\hline
			\textbf{Method} & \textbf{AUROC} & \textbf{AUPR}  & \textbf{p.9}   & \textbf{ECE}   & \textbf{BS} \\
			\hline
			\textbf{MP}    & 0.865 & 0.566 & 0.419 & 13.9\% & 0.151 \\
			\textbf{TS}    & 0.870 & 0.586 & 0.421 & 4.6\% & 0.116 \\
			\textbf{SB}    & 0.872 & 0.589 & 0.429 & 2.9\% & 0.113 \\
			\textbf{Dirichlet} & 0.870 & 0.586 & 0.422 & 4.5\% & 0.116 \\
			\textbf{\basicModel} & \textbf{0.907} & \textbf{0.715} & 0.489 & \textbf{1.4\%} & \textbf{0.095} \\
			\textbf{\advanceModel} & 0.906 & 0.696 & \textbf{0.493} & 1.9\% & 0.097\\
			\textbf{\transferMethod} & - & - & - & - & -\\
			\hline
		\end{tabular}%

	}
	\hfill
	\parbox{.48\linewidth}{
		\centering
		\caption{VGG16 on CIFAR-10 D2}
		\label{table:cifar10_in_ECE_D2}
		\begin{tabular}{lccccc}
			\hline
			\textbf{Method} & \textbf{AUROC} & \textbf{AUPR}  & \textbf{p.9}   & \textbf{ECE}   & \textbf{BS} \\
			\hline
			\textbf{MP}    & 0.814 & 0.664 & 0.560 & 28.2\% & 0.280 \\
			\textbf{TS}    & 0.823 & 0.677 & 0.573 & 9.2\% & 0.174 \\
			\textbf{SB}    & 0.839 & 0.688 & 0.606 & 3.6\% & 0.158 \\
			\textbf{Dirichlet} & 0.823 & 0.677 & 0.574 & 8.7\% & 0.173 \\
			\textbf{\basicModel} & \textbf{0.879} & {0.773} & 0.649 & \textbf{2.3\%} & \textbf{0.137} \\
			\textbf{\advanceModel} & 0.878 & 0.764 & \textbf{0.652} & 3.0\% & 0.138\\
			\textbf{\transferMethod} & 0.876 & \textbf{0.775} & 0.642 & 3.8\% & 0.140  \\
			\hline
		\end{tabular}%

	}
\end{table}

\begin{table}[t!]
	\centering
	\setlength{\tabcolsep}{1.5pt}
	\parbox{.48\linewidth}{
		\centering
		\caption{VGG16 on  CIFAR-10 OOD}
		\label{table:cifar10_in_ECE_OOD}
		\begin{tabular}{lccccc}
			\hline
			\textbf{Method} & \textbf{AUROC} & \textbf{AUPR}  & \textbf{p.9}   & \textbf{ECE}   & \textbf{BS} \\
			\hline
			\textbf{MP}    & 0.923 & 0.816 & 0.640 & 24.3\% & 0.225 \\
			\textbf{TS}    & 0.927 & 0.833 & 0.650 & 14.4\% & 0.140 \\
			\textbf{SB}    & 0.921 & 0.834 & 0.604 & 11.8\% & 0.119 \\
			\textbf{Dirichlet} & 0.923 & 0.816 & 0.640 & 15.9\% & 0.126 \\
			\textbf{\basicModel} & \textbf{0.940} & \textbf{0.870} & \textbf{0.674} & \textbf{2.0\%} & \textbf{0.087} \\
			\textbf{\advanceModel} & 0.934 & 0.861 & 0.645 & 2.0\% & 0.091 \\
			\textbf{\transferMethod} & 0.894 & 0.802 & 0.512 & 3.1\% & 0.108\\
			\hline
		\end{tabular}%
		
	}
	\hfill
	\parbox{.48\linewidth}{
		\centering
		\caption{VGG16 on CIFAR-10 AD}		
		\label{table:cifar10_in_ECE_AD}
		\begin{tabular}{lccccc}
			\hline
			\textbf{Method} & \textbf{AUROC} & \textbf{AUPR}  & \textbf{p.9}   & \textbf{ECE}   & \textbf{BS} \\
			\hline
			\textbf{MP}    & 0.923 & 0.889 & 0.839 & 48.9\% & 0.461 \\
			\textbf{TS}    & 0.913 & 0.885 & 0.825 & 34.1\% & 0.230 \\
			\textbf{SB}    & 0.890 & 0.876 & 0.770 & 20.5\% & 0.197 \\
			\textbf{Dirichlet} & 0.685 & 0.657 & 0.625 & 8.8\% & 0.220 \\
			\textbf{\basicModel} & \textbf{0.935} & \textbf{0.926} & 0.856 & 3.5\% & \textbf{0.097} \\
			\textbf{\advanceModel} & 0.934 & 0.925 & \textbf{0.860} & \textbf{3.3\%} & 0.098\\
			\textbf{\transferMethod} & 0.872 & 0.864 & 0.749 & 3.8\% & 0.143 \\
			\hline
		\end{tabular}%
		
	}
\end{table}

\begin{figure*}[!t]
	\centering
	\subfigure[]{\includegraphics[trim=0cm 0cm 0cm 0cm,clip,  width=0.195\textwidth]{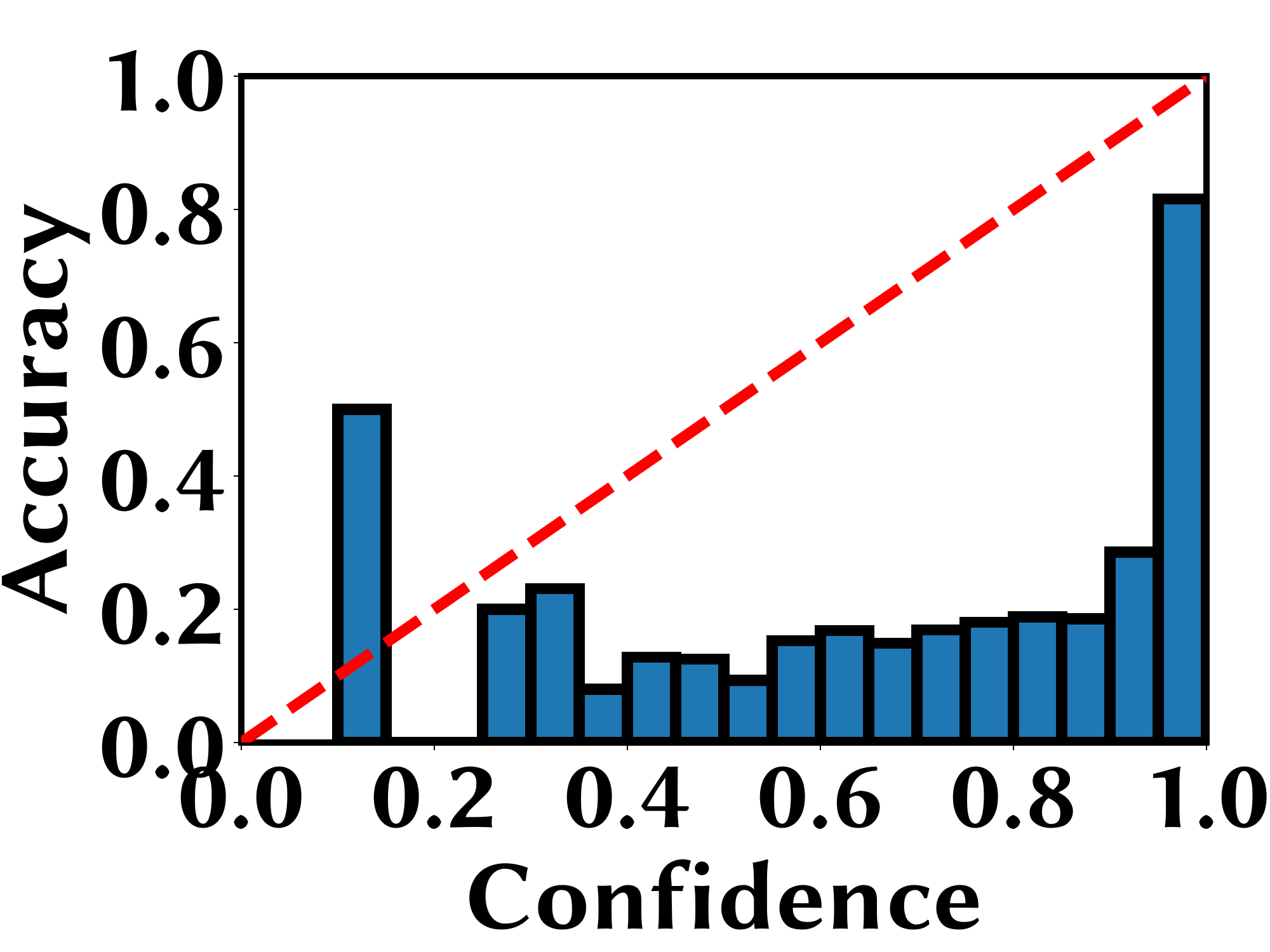}}
	\subfigure[]{\includegraphics[trim=0cm 0cm 0cm 0cm,clip,  width=0.195\textwidth]{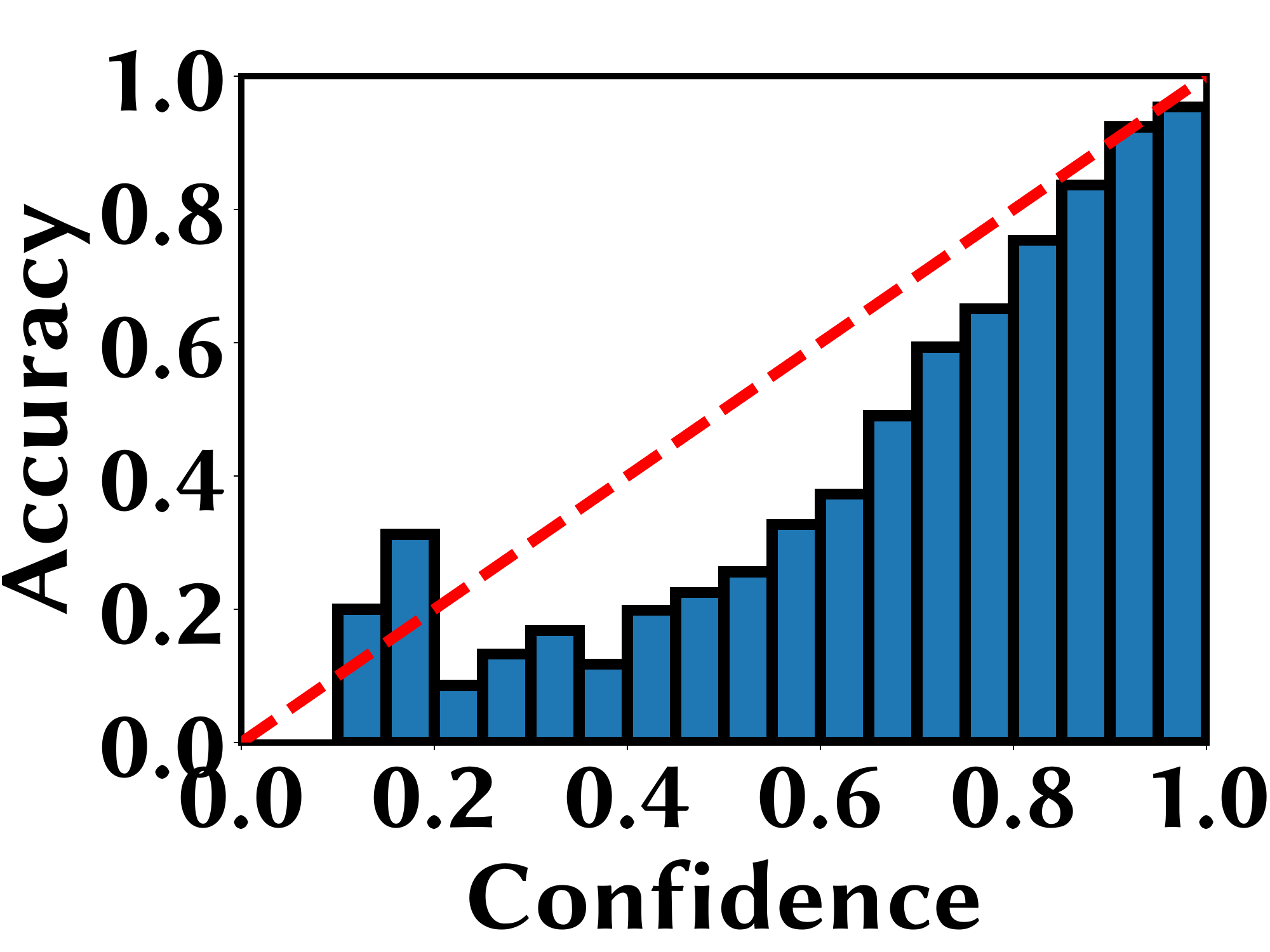}}
	\subfigure[]{\includegraphics[trim=0cm 0cm 0cm 0cm,clip,  width=0.195\textwidth]{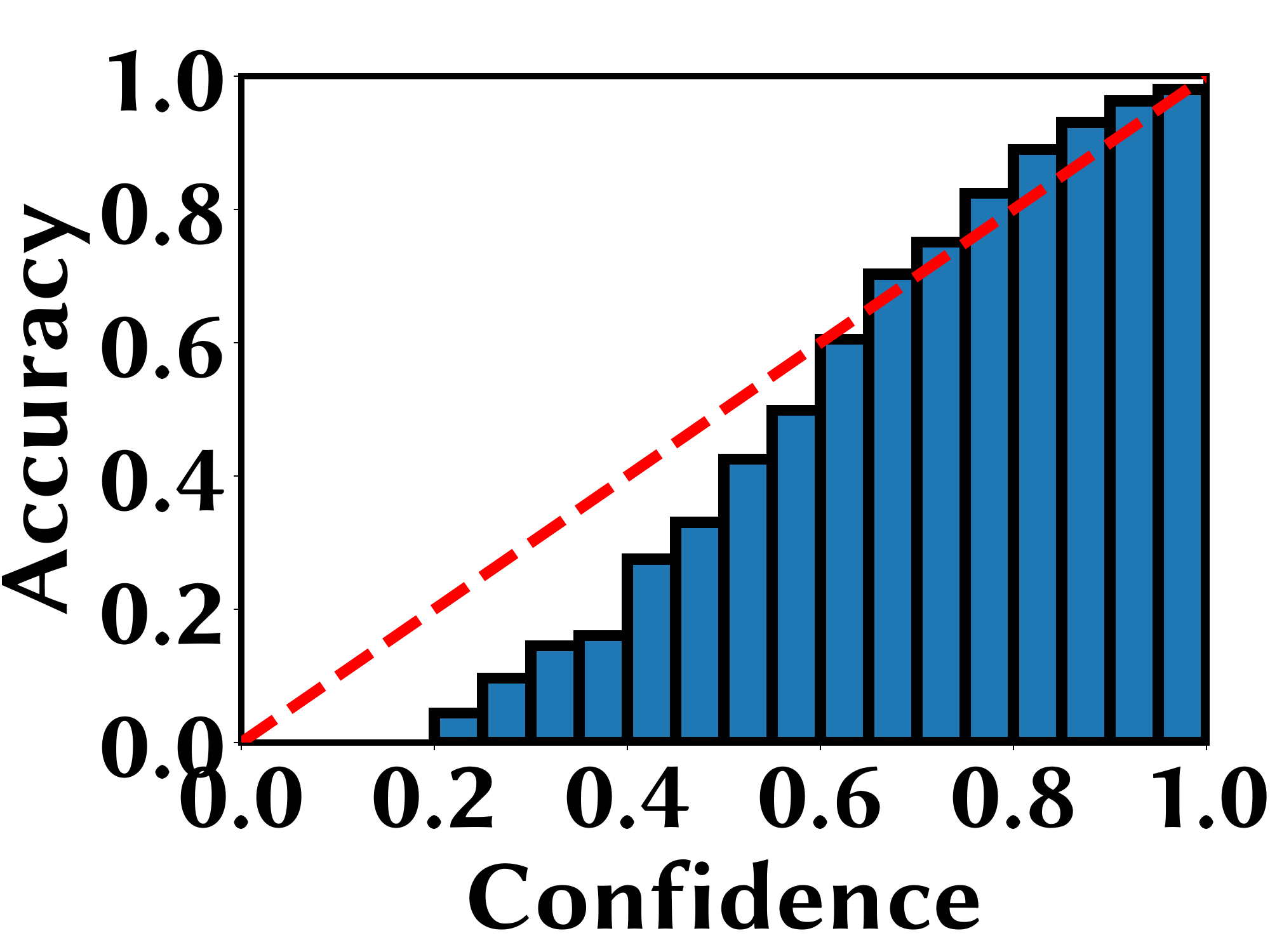}}
	\subfigure[]{\includegraphics[trim=0cm 0cm 0cm 0cm,clip,  width=0.195\textwidth]{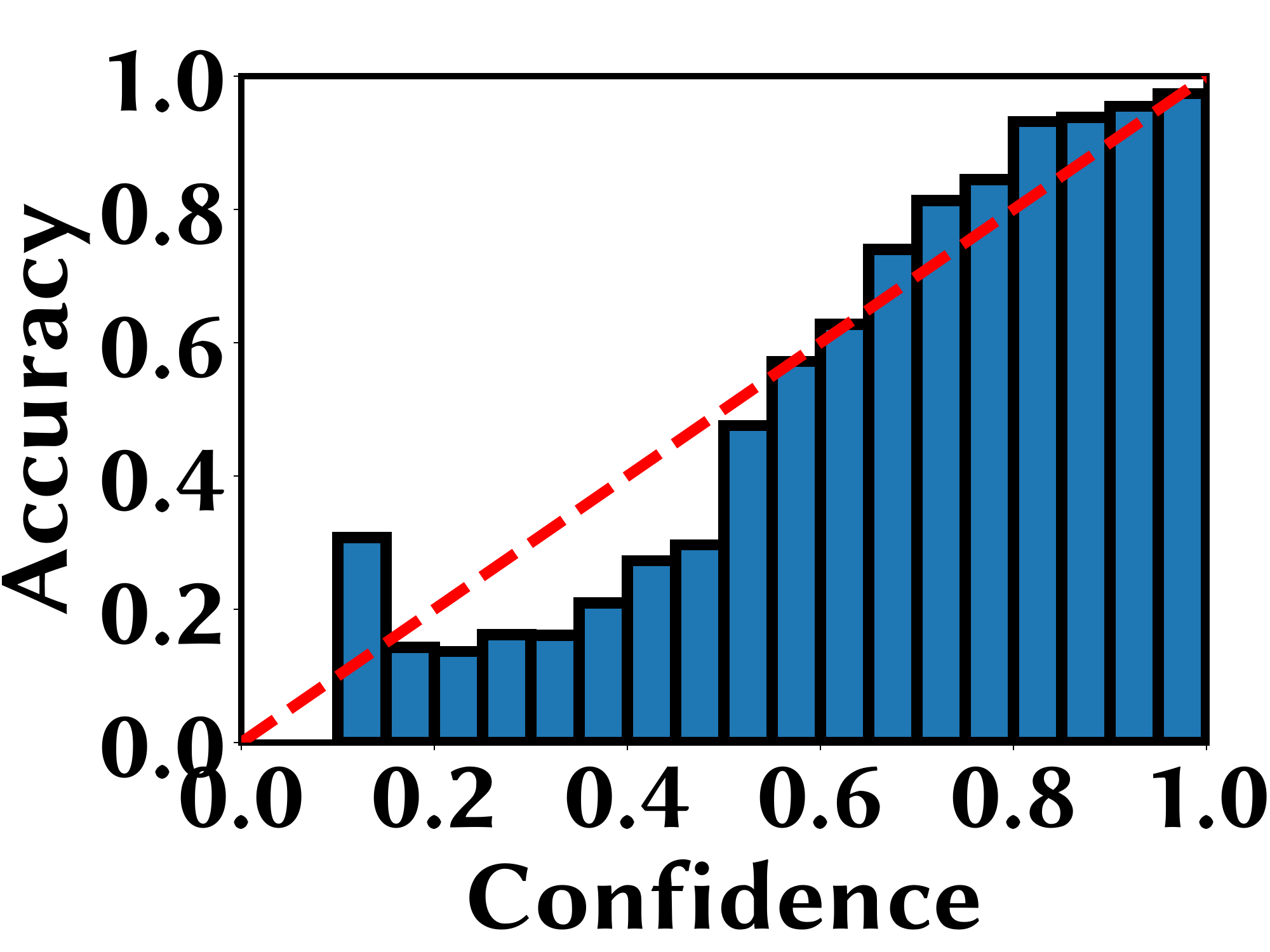}}
	\subfigure[]{\includegraphics[trim=0cm 0cm 0cm 0cm,clip,  width=0.195\textwidth]{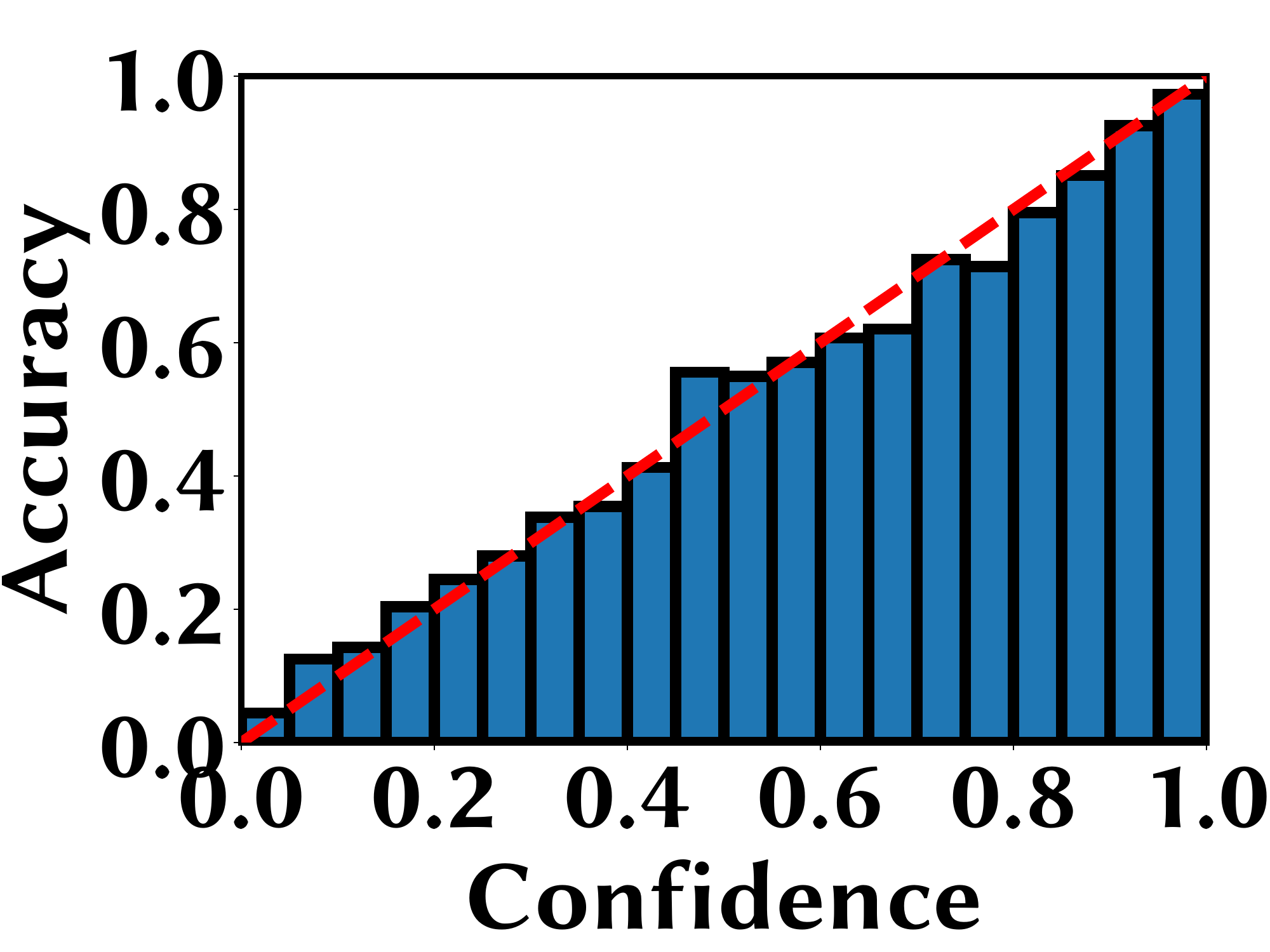}}
	\caption{Reliability diagrams of prediction confidences for VGG16 on CIFAR-10 OOD dataset. (a) Un-calibrated. (b) Temperature scaling. (c) Scaling-binning. (d) Dirichlet calibration. (e) \ouralg.}\label{fig:RD_plot_cifar10_OOD}
	\vspace{-0.3cm}
\end{figure*}

\textbf{Results. } The calibration results are presented in Table~\ref{table:cifar10_in_ECE_D1}, \ref{table:cifar10_in_ECE_D2}, \ref{table:cifar10_in_ECE_OOD} , \ref{table:cifar10_in_ECE_AD} for D1, D2, OOD and AD datasets, respectively. We highlight the best performance among all the calibration methods with bold font, including the highest AUROC/AUPR/p.9 for mis-classification detection and the lowest ECE/BS for confidence calibration.
The results show that the proposed methods (\basicModel, \advanceModel) outperform the baselines in terms of both mis-classification detection and confidence calibration, with more significant improvement on OOD and AD datasets. In addition, even though \transferMethod model is transferred to D2/OOD/AD datasets with fewer training samples than the baselines, it still offers a better calibration performance in terms of ECE and BS.

We further show the histograms of confidences for correct and wrong predictions in Fig.~\ref{fig:cifar10_OOD_histogram} on the OOD dataset. The corresponding reliability diagrams also are shown in Fig.~\ref{fig:RD_plot_cifar10_OOD} on the OOD dataset with different calibration methods. Without calibration, most confidences are high ($\sim 1.0$) for correct and wrong predictions, resulting in confidently wrong predictions and over-confident reliability diagrams.
While reducing the confidence for mis-classified samples, the baseline calibration methods also tend to decrease the confidences of correct predictions. The reliability diagrams shows over-confidence for low confidences, and slight under-confidence for high confidences. Nevertheless,  \basicModel can provide better calibration performance by separating mis-classified samples from correctly sampled ones.

\begin{table}
	\centering
	\setlength{\tabcolsep}{1.5pt}
	\parbox{.48\linewidth}{
		\centering
		\caption{ResNet-50 on CIFAR-10 D1}
		\label{table:cifar10_resnet_in_ECE_D1}
		\begin{tabular}{lccccc}
			\hline
			\textbf{Method} & \textbf{AUROC} & \textbf{AUPR} & \textbf{p.9} & \textbf{ECE} & \textbf{BS}\\
			\hline
			\textbf{MP}    & 0.834 & 0.576 & 0.430 & 16.8\% & 0.182 \\
			\textbf{TS}    & 0.833 & 0.575 & 0.427 & 4.4\% & 0.139 \\
			\textbf{SB}    & 0.843 & 0.585 & 0.438 & 4.7\% & 0.137 \\
			\textbf{Dirichlet} & 0.833 & 0.575 & 0.427 & 4.3\% & 0.139 \\
			\textbf{\basicModel} & \textbf{0.853} & \textbf{0.605} & 0.457 & 2.3\% & 0.132 \\
			\textbf{\advanceModel} & 0.853 & 0.604 & \textbf{0.458} & \textbf{1.4\%} & \textbf{0.130} \\
			\textbf{\transferMethod} & - & - & - & - & -\\
			\hline
		\end{tabular}%
		
	}
	\hfill
	\parbox{.48\linewidth}{
		\centering
		\caption{ResNet-50 on CIFAR-10 D2}
		\label{table:cifar10_resnet_in_ECE_D2}
		\begin{tabular}{lccccc}
			\hline
			\textbf{Method} & \textbf{AUROC} & \textbf{AUPR} & \textbf{p.9} & \textbf{ECE} & \textbf{BS}\\
			\hline
			\textbf{MP}    & 0.785 & 0.665 & 0.551 & 29.1\% & 0.292\\
			\textbf{TS}    & 0.784 & 0.668 & 0.546 & 3.8\% & 0.186 \\
			\textbf{SB}    & 0.792 & 0.668 & 0.560 & 3.9\% & 0.183 \\
			\textbf{Dirichlet} & 0.784 & 0.668 & 0.546 & 3.8\% & 0.186 \\
			\textbf{\basicModel} & 0.811 & \textbf{0.692} & 0.590 & 3.4\% & \textbf{0.037} \\
			\textbf{\advanceModel} & \textbf{0.813} & 0.692 & 0.589 & \textbf{2.2\%} & 0.172 \\
			\textbf{\transferMethod} & 0.809 & 0.690 & \textbf{0.593} & 3.2\% & 0.174 \\
			\hline
		\end{tabular}%
	}
\end{table}

\begin{table}
	\centering
	\setlength{\tabcolsep}{1.5pt}
	\parbox{.48\linewidth}{
		\centering
		\caption{ResNet-50 on CIFAR-10 OOD}
		\label{table:cifar10_resnet_in_ECE_OOD}
		\begin{tabular}{lccccc}
			\hline
			\textbf{Method} & \textbf{AUROC} & \textbf{AUPR} & \textbf{p.9} & \textbf{ECE} & \textbf{BS}\\
			\hline
			\textbf{MP}    & 0.883 & 0.761 & 0.548 & 25.0\% & 0.233\\
			\textbf{TS}    & 0.879 & 0.752 & 0.535 & 9.0\% & 0.138 \\
			\textbf{SB}    & 0.886 & 0.779 & 0.545 & 7.4\% & 0.127 \\
			\textbf{Dirichlet} & 0.875 & 0.746 & 0.529 & 7.6\% & 0.131 \\
			\textbf{\basicModel} & \textbf{0.913} & \textbf{0.832} & \textbf{0.615} & 1.8\% & \textbf{0.104} \\
			\textbf{\advanceModel} & 0.890 & 0.785 & 0.552 & \textbf{1.6\%} & 0.118 \\
			\textbf{\transferMethod} & 0.874 & 0.761 & 0.520 & 1.9\% & 0.126 \\
			\hline
		\end{tabular}%
		
	}
	\hfill
	\parbox{.48\linewidth}{
		\centering
		\caption{ResNet-50 on CIFAR-10 AD}
		\label{table:cifar10_resnet_in_ECE_AD}
		\begin{tabular}{lccccc}
			\hline
			\textbf{Method} & \textbf{AUROC} & \textbf{AUPR} & \textbf{p.9} & \textbf{ECE} & \textbf{BS}\\
			\hline
			\textbf{MP}    & 0.730 & 0.604 & 0.421 & 27.8\% & 0.283\\
			\textbf{TS}    & 0.724 & 0.603 & 0.412 & 6.4\% & 0.201 \\
			\textbf{SB}    & 0.752 & 0.616 & 0.452 & 4.9\% & 0.193 \\
			\textbf{Dirichlet} & 0.724 & 0.603 & 0.412 & 6.5\% & 0.202 \\
			\textbf{\ouralg} & 0.780 & 0.654 & \textbf{0.481} & 2.5\% & 0.181 \\
			\textbf{\ouralgtwo} & \textbf{0.783} & \textbf{0.663} & 0.479 & \textbf{2.0\%} & \textbf{0.179} \\
			\textbf{\ouralgthree} & 0.732 & 0.581 & 0.433 & 4.5\% & 0.199   \\
			\hline
		\end{tabular}%
		
	}
\end{table}

\begin{figure*}[!t]
	\centering
	\subfigure[]{\includegraphics[trim=0cm 0cm 0cm 0cm,clip,  width=0.195\textwidth]{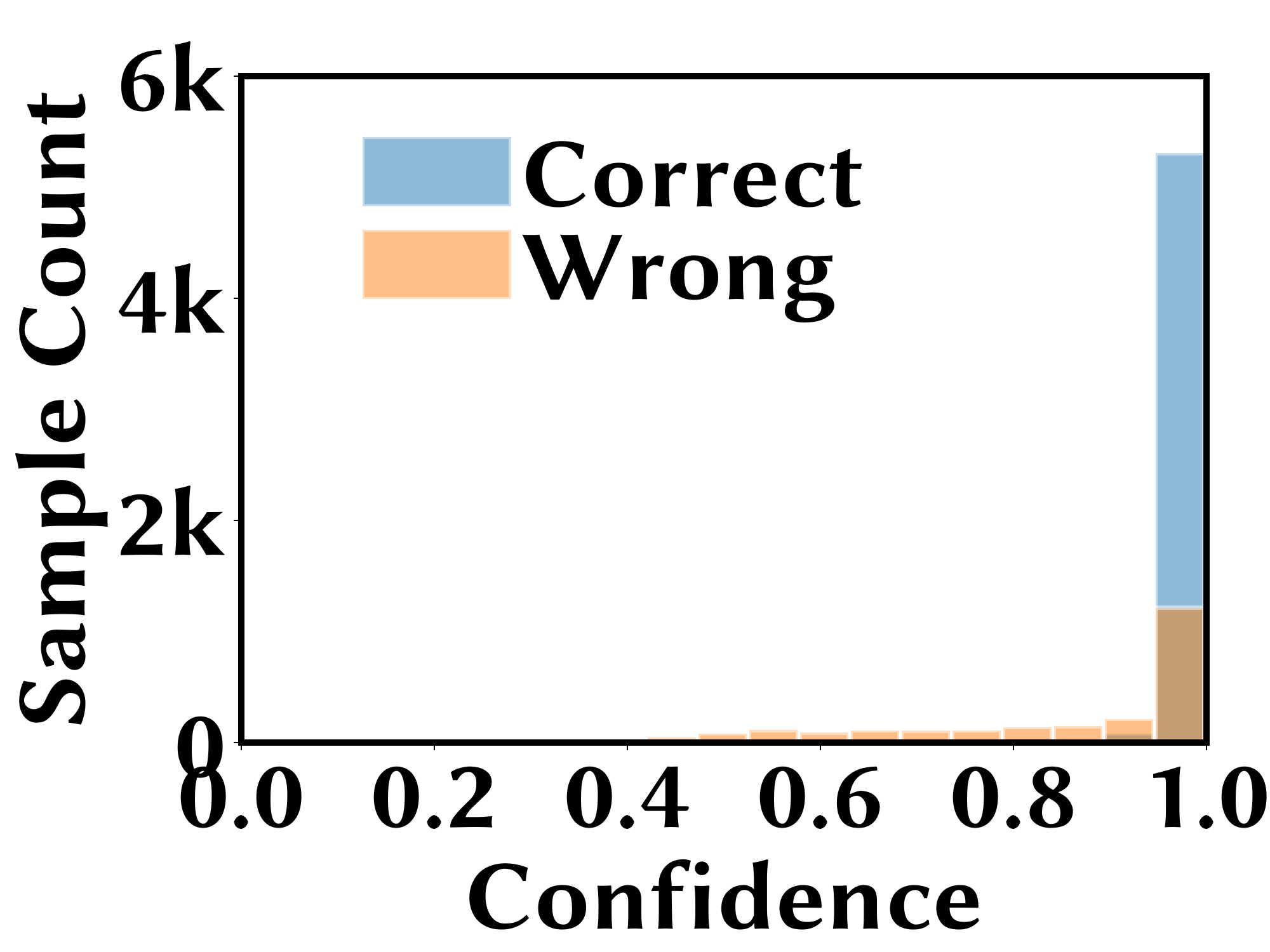}}
	\subfigure[]{\includegraphics[trim=0cm 0cm 0cm 0cm,clip,  width=0.195\textwidth]{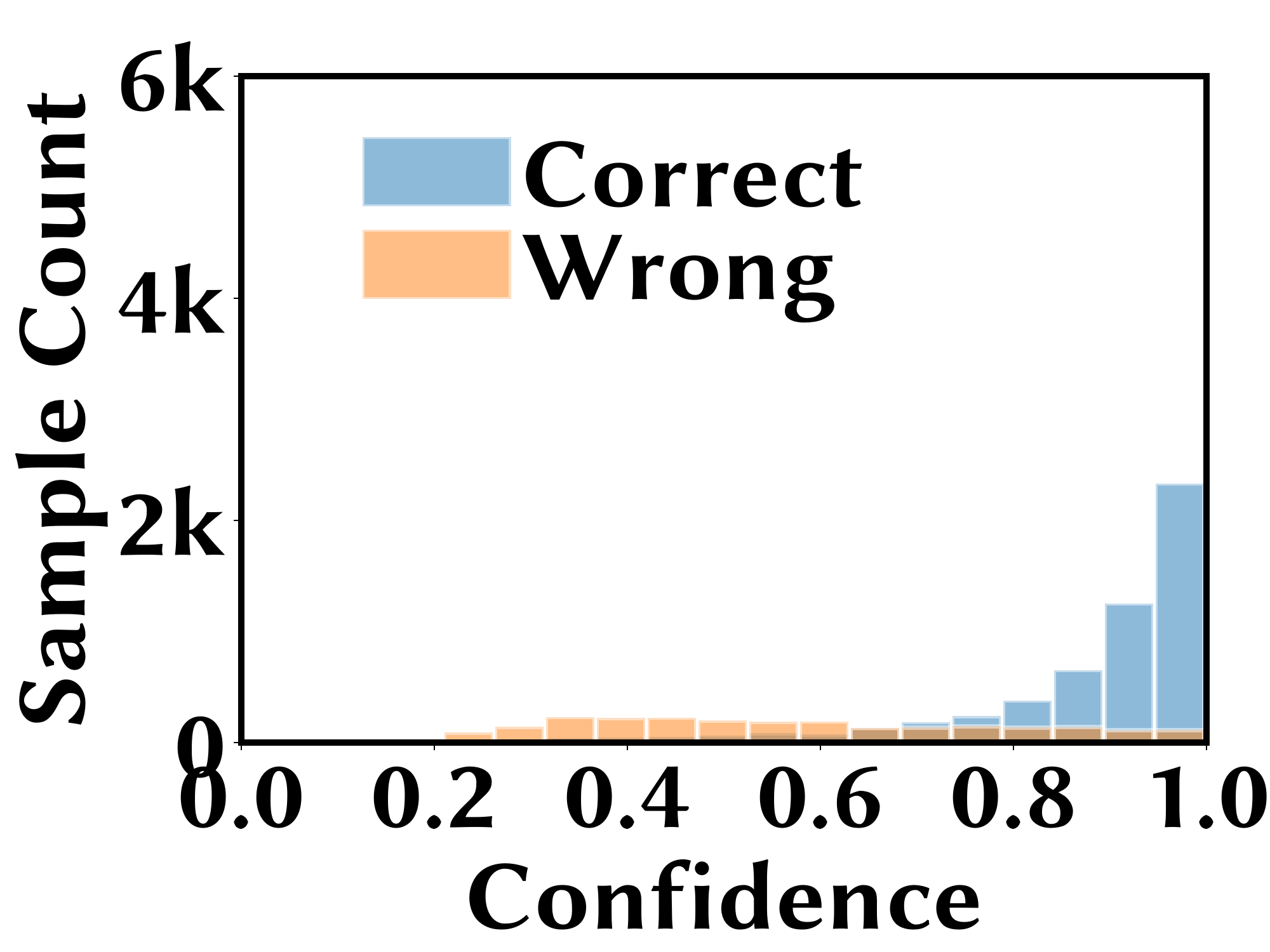}}
	\subfigure[]{\includegraphics[trim=0cm 0cm 0cm 0cm,clip,  width=0.195\textwidth]{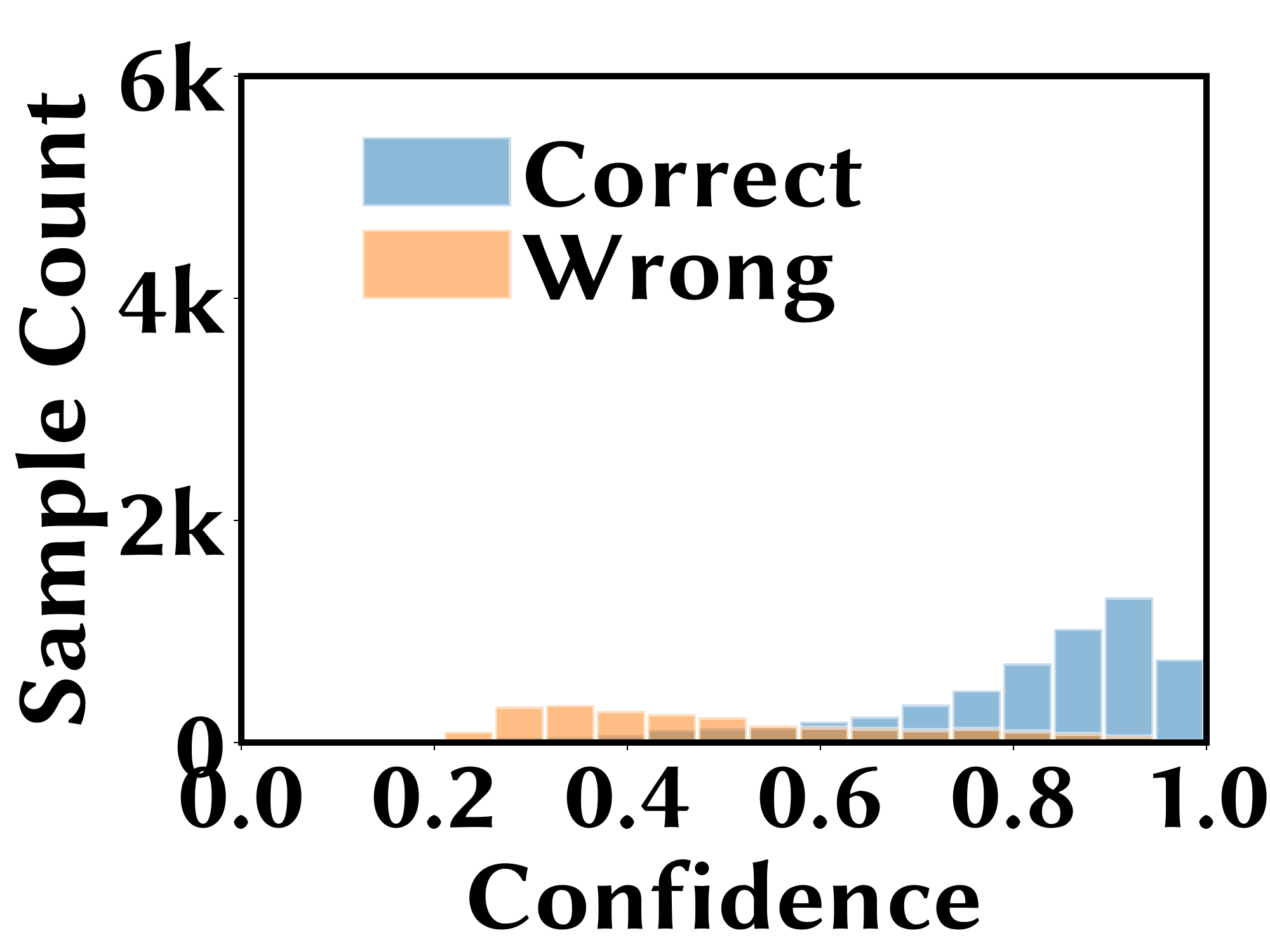}}
	\subfigure[]{\includegraphics[trim=0cm 0cm 0cm 0cm,clip,  width=0.195\textwidth]{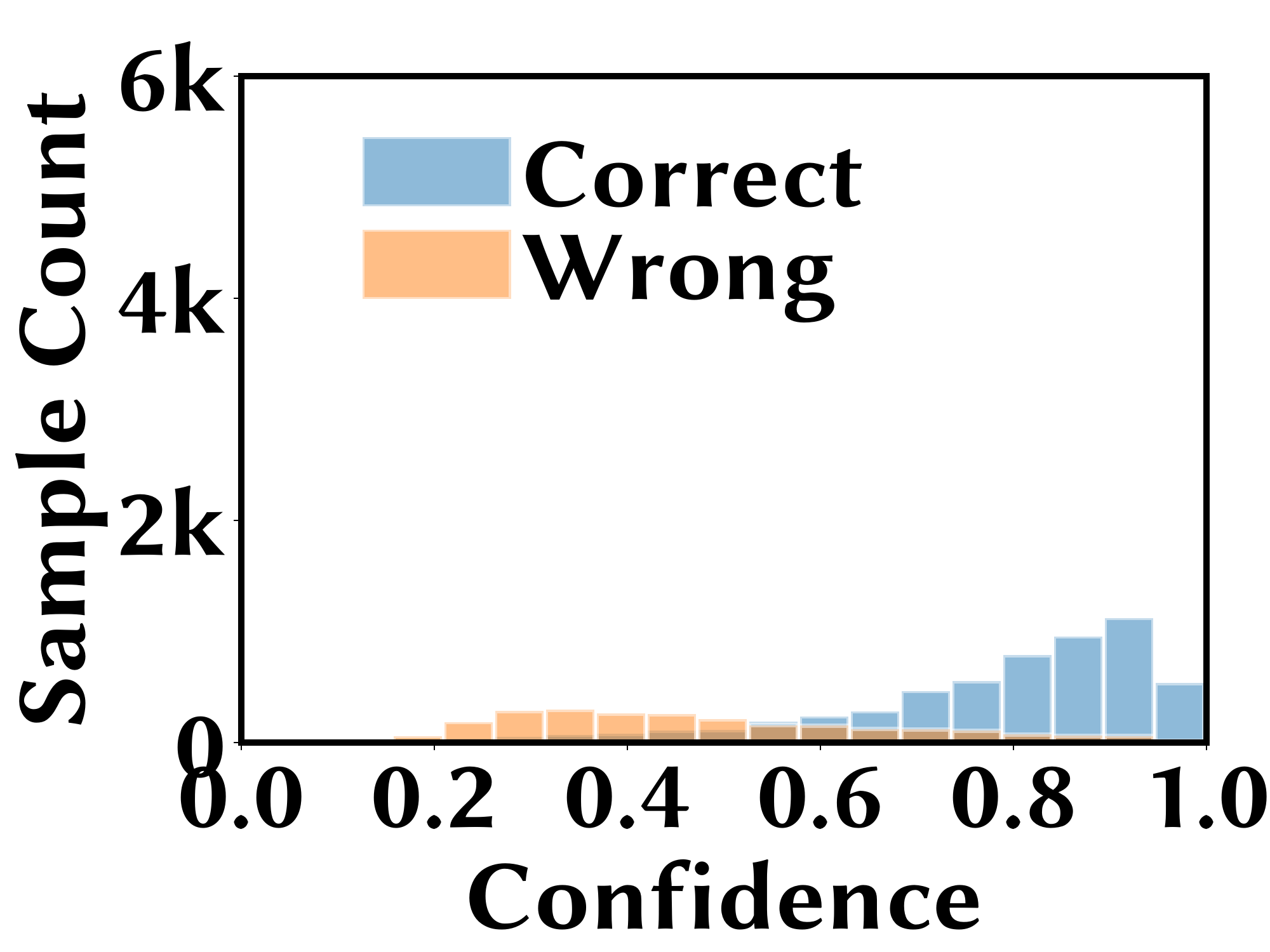}}
	\subfigure[]{\includegraphics[trim=0cm 0cm 0cm 0cm,clip,  width=0.195\textwidth]{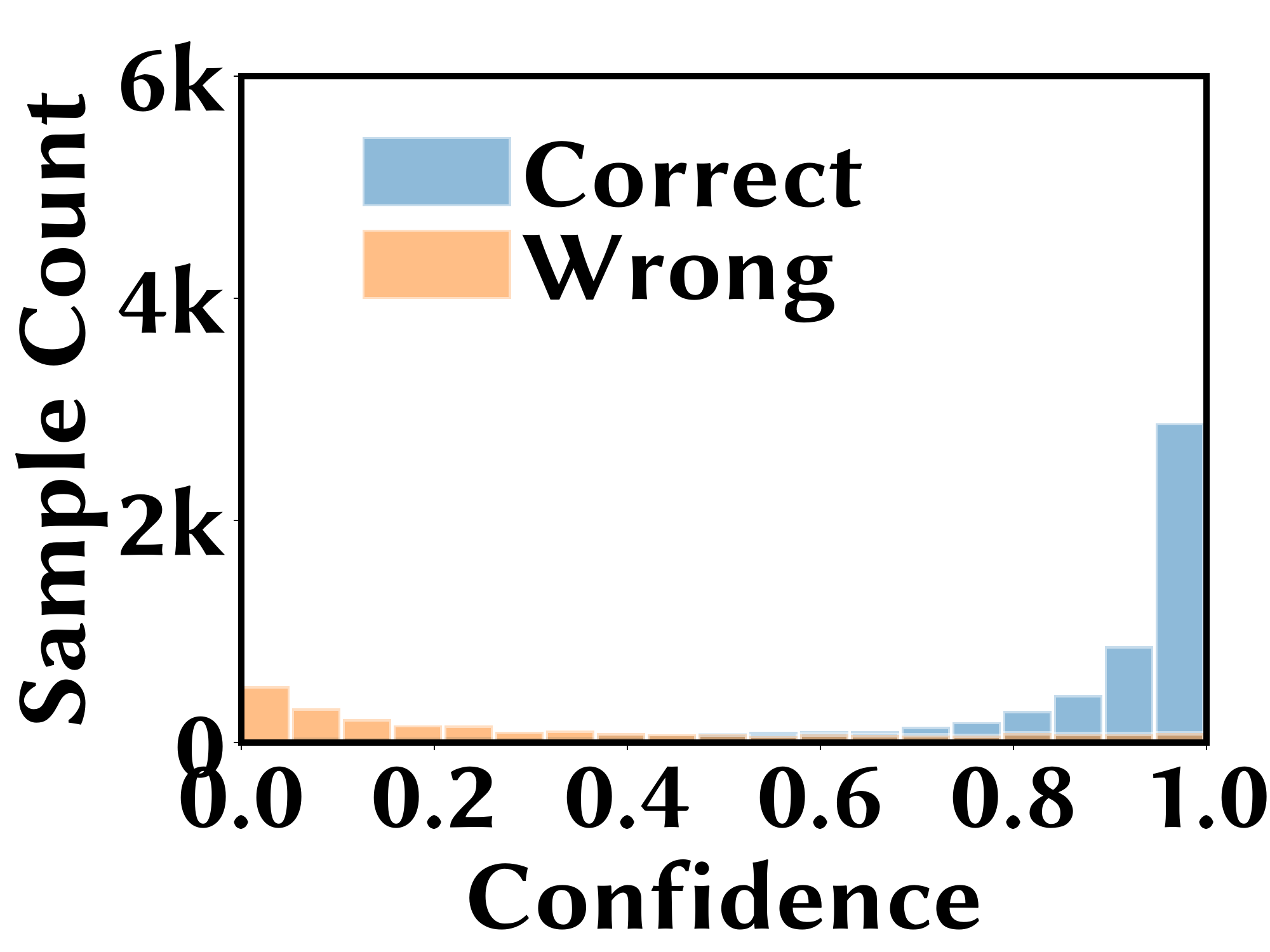}}
	\vspace{-0.3cm}
	\caption{Histograms of prediction confidences for ResNet-50 on CIFAR-10 OOD dataset. (a) Un-calibrated. (b) Temperature scaling. (c) Scaling-binning. (d) Dirichlet calibration. (e) \ouralg.}\label{fig:cifar10_OOD_resnet_histogram}
	\vspace{-0.3cm}
\end{figure*}

\begin{figure*}[!t]
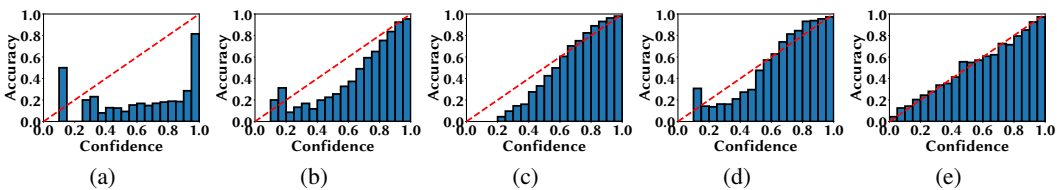

	\centering
	\subfigure[]{\includegraphics[trim=0cm 0cm 0cm 0cm,clip,  width=0.195\textwidth]{figures/cifar10_ResNet/Cifar10_OOD_MP_ECE.pdf}}
	\subfigure[]{\includegraphics[trim=0cm 0cm 0cm 0cm,clip,  width=0.195\textwidth]{figures/cifar10_ResNet/Cifar10_OOD_TS_ECE.pdf}}
	\subfigure[]{\includegraphics[trim=0cm 0cm 0cm 0cm,clip,  width=0.195\textwidth]{figures/cifar10_ResNet/Cifar10_OOD_SB_ECE.pdf}}
	\subfigure[]{\includegraphics[trim=0cm 0cm 0cm 0cm,clip,  width=0.195\textwidth]{figures/cifar10_ResNet/Cifar10_OOD_Dirichlet_ECE.pdf}}
	\subfigure[]{\includegraphics[trim=0cm 0cm 0cm 0cm,clip,  width=0.195\textwidth]{figures/cifar10_ResNet/Cifar10_OOD_Calib1_ECE.pdf}}
	\caption{Reliability diagrams of prediction confidences for ResNet-50 on CIFAR-10 OOD dataset. (a) Un-calibrated. (b) Temperature scaling. (c) Scaling-binning. (d) Dirichlet calibration. (e) \ouralg.}\label{fig:RD_plot_cifar10_resnet_OOD}
	\vspace{-0.3cm}
\end{figure*}

\subsubsection{10-class Image Classification with ResNet-50 DNN}

\textbf{Target DNN model. } The target DNN model is a modified ResNet-50 for tiny images (CIFAR). The target model includes batch normalization layers and dropout layers to improve classification performance. It is trained in \textit{PyTorch} on CIFAR-10 training dataset (50k images). The pre-trained weights of target DNN model are downloaded from \cite{resnet_weights_github_2018}.

All the other settings, such as the datasets and architecture
of the neural network in \ouralg/\ouralgtwo, are the same as those for the 10-class VGG16 DNN
described in Section~\ref{sec:10class_vgg}. The inference accuracies on the four test datasets are 75\% (D1), 59\% (D2), 70\% (OOD), and 63\% (AD).

\textbf{Results. } The calibration results are presented in Table~\ref{table:cifar10_resnet_in_ECE_D1}, \ref{table:cifar10_resnet_in_ECE_D2}, \ref{table:cifar10_resnet_in_ECE_OOD}, \ref{table:cifar10_resnet_in_ECE_AD} for D1, D2, OOD and AD datasets, respectively.

The results show that the proposed methods (\basicModel, \advanceModel) outperform the baselines in terms of both mis-classification detection and confidence calibration, with significant improvement on OOD and AD datasets.
We further show the histograms of confidences for correct and wrong predictions in Fig.~\ref{fig:cifar10_OOD_resnet_histogram} on the OOD dataset. The corresponding reliability diagrams also are shown in Fig.~\ref{fig:RD_plot_cifar10_resnet_OOD} on the OOD dataset with different calibration methods.

\subsubsection{100-class Image Classification with VGG16 DNN}\label{sec:100class_vgg}

\textbf{Target DNN model. } The target DNN model is a modified VGG16 for tiny images (CIFAR). The target model includes batch normalization layers and dropout layers to improve classification performance. It is trained in \textit{Tensorflow} on CIFAR-100 training dataset (50k images). The pre-trained weights of target DNN model is downloaded from \cite{vgg16_weights_github_2018}.

\textbf{Datasets.} We evaluate the calibration performance on four datasets generated from CIFAR-100 and CIFAR-10 \cite{cifar10_100_dataset_2009}, including two augmented datasets (D1 and D2), one out-of-distribution dataset (OOD) and one adversarial dataset (AD). We first randomly select 30k samples from CIFAR-100 training dataset and 10k samples from CIFAR-100 testing dataset, and then perform augmentation operations on the selected samples. For D1, the augment operations and parameters include rotation within $[-20,20]$ degrees, vertical/horizontal shift within $[-0.25,0.25]$, zoom with $0.4$, and horizontal flip. For D2, the augmentation parameters are: rotation within $[-40,40]$ degrees, vertical/horizontal shift within $[-0.4,0.4]$, zoom with $0.5$, and horizontal flip. All augmentations are performed via \textit{ImageDataGenerator} function in \textit{Tensorflow}. The OOD dataset also includes 40k samples, with 12k OOD samples randomly selected from CIFAR-10 training dataset and 28k samples from CIFAR-100 training dataset. As for 12k OOD samples from CIFAF-10, the true labels are mapped to CIFAR-100's class labels. Specifically, we consider OOD samples with label ``truck'' as ``pick-up truck'' in CIFAR-100, and OOD samples with label ``bus'' as ``automobile'' in CIFAR-100. The other OOD samples are considered with ``NULL'' labels, indicating not belonging to any of the 100 classes in CIFAR-100. For the AD dataset, the 40k samples include 20k normal samples and  20k adversarial samples. The 20k normal samples contain 10k images from the CIFAR-100 training dataset and 10k images from the CIFAR-100 testing dataset, while
the 20k adversarial samples are generated based on 20k randomly selected
samples from the CIFAR-100 training dataset with DeepFool-Attack using \textit{foolbox} package \cite{adversarial_toolbox_foolbox_metthias_ICML_2017}. The inference accuracies on the four test datasets are 65\% (D1), 47\% (D2), 77\% (OOD), and 54\% (AD). The 40k samples (in D1, D2, OOD or AD) are randomly split into a 30k dataset for training, a 2k validation dataset for hyperparameter tuning, and a 8k testing dataset for performance evaluation.

\textbf{Baselines. } For each dataset, the baselines of TS, SB, and Dirichlet calibration are trained on the respective 30k training dataset.
For Dirichlet calibration, the regularization hyperparameters are tuned with a minimal ECE on the 2k validation dataset. The 30k training and 2k validation dataset are the same as those used for
training \ouralg and \ouralgtwo.

\textbf{Our method. } The neural network of \ouralg is implemented with one hidden layers with 100 hidden nodes.
The input layer contains 100 nodes and the output layer contains 101 nodes (including 100 classes and one ``mis-classification'' class).  \ouralg is trained over 1000 epochs using
the Adam optimizer (learning rate $10^{-3}$). The loss function hyperparameters ($\lambda_1$ and $\lambda_2$) and the confidence calculation method are selected with a minimal ECE on the validation dataset. For \ouralg, the neural network is implemented with the same structure as \basicModel, but the output layer contains only 1 node, representing the ``mis-classification'' class. The temperature scaling layer is implemented with a self-defined customer Keras layer with one learnable weight $T$. The training settings of \ouralgtwo are the same as those of \basicModel. For the transferred model \transferMethod, we first pre-train \advanceModel on dataset D1. Then, when applied to another target dataset, a set of 320 samples are randomly selected from the 30k training dataset for 30k training dataset . In addition, we select another 200 samples from the 2k validation dataset  to tune corresponding hyperparameters and confidence calculation method in \ouralgthree.

\begin{figure*}[!t]
	\centering
	\subfigure[]{\includegraphics[trim=0cm 0cm 0cm 0cm,clip,  width=0.195\textwidth]{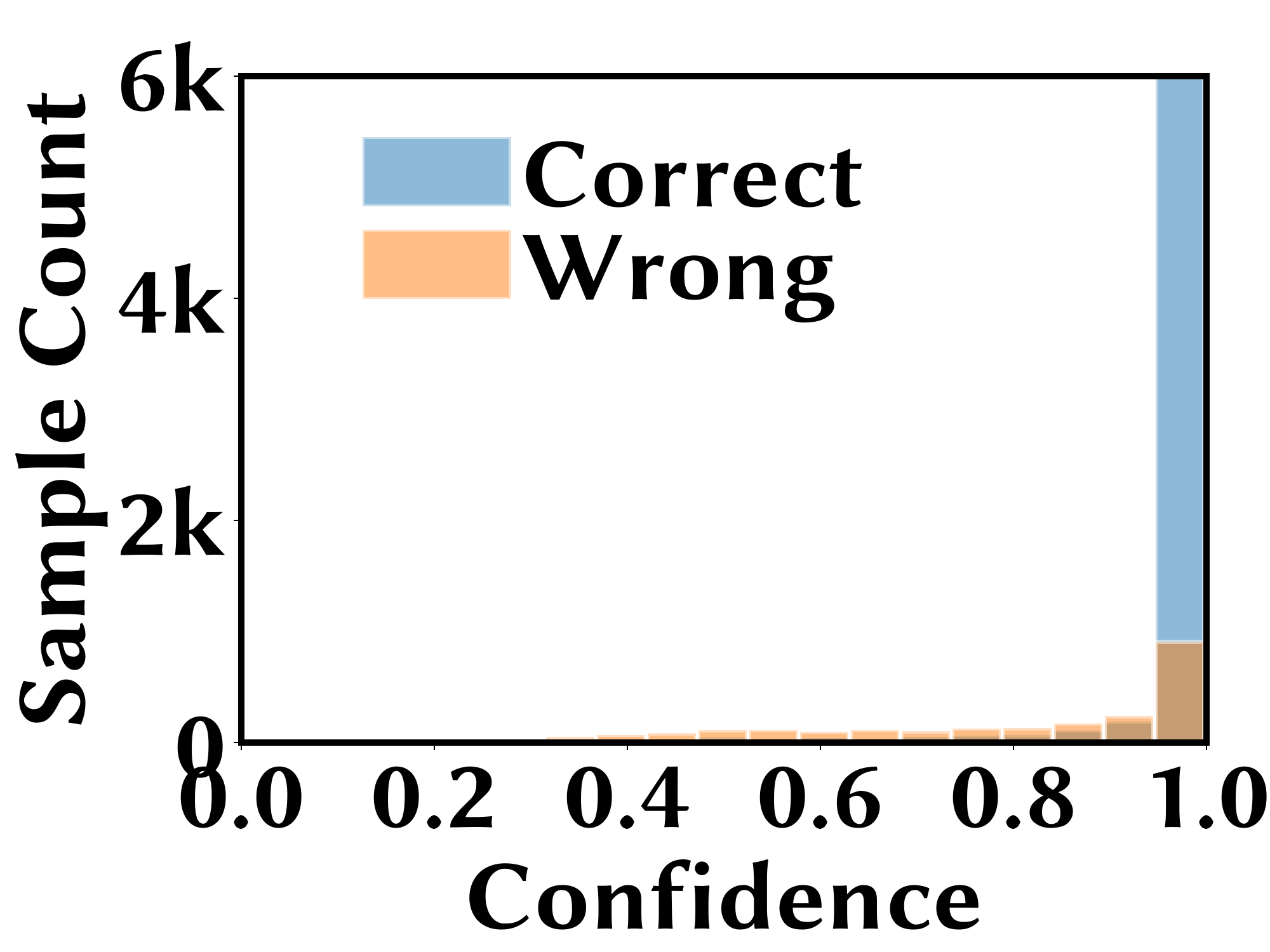}}
	\subfigure[]{\includegraphics[trim=0cm 0cm 0cm 0cm,clip,  width=0.195\textwidth]{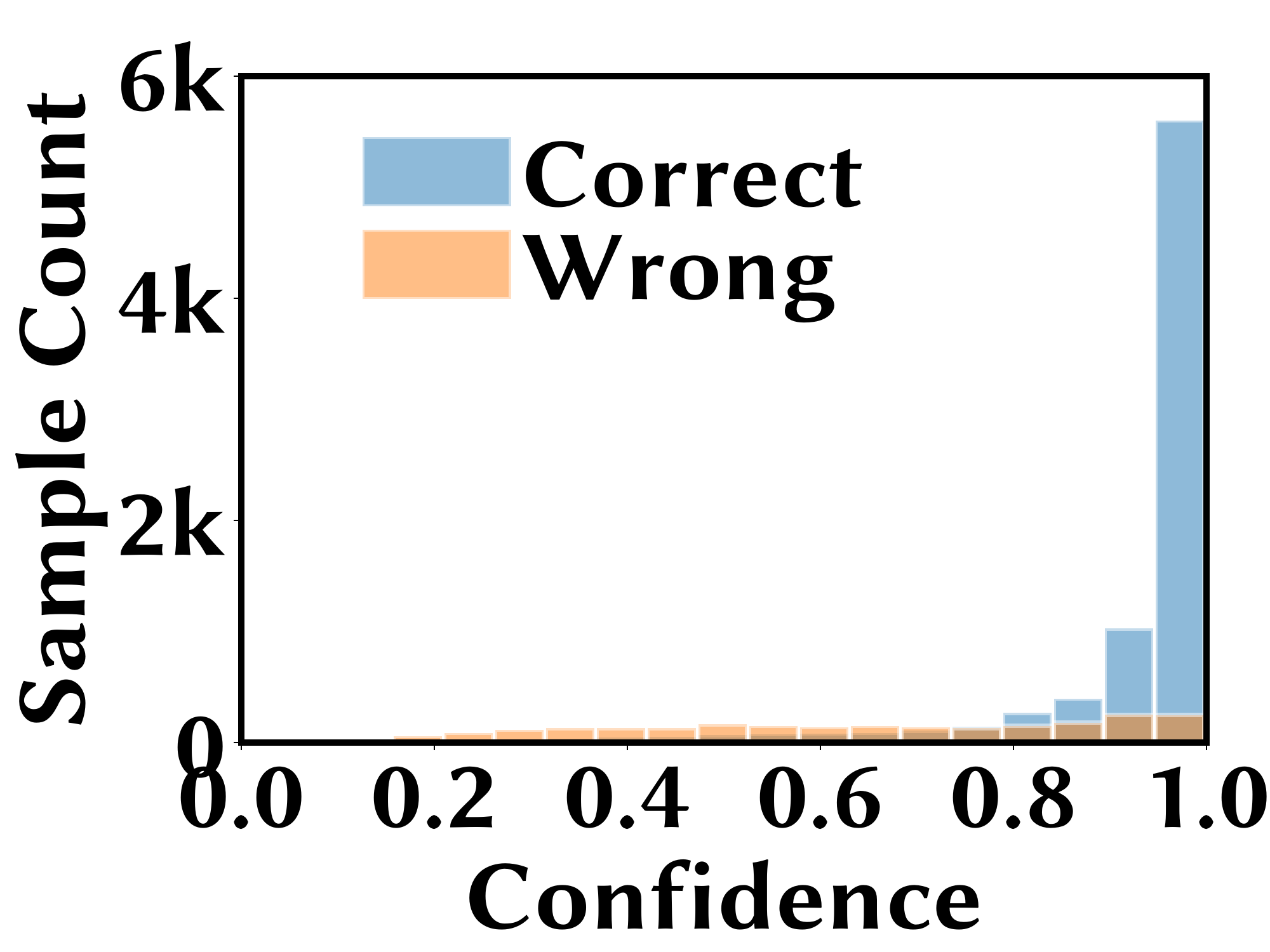}}
	\subfigure[]{\includegraphics[trim=0cm 0cm 0cm 0cm,clip,  width=0.195\textwidth]{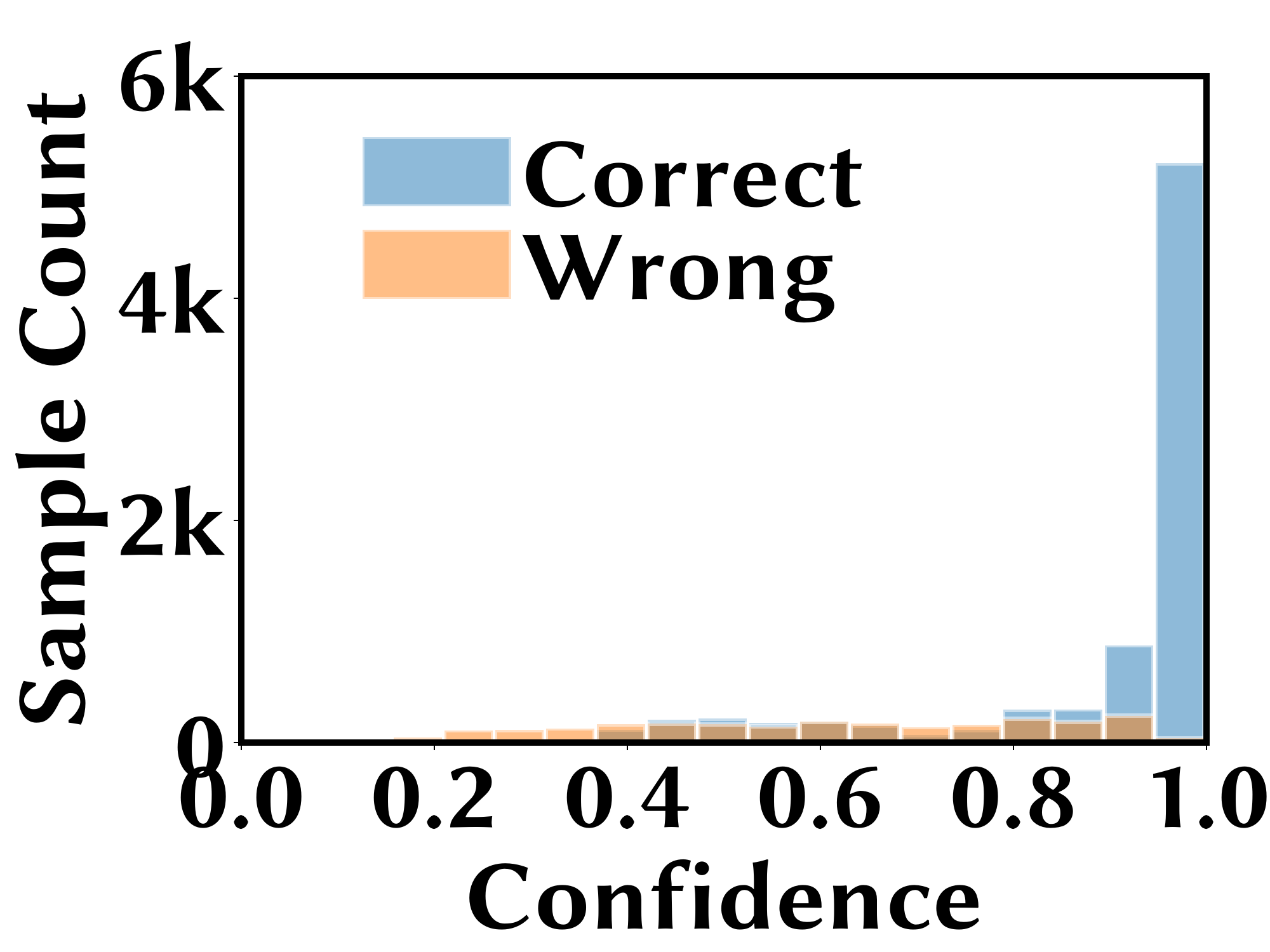}}
	\subfigure[]{\includegraphics[trim=0cm 0cm 0cm 0cm,clip,  width=0.195\textwidth]{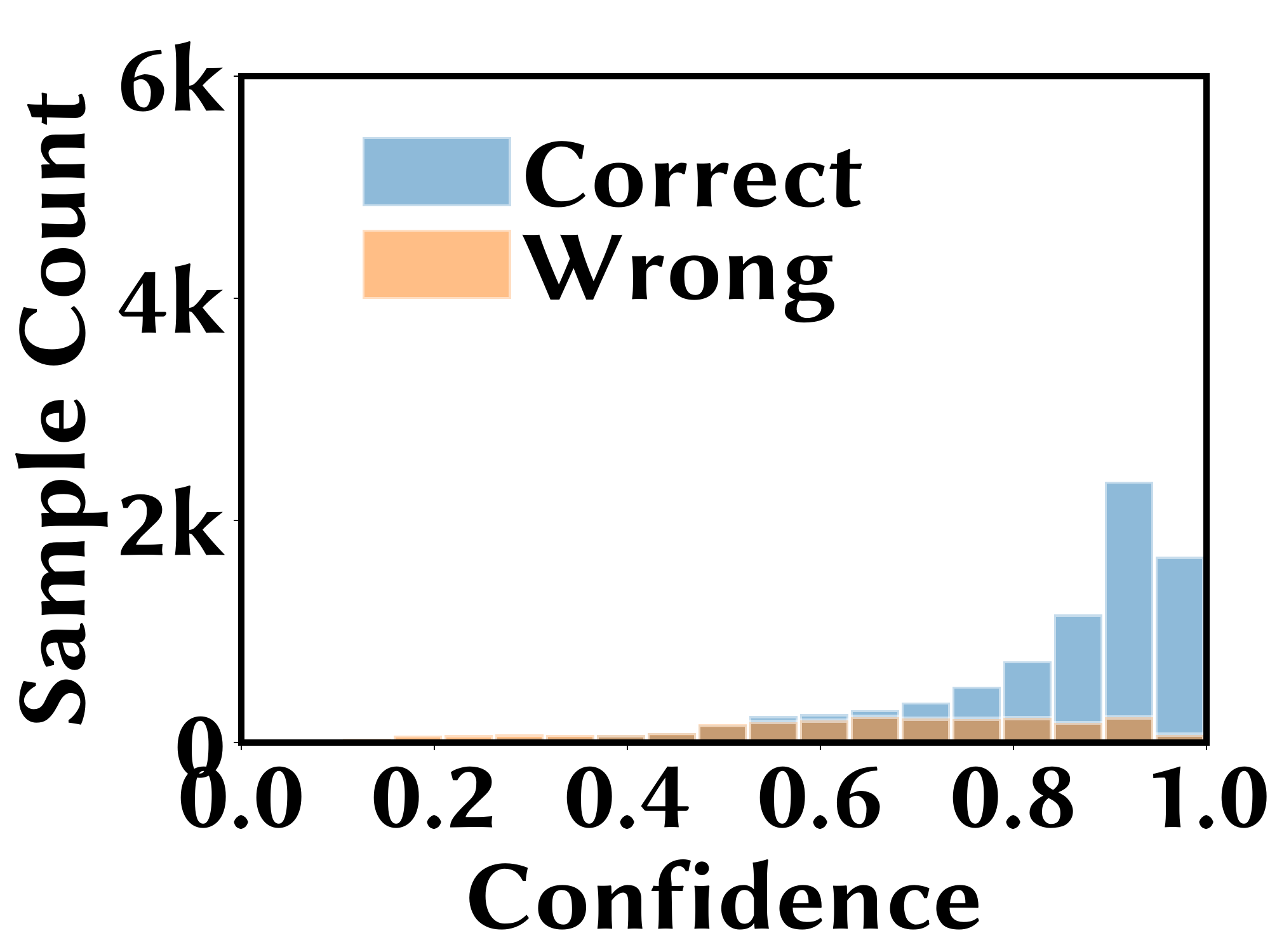}}
	\subfigure[]{\includegraphics[trim=0cm 0cm 0cm 0cm,clip,  width=0.195\textwidth]{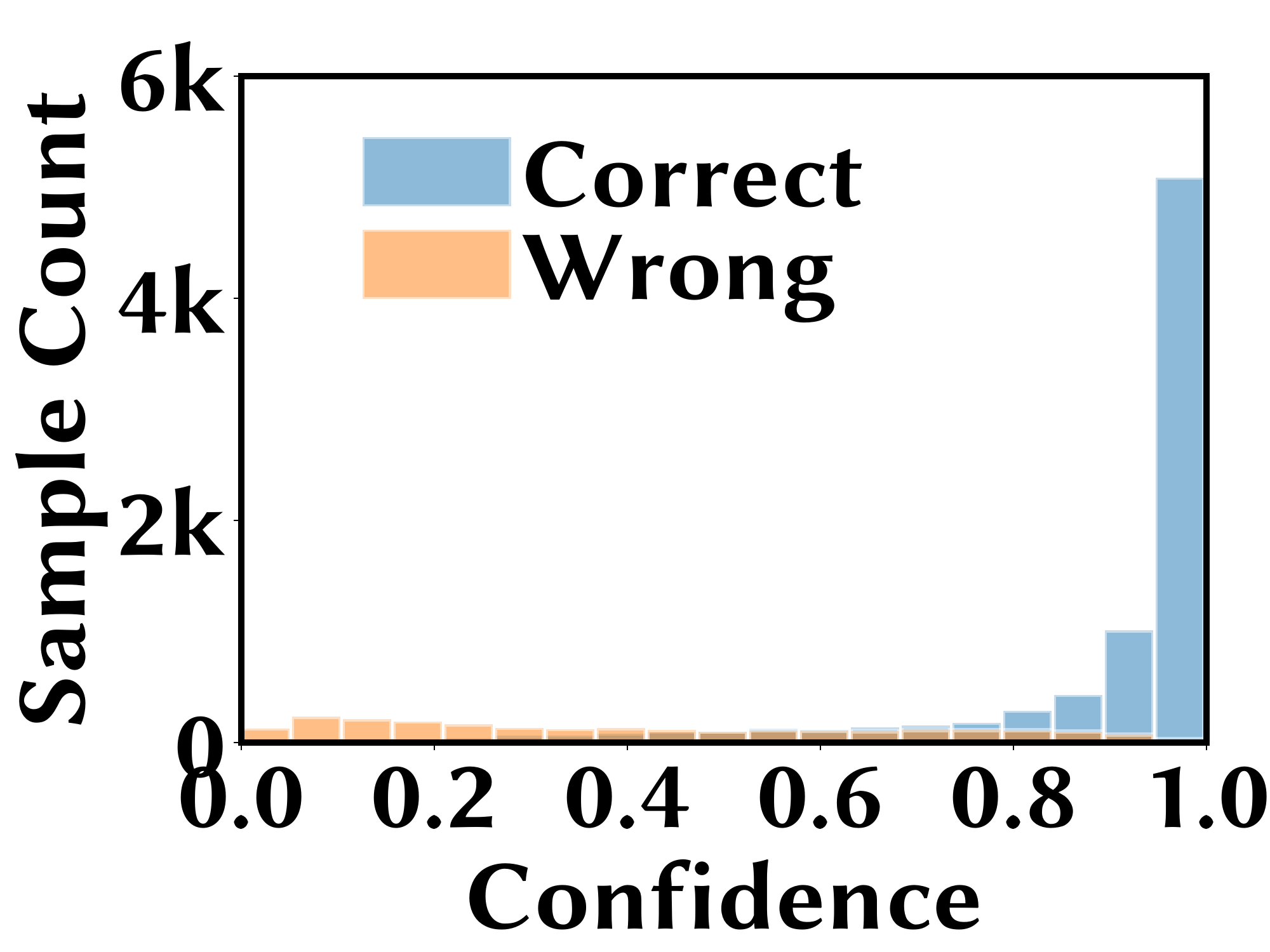}}
	\vspace{-0.3cm}
	\caption{Histograms of prediction confidences for VGG16 on CIFAR-100 OOD dataset. (a) Un-calibrated. (b) Temperature scaling. (c) Scaling-binning. (d) Dirichlet calibration. (e) \ouralg. }\label{fig:cifar100_OOD_histogram}
	\vspace{-0.3cm}
\end{figure*}

\begin{figure*}[!t]
	\centering
	\subfigure[]{\includegraphics[trim=0cm 0cm 0cm 0cm,clip,  width=0.195\textwidth]{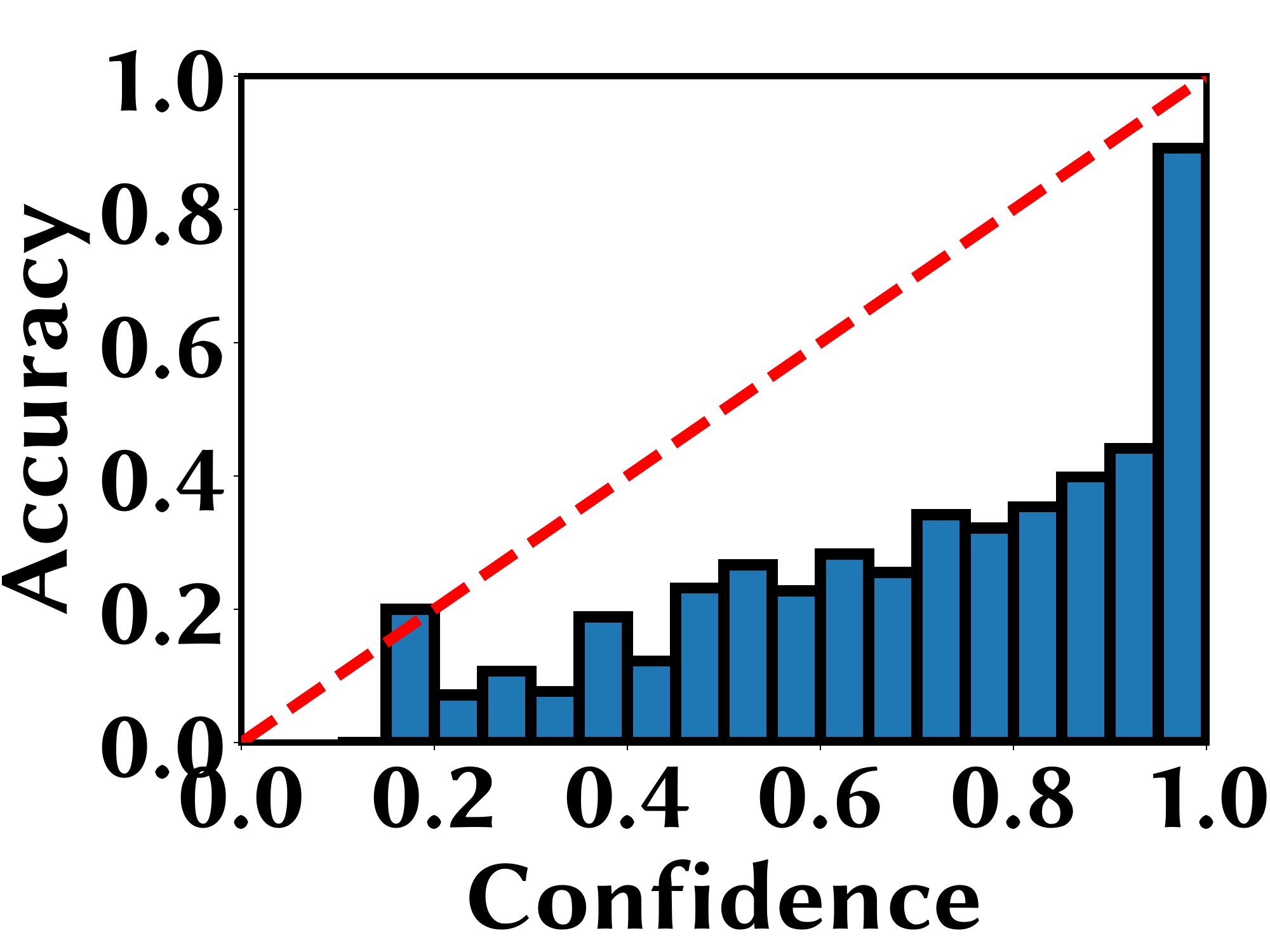}}
	\subfigure[]{\includegraphics[trim=0cm 0cm 0cm 0cm,clip,  width=0.195\textwidth]{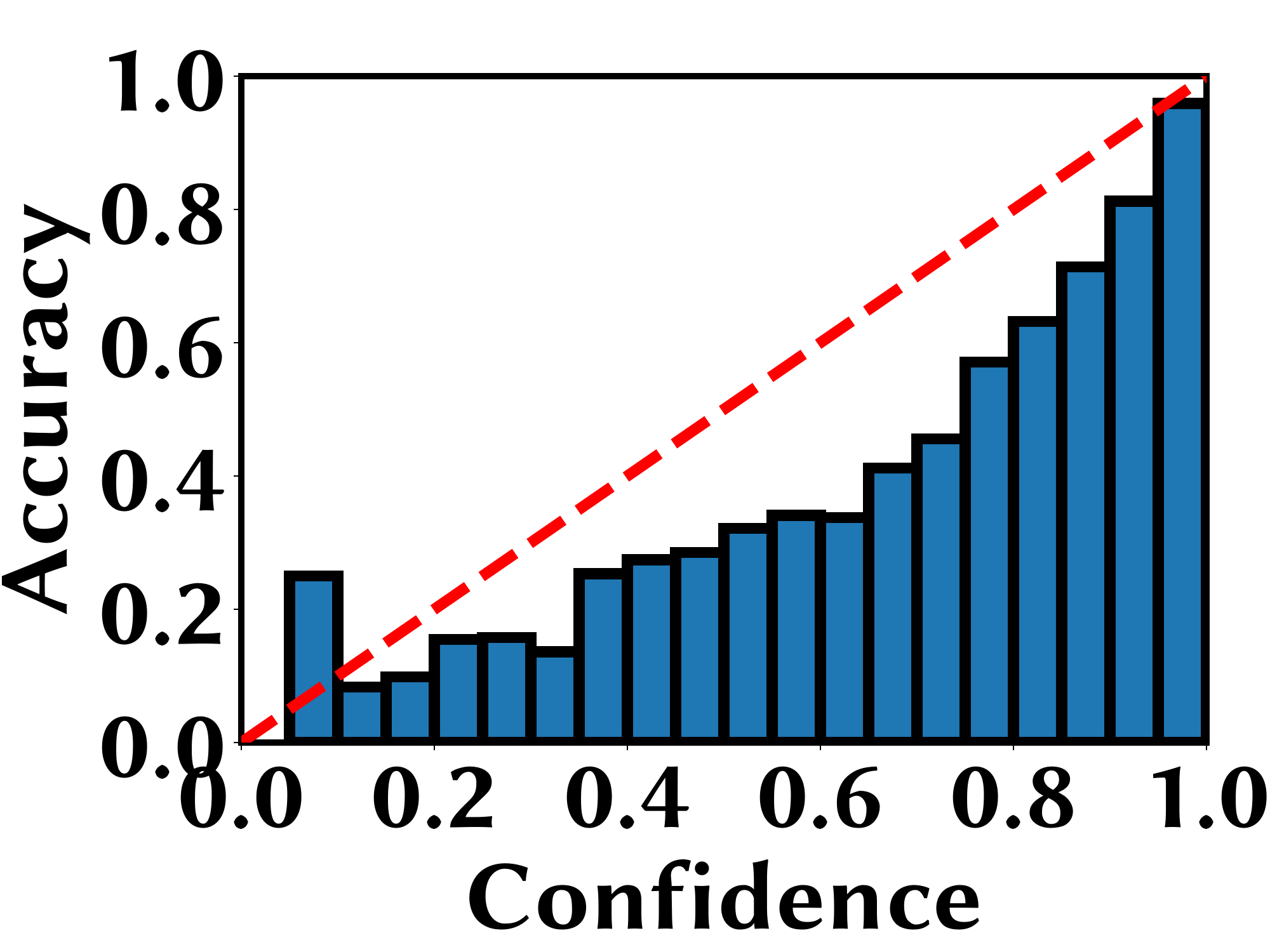}}
	\subfigure[]{\includegraphics[trim=0cm 0cm 0cm 0cm,clip,  width=0.195\textwidth]{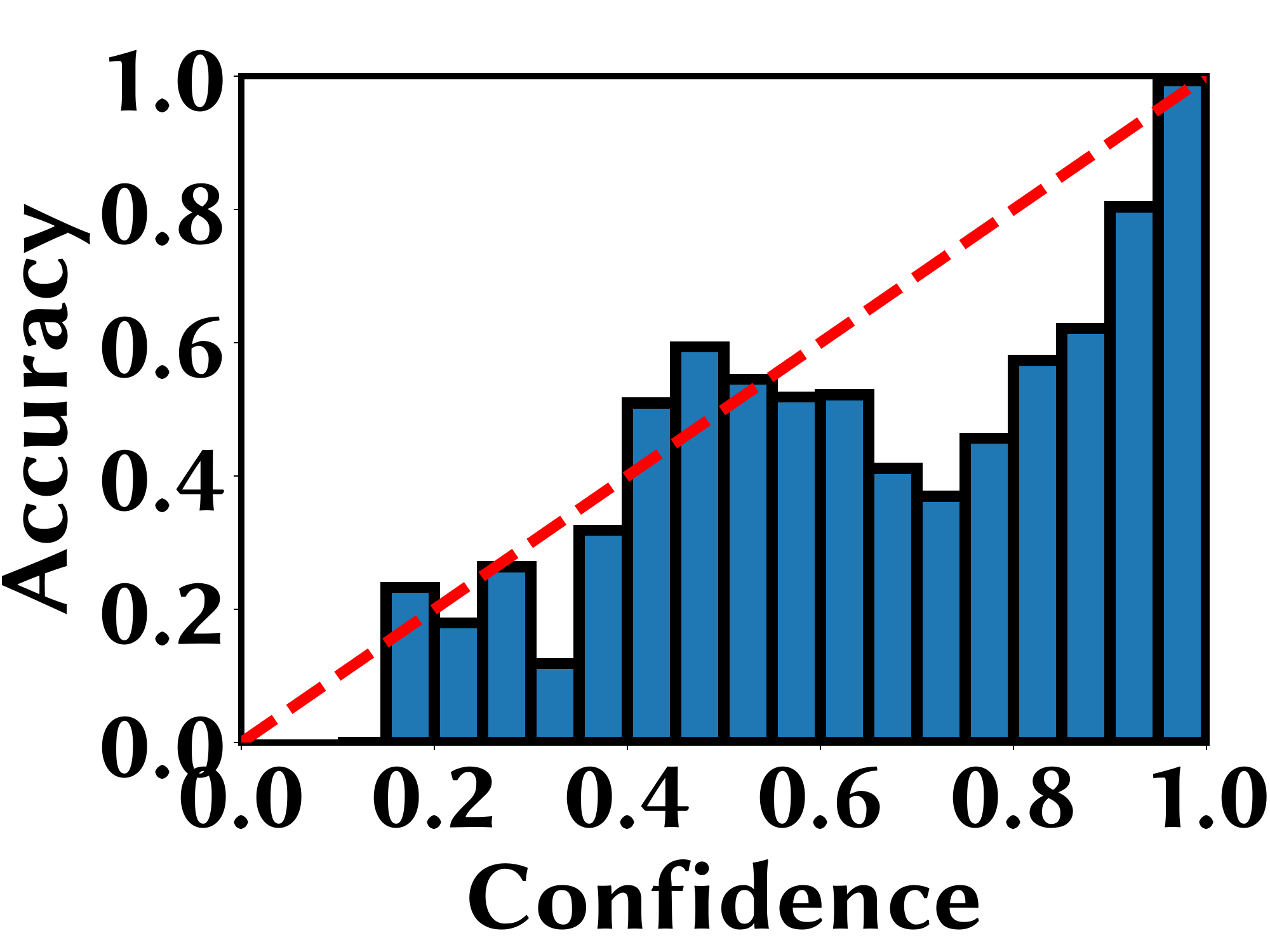}}
	\subfigure[]{\includegraphics[trim=0cm 0cm 0cm 0cm,clip,  width=0.195\textwidth]{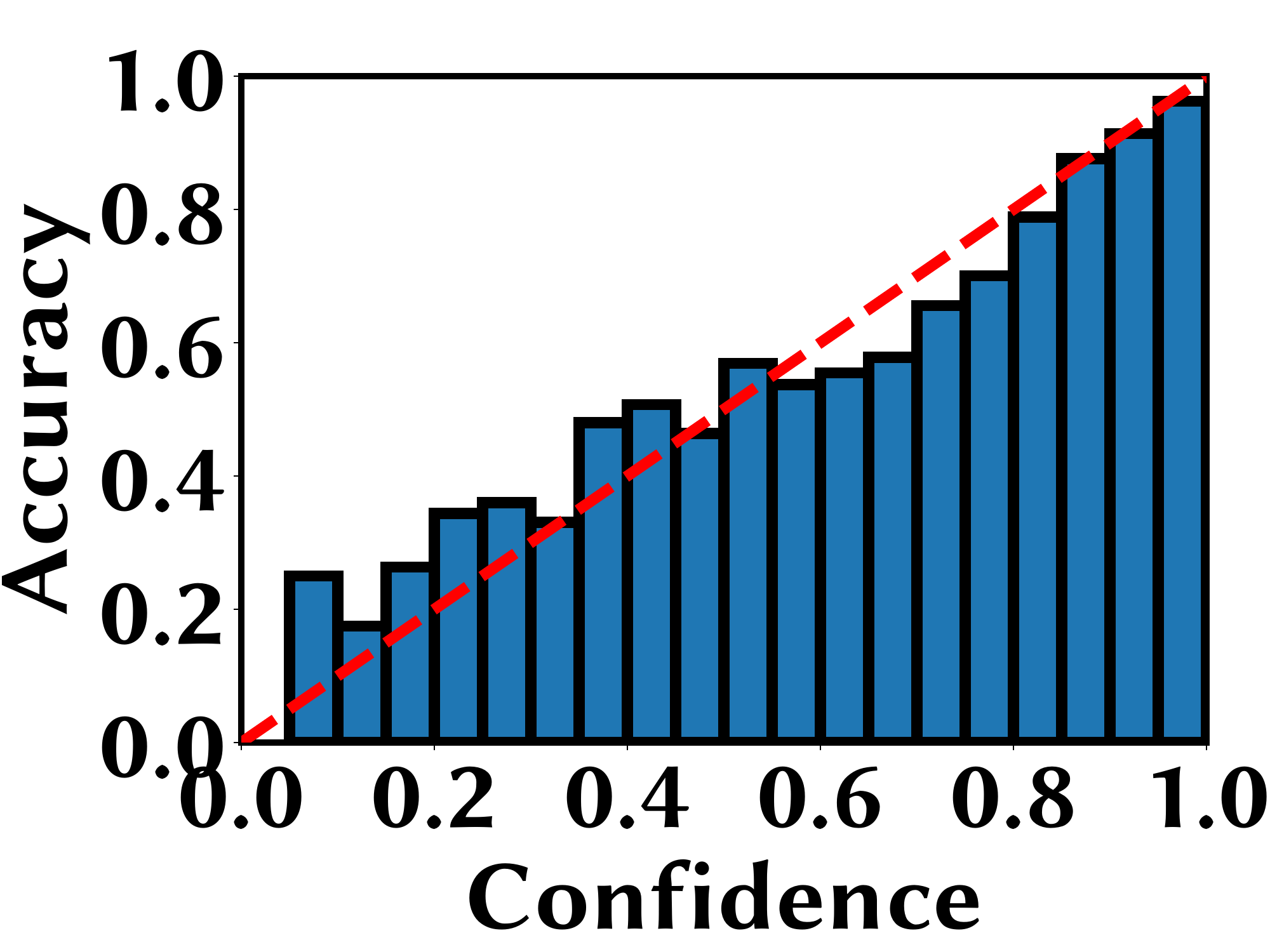}}
	\subfigure[]{\includegraphics[trim=0cm 0cm 0cm 0cm,clip,  width=0.195\textwidth]{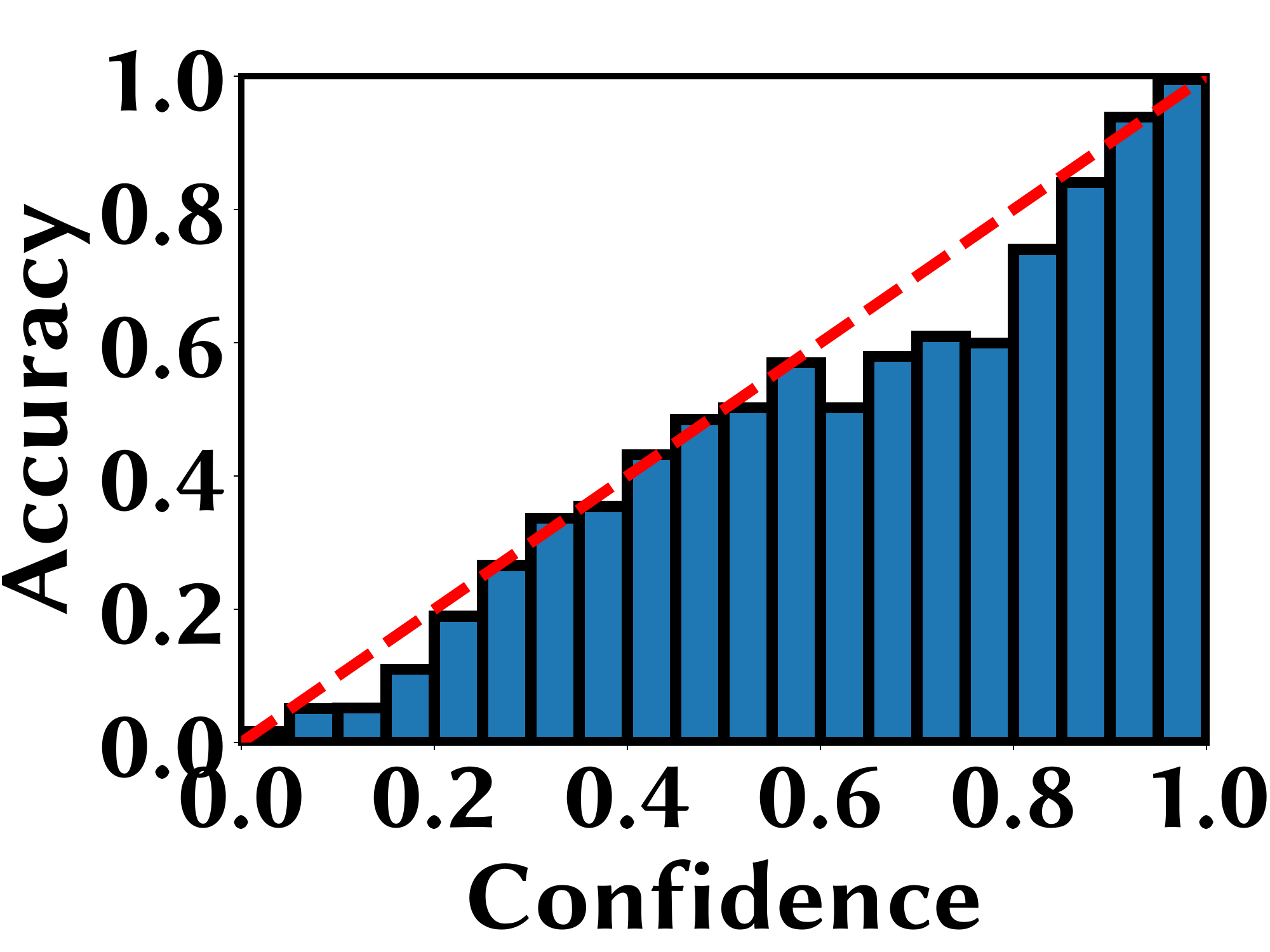}}
	\vspace{-0.3cm}
	\caption{Reliability diagrams of prediction confidences for VGG16 on CIFAR-100 OOD dataset. (a) Un-calibrated. (b) Temperature scaling. (c) Scaling-binning. (d) Dirichlet calibration. (e) \ouralg.}\label{fig:RD_plot_cifar100_OOD}
	\vspace{-0.3cm}
\end{figure*}

\textbf{Results. } The calibration results are presented in Table~\ref{table:cifar100_in_ECE_D1}, \ref{table:cifar100_in_ECE_D2}, \ref{table:cifar100_in_ECE_OOD} , \ref{table:cifar100_in_ECE_AD} for D1, D2, OOD and AD datasets, respectively.
The results show that the proposed method (\basicModel) outperforms the baselines in terms of mis-classification detection. For confidence calibration, the proposed methods provide the best calibrated confidence (lowest ECE) on D2, OOD and AD, with significant improvement on OOD and AD datasets. For D1, our proposed method achieves a well-calibrated confidence comparable to the best calibration. In addition, even though \transferMethod model is transferred to D2/OOD/AD datasets with fewer training samples than the baselines, it still offers a better calibration performance in terms of ECE.
We further show the histograms of confidences for correct and wrong predictions in Fig.~\ref{fig:cifar100_OOD_histogram} on OOD dataset. The corresponding reliability diagrams also are shown in Fig.~\ref{fig:RD_plot_cifar100_OOD} on OOD dataset with different calibration methods.

\begin{table}
	\centering
	\setlength{\tabcolsep}{1.5pt}
	\parbox{.48\linewidth}{
		\centering
		\caption{VGG16 on CIFAR-100 D1}
		\label{table:cifar100_in_ECE_D1}
		\begin{tabular}{lccccc}
			\hline
			\textbf{Method} & \textbf{AUROC} & \textbf{AUPR}  & \textbf{p.9}   & \textbf{ECE}   & \textbf{BS} \\
			\hline
			\textbf{MP}    & 0.841 & 0.708 & 0.538 & 19.6\% & 0.206\\
			\textbf{TS}    & 0.846 & 0.720 & 0.550 & 2.3\% & 0.151 \\
			\textbf{SB}    & 0.878 & 0.777 & 0.592 & \textbf{2.0\%} & 0.135 \\
			\textbf{Dirichlet} & 0.861 & 0.737 & 0.574 & 3.5\% & 0.144 \\
			\textbf{\basicModel} & \textbf{0.880} & \textbf{0.787} & \textbf{0.596} & 2.4\% & \textbf{0.134} \\
			\textbf{\advanceModel} & 0.862 & 0.756 & 0.566 & 2.1\% & 0.136\\
			\textbf{\transferMethod} &-& -&-&-&-\\
			\hline
		\end{tabular}%
		
	}
	\hfill
	\parbox{.48\linewidth}{
		\centering
		\caption{VGG16 on CIFAR-100 D2}
		\label{table:cifar100_in_ECE_D2}
		\begin{tabular}{lccccc}
			\hline
			\textbf{Method} & \textbf{AUROC} & \textbf{AUPR}  & \textbf{p.9}   & \textbf{ECE}   & \textbf{BS} \\
			\hline
			\textbf{MP}    & 0.800 & 0.790 & 0.683 & 31.8\% & 0.300\\
			\textbf{TS}    & 0.813 & 0.805 & 0.690 & 2.5\% & 0.174 \\
			\textbf{SB}    & 0.852 & 0.848 & 0.734 & 2.7\% & 0.155 \\
			\textbf{Dirichlet} & 0.840 & 0.829 & 0.724 & 4.4\% & 0.163 \\
			\textbf{\basicModel} & \textbf{0.865} & \textbf{0.867} & \textbf{0.741} & \textbf{1.6\%} & \textbf{0.148} \\
			\textbf{\advanceModel} & 0.851 & 0.851 & 0.726 & 1.7\% & 0.155 \\
			\textbf{\transferMethod} & 0.846 & 0.841 & 0.725 & 2.4\% & 0.158 \\
			\hline
		\end{tabular}%
		
	}
\end{table}

\begin{table}
	\centering
	\label{table:cifar100_ood_ad}
	\setlength{\tabcolsep}{1.5pt}
	\parbox{.48\linewidth}{
		\centering
		\caption{VGG16 on CIFAR-100 OOD}
		\label{table:cifar100_in_ECE_OOD}
		\begin{tabular}{lccccc}
			\hline
			\textbf{Method} & \textbf{AUROC} & \textbf{AUPR}  & \textbf{p.9}   & \textbf{ECE}   & \textbf{BS} \\
			\hline
			\textbf{MP}    & 0.890 & 0.690 & 0.477 & 16.5\% & 0.161 \\
			\textbf{TS}    & 0.890 & 0.701 & 0.463 & 8.9\% & 0.122 \\
			\textbf{SB}    & 0.877 & 0.617 & 0.522 & 8.2\% & 0.128 \\
			\textbf{Dirichlet} & 0.784 & 0.507 & 0.327 & 3.2\% & 0.145 \\
			\textbf{\basicModel} & \textbf{0.936} & \textbf{0.812} & \textbf{0.611} & 3.1\% & \textbf{0.087} \\
			\textbf{\advanceModel} & 0.921 & 0.769 & 0.555 & 2.7\% & 0.094\\
			\textbf{\transferMethod} & 0.919 & 0.748 & 0.551 & \textbf{2.6\%} & 0.095 \\
			\hline
		\end{tabular}%

	}
	\hfill
	\parbox{.48\linewidth}{
		\centering
		\caption{VGG16 on CIFAR-100 AD}
		\label{table:cifar100_in_ECE_AD}
		\begin{tabular}{lccccc}
			\hline
			\textbf{Method} & \textbf{AUROC} & \textbf{AUPR}  & \textbf{p.9}   & \textbf{ECE}   & \textbf{BS} \\
			\hline
			\textbf{MP}    & 0.916 & 0.858 & 0.788 & 37.2\% & 0.333\\
			\textbf{TS}    & 0.914 & 0.859 & 0.789 & 15.5\% & 0.146 \\
			\textbf{SB}    & 0.916 & 0.883 & 0.769 & 16.1\% & 0.158 \\
			\textbf{Dirichlet} & 0.894 & 0.849 & 0.739 & 13.4\% & 0.153 \\
			\textbf{\basicModel} & \textbf{0.933} & \textbf{0.892} & \textbf{0.833} & 4.9\% & \textbf{0.100} \\
			\textbf{\advanceModel} & 0.924 & 0.872 & 0.815 & 4.3\% & 0.106 \\
			\textbf{\transferMethod} & 0.858 & 0.803 & 0.673 & \textbf{3.9\%} & 0.153 \\
			\hline
		\end{tabular}%
		
	}
\end{table}

\subsubsection{100-class Image Classification with ResNet-50 DNN}
\textbf{Target DNN model.} The target DNN model is a modified ResNet-50 for tiny images (CIFAR). The target model includes batch normalization layers and dropout layers to improve classification performance. It is trained in \textit{PyTorch} on CIFAR-100 training dataset (50k images). The pre-trained weights of target DNN model are downloaded from \cite{resnet_weights_github_2018}.

All the other settings, such as the datasets and architecture
of the neural network in \ouralg/\ouralgtwo, are the same as those for the 100-class VGG16 DNN
described in Section~\ref{sec:100class_vgg}. The inference accuracies on the four test datasets are 55\% (D1), 38\% (D2), 65\% (OOD), and 37\% (AD).

\textbf{Results. } The calibration results are presented in Table~\ref{table:cifar100_resnet_in_ECE_D1}, \ref{table:cifar100_resnet_in_ECE_D2}, \ref{table:cifar100_resnet_in_ECE_OOD}, \ref{table:cifar100_resnet_in_ECE_AD} for D1, D2, OOD and AD datasets, respectively.

The results show that the proposed method (\basicModel) outperform the baselines in terms of both mis-classification detection and confidence calibration on OOD and AD datasets. As for augmented datasets D1 and D2, our proposed methods can still achieve better mis-classification detection performance and the calibration performance is almost as good as the best calibration in terms of ECE. In addition, even though \transferMethod model is transferred to OOD/AD datasets with fewer training samples than the baselines, it still offers a good calibration performance in terms of ECE and BS.

\begin{figure*}[!t]
	\centering
	\subfigure[]{\includegraphics[trim=0cm 0cm 0cm 0cm,clip,  width=0.195\textwidth]{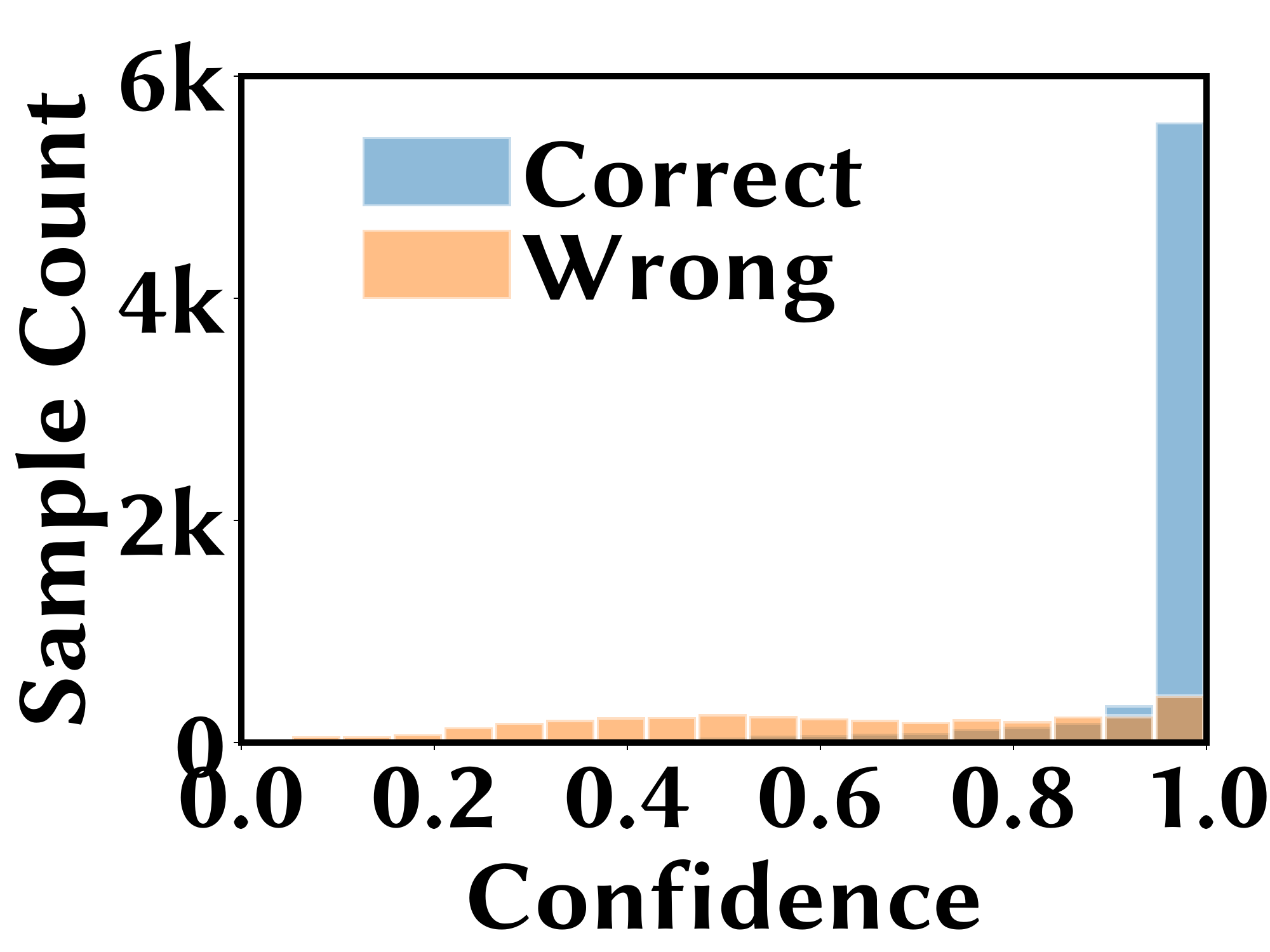}}
	\subfigure[]{\includegraphics[trim=0cm 0cm 0cm 0cm,clip,  width=0.195\textwidth]{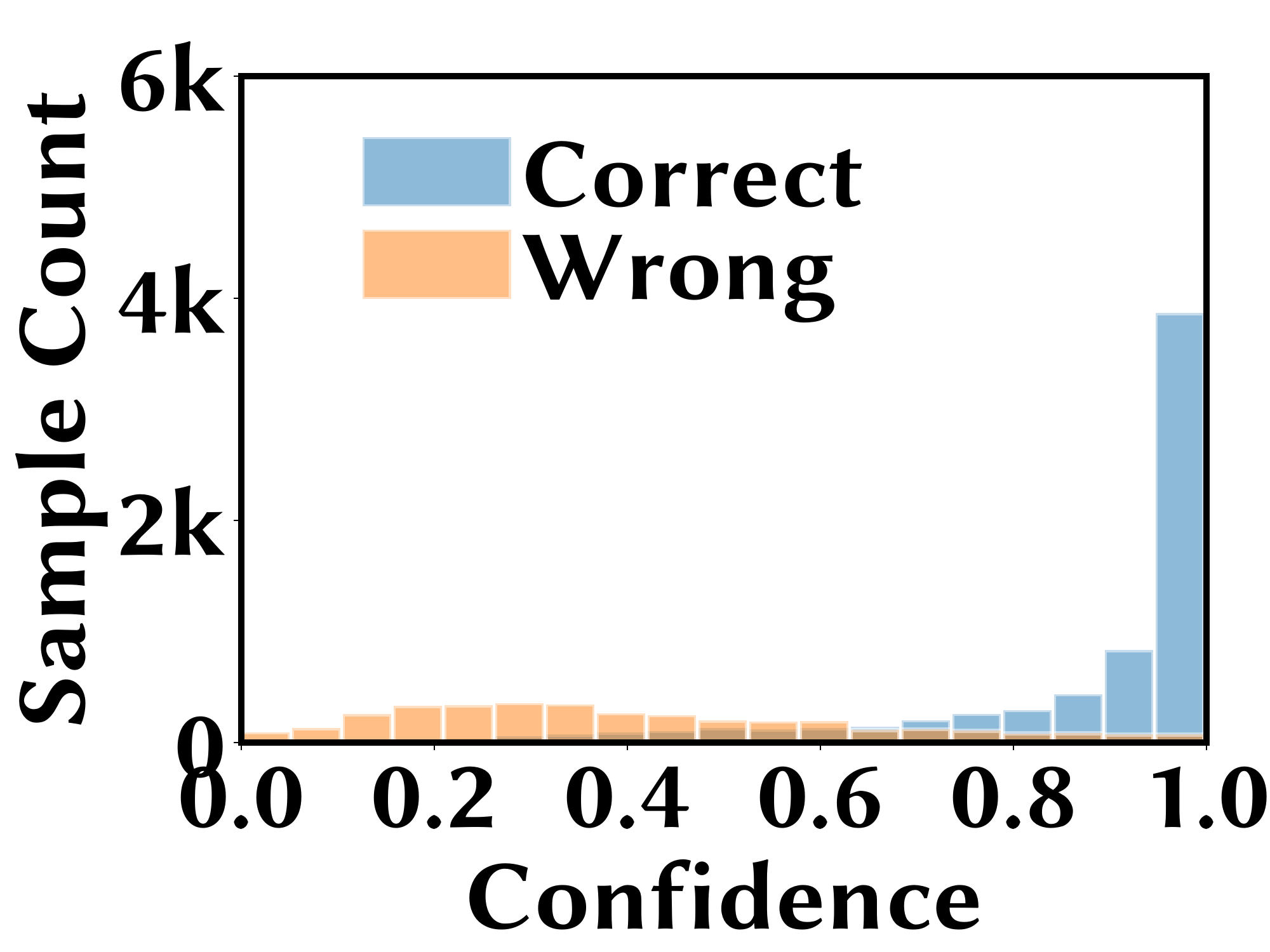}}
	\subfigure[]{\includegraphics[trim=0cm 0cm 0cm 0cm,clip,  width=0.195\textwidth]{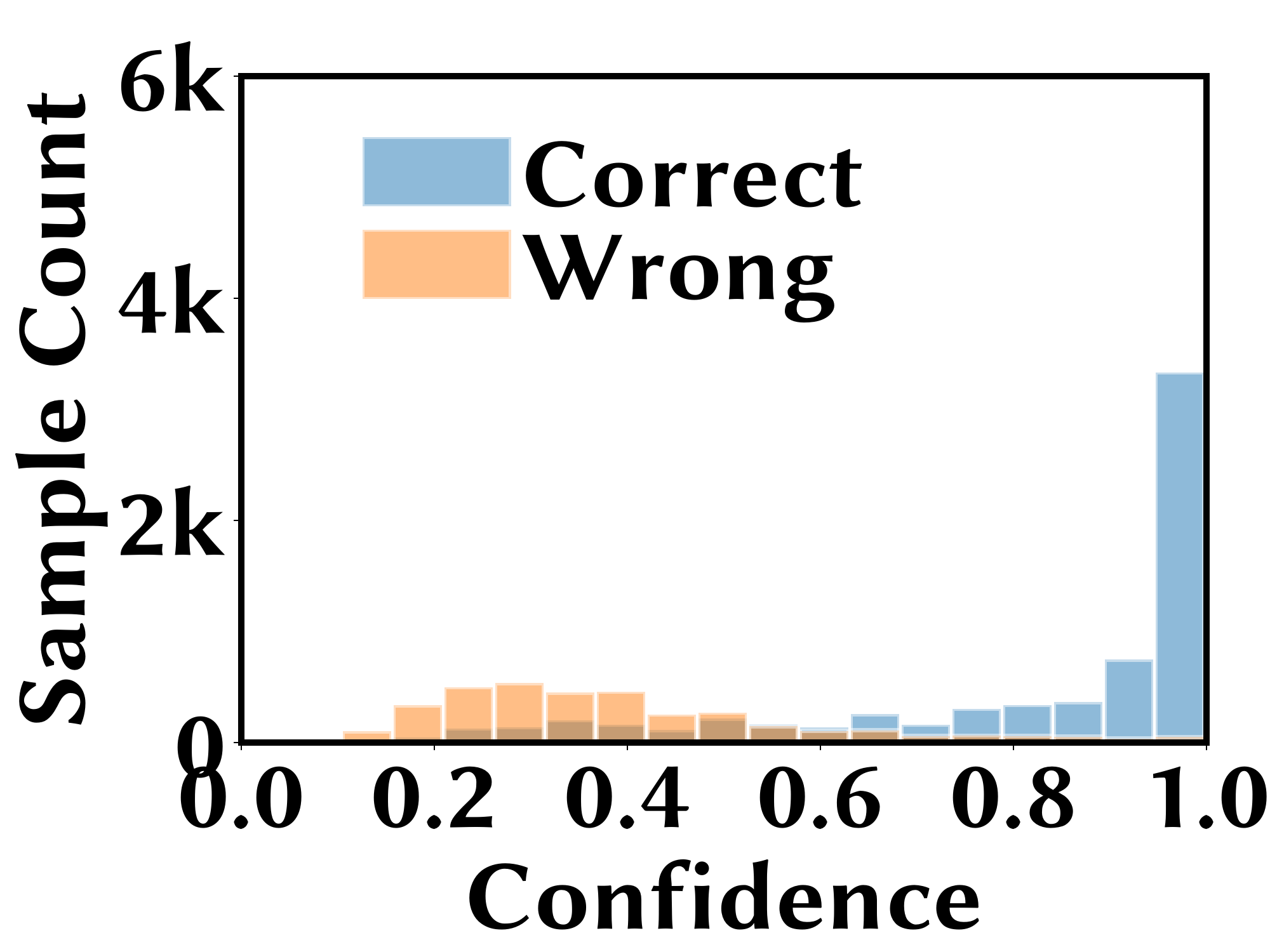}}
	\subfigure[]{\includegraphics[trim=0cm 0cm 0cm 0cm,clip,  width=0.195\textwidth]{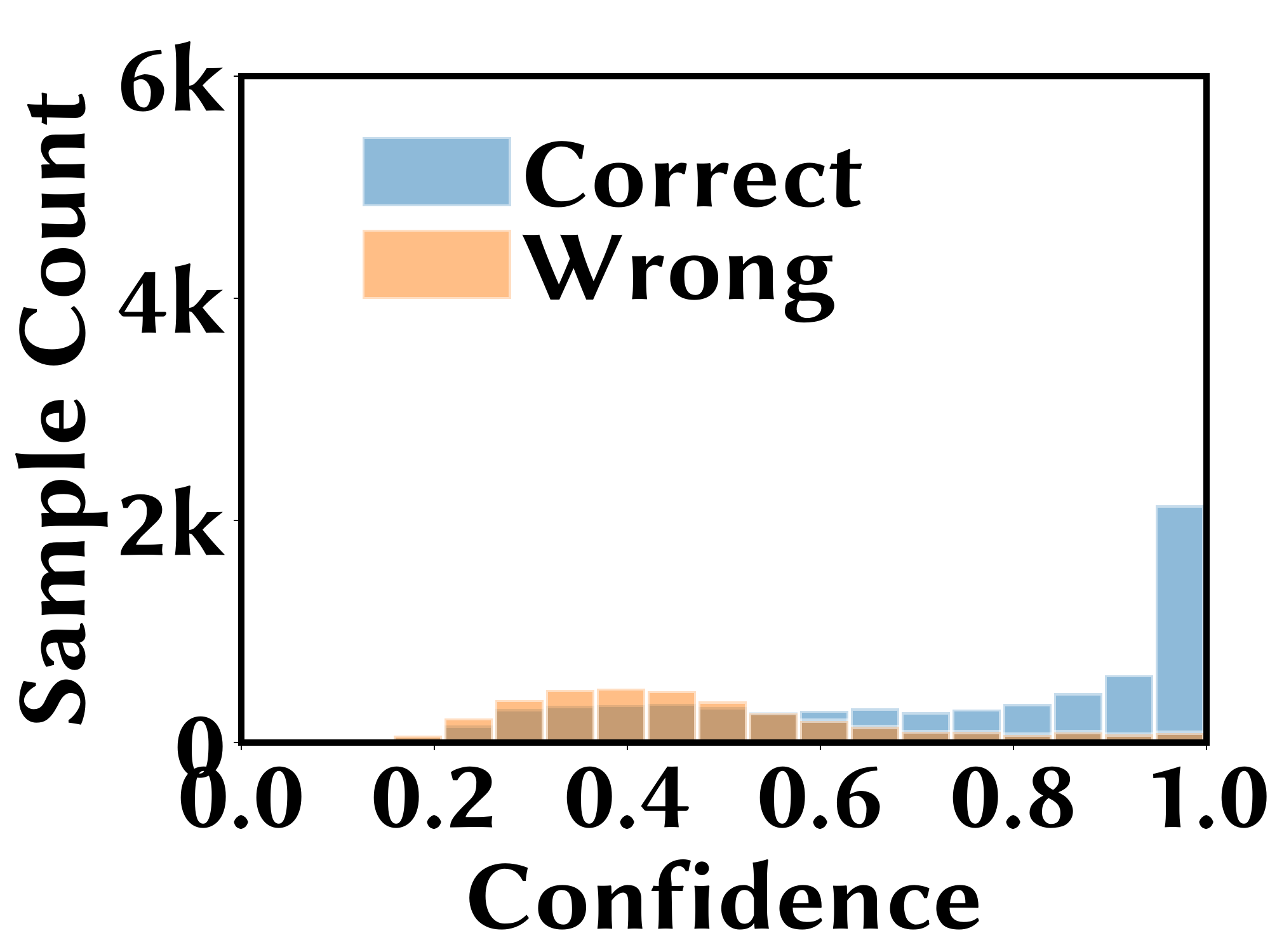}}
	\subfigure[]{\includegraphics[trim=0cm 0cm 0cm 0cm,clip,  width=0.195\textwidth]{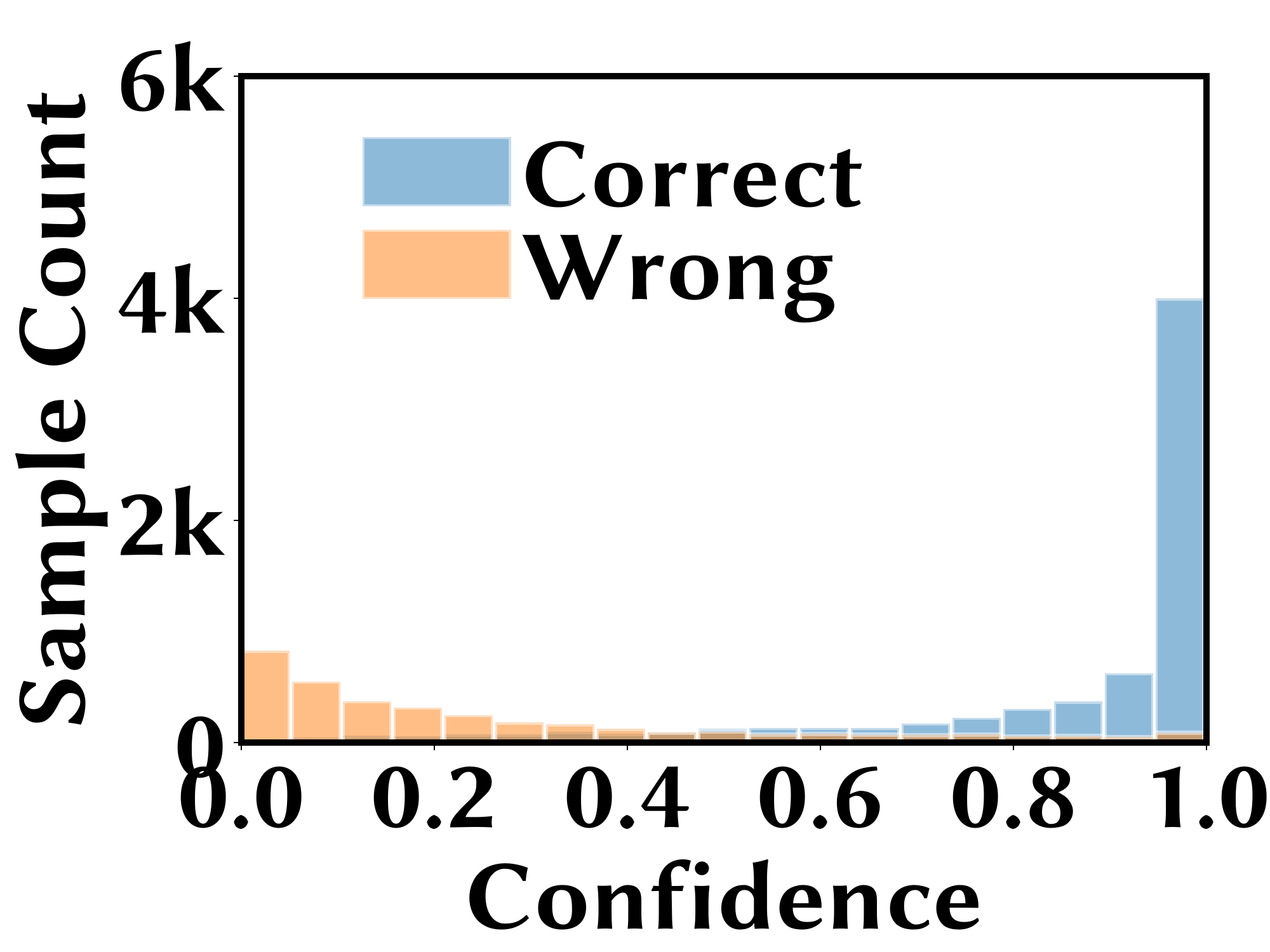}}
	\vspace{-0.3cm}
	\caption{Histograms of prediction confidences for ResNet-50 on CIFAR-100 OOD dataset. (a) Un-calibrated. (b) Temperature scaling. (c) Scaling-binning. (d) Dirichlet calibration. (e) \ouralg.}\label{fig:cifar100_resnet_OOD_histogram}
	\vspace{-0.3cm}
\end{figure*}

\begin{figure*}[!t]
	\centering
	\subfigure[]{\includegraphics[trim=0cm 0cm 0cm 0cm,clip,  width=0.195\textwidth]{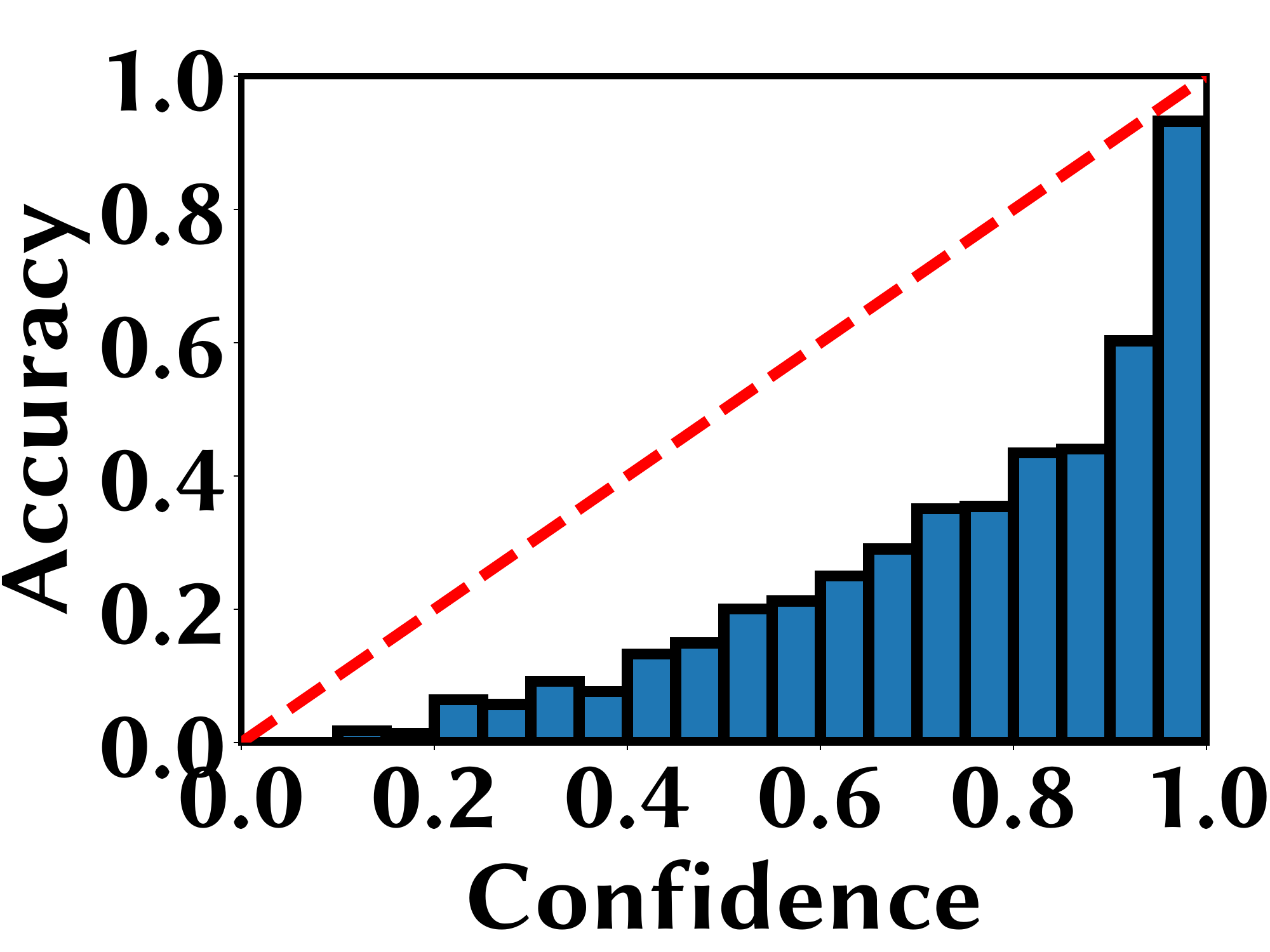}}
	\subfigure[]{\includegraphics[trim=0cm 0cm 0cm 0cm,clip,  width=0.195\textwidth]{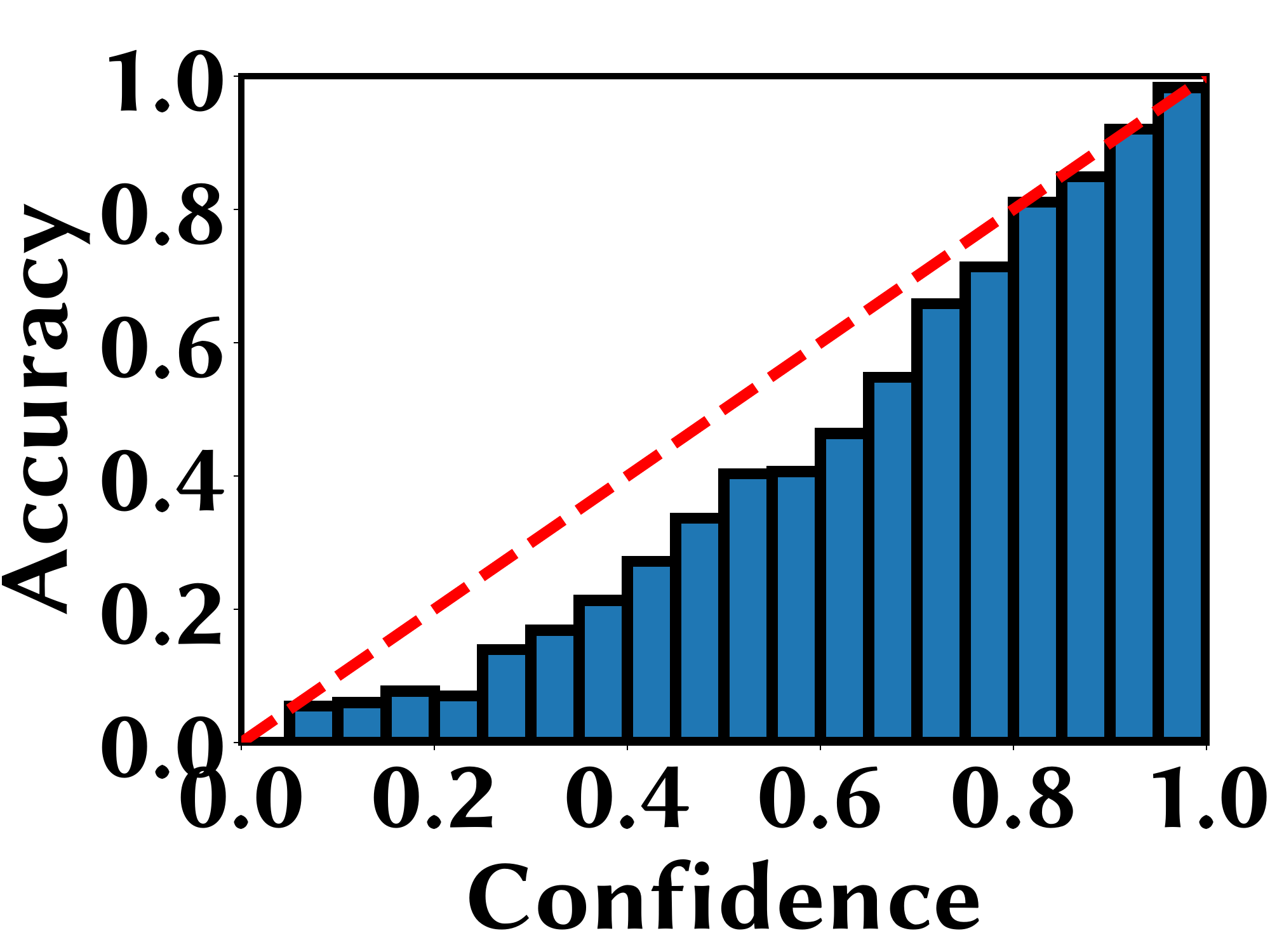}}
	\subfigure[]{\includegraphics[trim=0cm 0cm 0cm 0cm,clip,  width=0.195\textwidth]{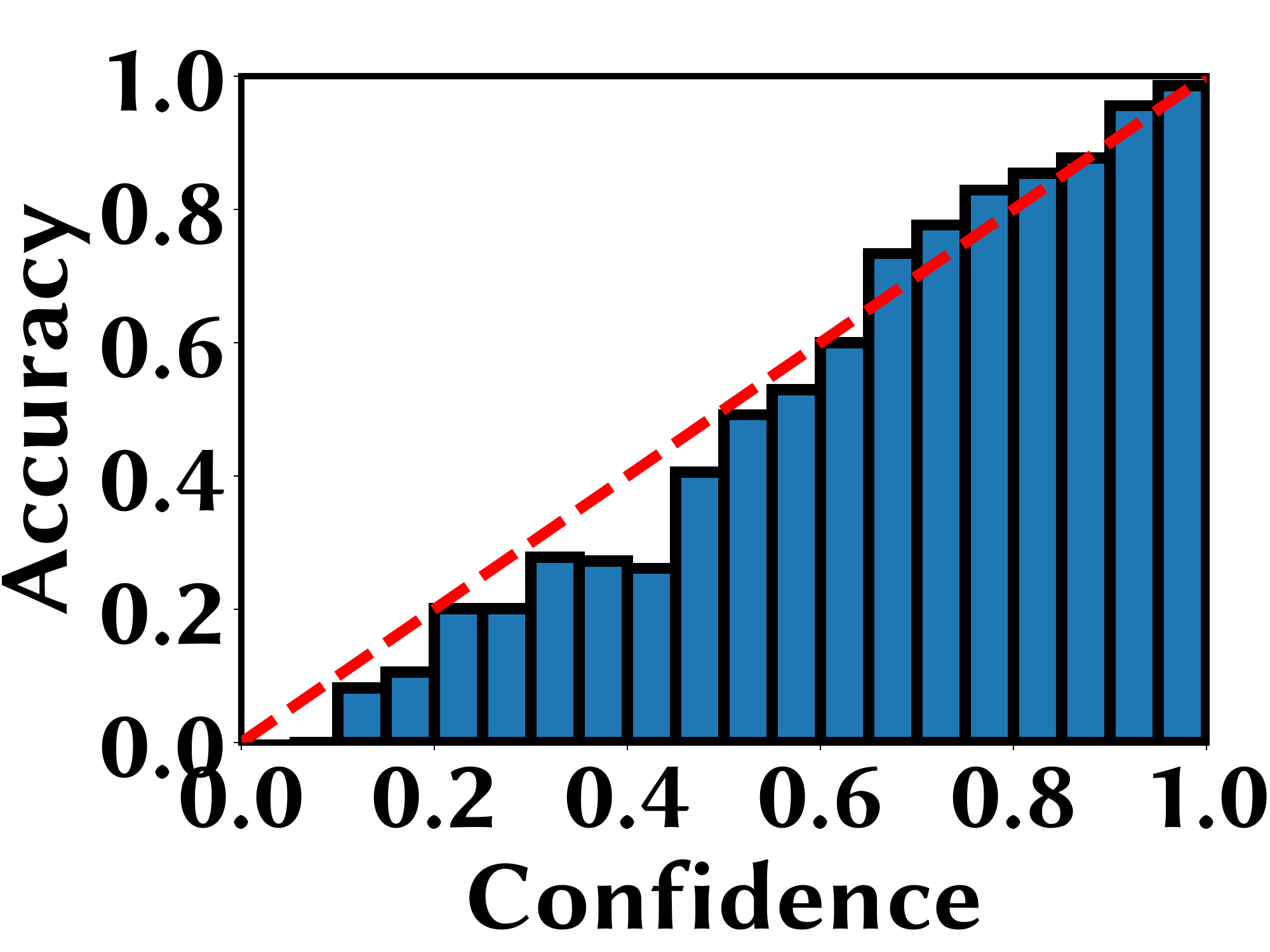}}
	\subfigure[]{\includegraphics[trim=0cm 0cm 0cm 0cm,clip,  width=0.195\textwidth]{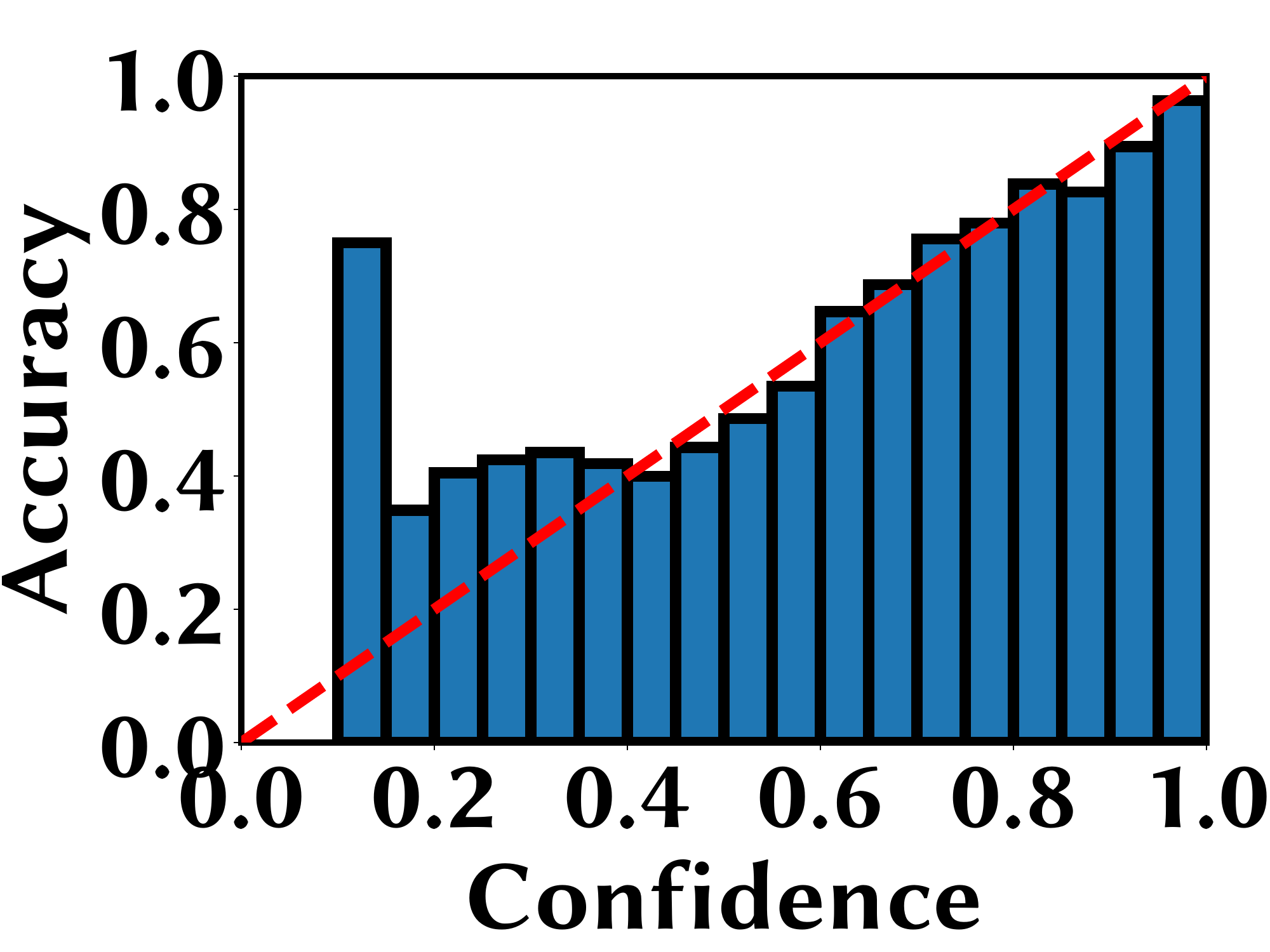}}
	\subfigure[]{\includegraphics[trim=0cm 0cm 0cm 0cm,clip,  width=0.195\textwidth]{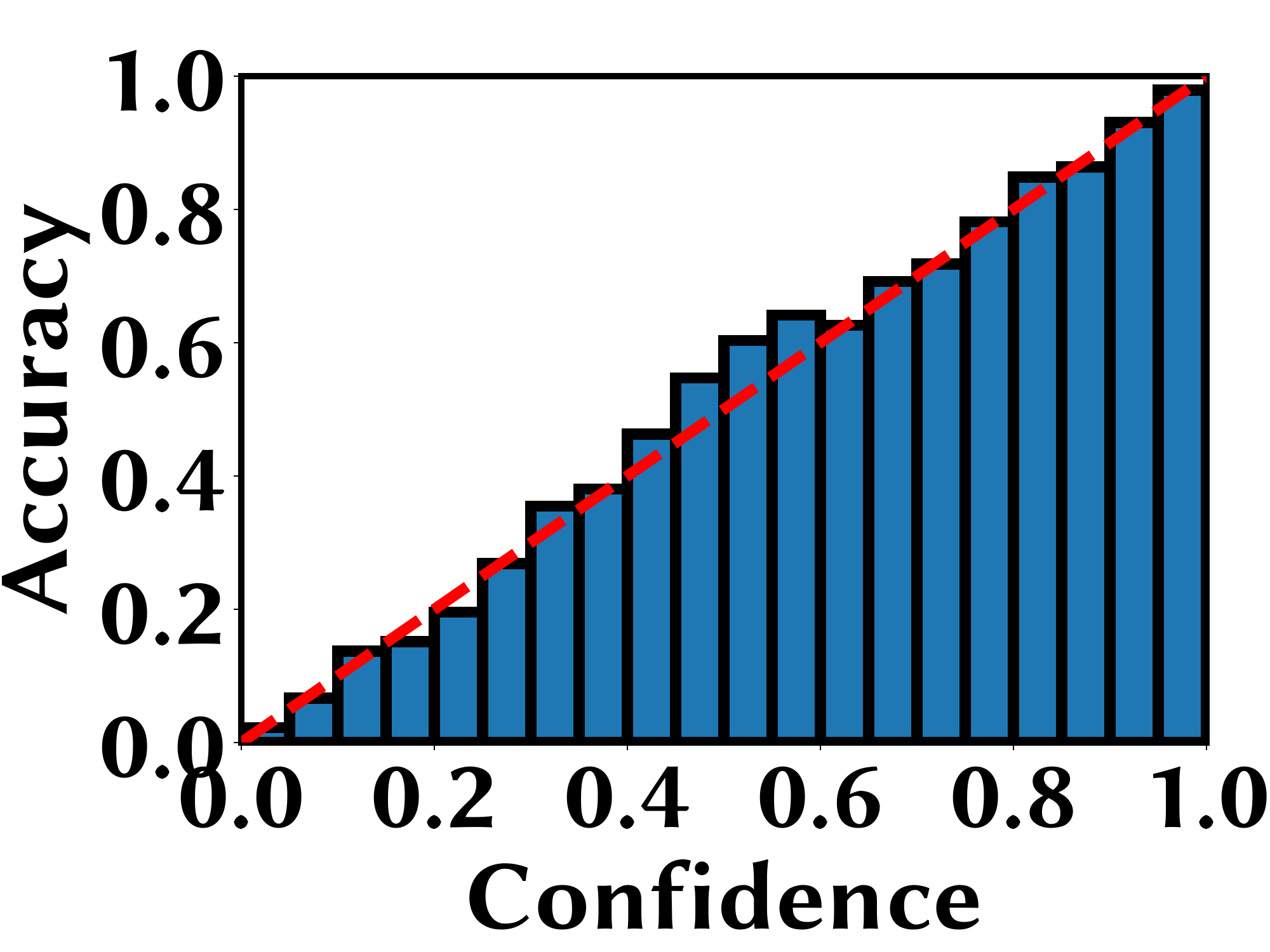}}
	\caption{Reliability diagrams of prediction confidences for ResNet-50 on CIFAR-100 OOD dataset. (a) Un-calibrated. (b) Temperature scaling. (c) Scaling-binning. (d) Dirichlet calibration. (e) \ouralg.}\label{fig:RD_plot_cifar100_resnet_OOD}
	\vspace{-0.3cm}
\end{figure*}

We further show the histograms of confidences for correct and wrong predictions in Fig.~\ref{fig:cifar100_resnet_OOD_histogram} on the OOD dataset. The corresponding reliability diagrams also are shown in Fig.~\ref{fig:RD_plot_cifar100_resnet_OOD} on the OOD dataset with different calibration methods.

\begin{table}
	\centering
	\setlength{\tabcolsep}{1.5pt}
	\parbox{.48\linewidth}{
		\centering
		\caption{ResNet-50 on CIFAR-100 D1}
		\label{table:cifar100_resnet_in_ECE_D1}
		\begin{tabular}{lccccc}
			\hline
			\textbf{Method} & \textbf{AUROC} & \textbf{AUPR} & \textbf{p.9} & \textbf{ECE} & \textbf{BS}\\
			\hline
			\textbf{MP}    & 0.824 & {0.767} & 0.623 & 23.1\% & 0.237 \\
			\textbf{TS}    & 0.831 & \textbf{0.776} & 0.628 & \textbf{1.6\%} & 0.167 \\
			\textbf{SB}    & 0.833 & 0.774 & {0.638} & 2.1\% & {0.166} \\
			\textbf{Dirichlet} & 0.732 & 0.645 & 0.549 & 6.7\% & 0.211 \\
			\textbf{\basicModel} & \textbf{0.836} & 0.762 & \textbf{0.643} & 3.3\% & \textbf{0.165}  \\
			\textbf{\advanceModel} & 0.821 & 0.752 & 0.621 & 2.4\% & 0.172  \\
			\textbf{\transferMethod} &-&-&-&-&-  \\
			\hline
		\end{tabular}%
		
	}
	\hfill
	\parbox{.48\linewidth}{
		\centering
		\caption{ResNet-50 on CIFAR-100 D2}
		\label{table:cifar100_resnet_in_ECE_D2}
		\begin{tabular}{lccccc}
			\hline
			\textbf{Method} & \textbf{AUROC} & \textbf{AUPR} & \textbf{p.9} & \textbf{ECE} & \textbf{BS}\\
			\hline
			\textbf{MP}    & 0.779 & 0.832 & 0.732 & 34.5\% & 0.315 \\
			\textbf{TS}    & 0.785 & 0.833 & 0.740 & \textbf{2.1\%} & 0.176 \\
			\textbf{SB}    & 0.794 & 0.838 & 0.748 & 2.8\% & 0.174 \\
			\textbf{Dirichlet} & 0.702 & 0.756 & 0.699 & 3.5\% & 0.204 \\
			\textbf{\basicModel} & \textbf{0.817} & \textbf{0.851} & \textbf{0.768} & 3.9\% & \textbf{0.163} \\
			\textbf{\advanceModel} & 0.765 & 0.810 & 0.738 & 2.4\% & 0.179 \\
			\textbf{\transferMethod} & 0.789 & 0.831 & 0.744 & 3.6\% & 0.173 \\
			\hline
		\end{tabular}
	}\vspace{-0.3cm}
\end{table}

\begin{table}
	\centering
	\setlength{\tabcolsep}{1.5pt}
	\parbox{.48\linewidth}{
		\centering
		\caption{ResNet-50 on CIFAR-100 OOD}
		\label{table:cifar100_resnet_in_ECE_OOD}
		\begin{tabular}{lccccc}
			\hline
			\textbf{Method} & \textbf{AUROC} & \textbf{AUPR} & \textbf{p.9} & \textbf{ECE} & \textbf{BS}\\
			\hline
			\textbf{MP}    & 0.922 & 0.851 & 0.695 & 17.8\% & 0.161 \\
			\textbf{TS}    & 0.929 & 0.862 & 0.721 & 6.1\% & 0.108 \\
			\textbf{SB}    & 0.903 & 0.788 & 0.689 & 3.8\% & 0.119 \\
			\textbf{Dirichlet} & 0.776 & 0.563 & 0.507 & 4.6\% & 0.179 \\
			\textbf{\basicModel} & \textbf{0.939} & \textbf{0.894} & \textbf{0.726} & \textbf{1.5\%} & \textbf{0.092} \\
			\textbf{\advanceModel} & 0.902 & 0.784 & 0.665 & 2.9\% & 0.120 \\
			\textbf{\transferMethod} & 0.926 & 0.854 & 0.707 & 3.2\% & 0.105 \\
			\hline
		\end{tabular}%
		
	}
	\hfill
	\parbox{.48\linewidth}{
		\centering
		\caption{ResNet-50 on CIFAR-100 AD}
		\label{table:cifar100_resnet_in_ECE_AD}
		\begin{tabular}{lccccc}
			\hline
			\textbf{Method} & \textbf{AUROC} & \textbf{AUPR} & \textbf{p.9} & \textbf{ECE} & \textbf{BS}\\
			\hline
			\textbf{MP} & 0.738 & 0.824 & 0.698 & 42.1\% & 0.387\\
			\textbf{TS} & 0.743 & 0.830 & 0.699 & 8.8\% & 0.203 \\
			\textbf{SB} & 0.766 & 0.831 & 0.721 & 9.3\% & 0.196 \\
			\textbf{Dirichlet} & 0.700 & 0.768 & 0.700 & 8.0\% & 0.210 \\
			\textbf{\basicModel} & \textbf{0.790} & 0.842 & \textbf{0.756} & 5.5\% & \textbf{0.176} \\
			\textbf{\advanceModel} & 0.786 & \textbf{0.857} & 0.732 & \textbf{5.1\%} & 0.182 \\
			\textbf{\transferMethod} & 0.756 & 0.819 & 0.717 & 6.9\% & 0.199 \\
			\hline
		\end{tabular}
		
	}\vspace{-0.5cm}
\end{table}

\subsubsection{1000-class Image Classification with VGG16 DNN}\label{sec:1000class_vgg}
\textbf{Target DNN model. } The target DNN model is the VGG16 for 1000-class image classification. The pre-trained weights of VGG16 model are directly downloaded from the Keras application package \cite{keras_2015}, which are trained on ImageNet 2012 training dataset \cite{imagenet_dataset_2012_IJCV}.

\textbf{Datasets.} We evaluate the calibration performance on two datasets generated from ImageNet 2012 validation dataset \cite{imagenet_dataset_2012_IJCV}, including augmented dataset D1 and one out-of-distribution dataset (OOD). We first randomly select 30k samples from ImageNet 2012 validation, and then perform augmentation operations on the selected samples via \textit{ImageDataGenerator} in \textit{Tensorflow}. For D1, the augment operations and parameters include rotation within $[-40,40]$ degrees, vertical/horizontal shift within $[-0.5,0.5]$, and horizontal flip. The OOD dataset includes 15k samples from ImageNet validation dataset and 15k samples from CIFAR-100 dataset serving as OOD samples. To fit for the input size of DNNs, the OOD samples from CIFAR-100 dataset are resized into $(224,224,3)$ using \textit{tf.image.resize} with default parameters. The OOD samples are treated with ``NULL'' label, indicating not belonging to any of the 1000 classes. The inference accuracies on the two test datasets are 51\% (D1) and 32\% (OOD). Additionally, the 30k samples (in D1 or OOD ) are randomly split into a 10k dataset for training, a 2k validation dataset for hyperparameter tuning, and a 18k testing dataset for performance evaluation.

\begin{table}
	\centering
	\setlength{\tabcolsep}{1.5pt}
	\parbox{.48\linewidth}{
		\centering
		\caption{VGG16 on ImageNet D1}
		\label{table:imagenet_vgg_ECE_D1}
		\begin{tabular}{lccccc}
			\hline
			\textbf{Method} & \textbf{AUROC} & \textbf{AUPR} & \textbf{p.9} & \textbf{ECE} & \textbf{BS} \\
			\hline
			\textbf{MP} & \textbf{0.846} & \textbf{0.815} & \textbf{0.686} & 2.5\% & \textbf{0.160} \\
			\textbf{TS} & 0.846 & 0.815 & 0.686 & 2.5\% & 0.160 \\
			\textbf{\advanceModel} & 0.784 & 0.743 & 0.621 & \textbf{2.4\%} & 0.189 \\
			\hline
		\end{tabular}%
	}
	\hfill
	\parbox{.48\linewidth}{
		\centering
		\caption{VGG16 on ImageNet OOD}
		\label{table:imagenet_vgg_ECE_OOD}
		\begin{tabular}{lccccc}
			\hline
			\textbf{Method} & \textbf{AUROC} & \textbf{AUPR} & \textbf{p.9} & \textbf{ECE} & \textbf{BS}\\
			\hline
			\textbf{MP} & 0.925 & 0.961 & 0.887 & 14.0\% & 0.124\\
			\textbf{TS} & 0.931 & 0.964 & 0.896 & 7.2\% & 0.102 \\
			\textbf{\advanceModel} & \textbf{0.953} & \textbf{0.979} & \textbf{0.915} & \textbf{2.0\%} & \textbf{0.083} \\
			\hline
		\end{tabular}%
		
	}
\end{table}

\begin{figure*}[!t]
	\centering
	\subfigure[]{\includegraphics[trim=0cm 0cm 0cm 0cm,clip,  width=0.24\textwidth]{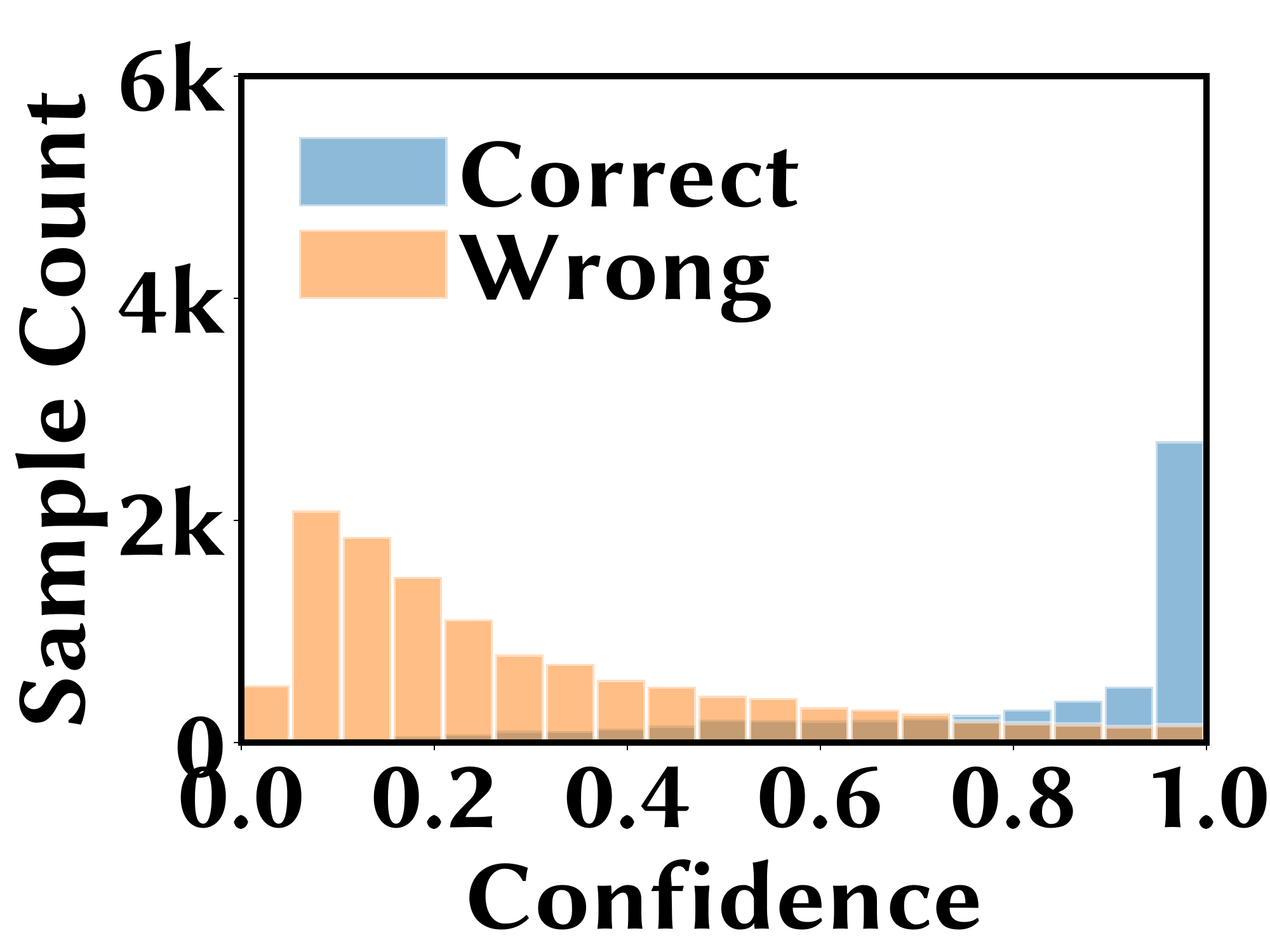}}
	\subfigure[]{\includegraphics[trim=0cm 0cm 0cm 0cm,clip,  width=0.24\textwidth]{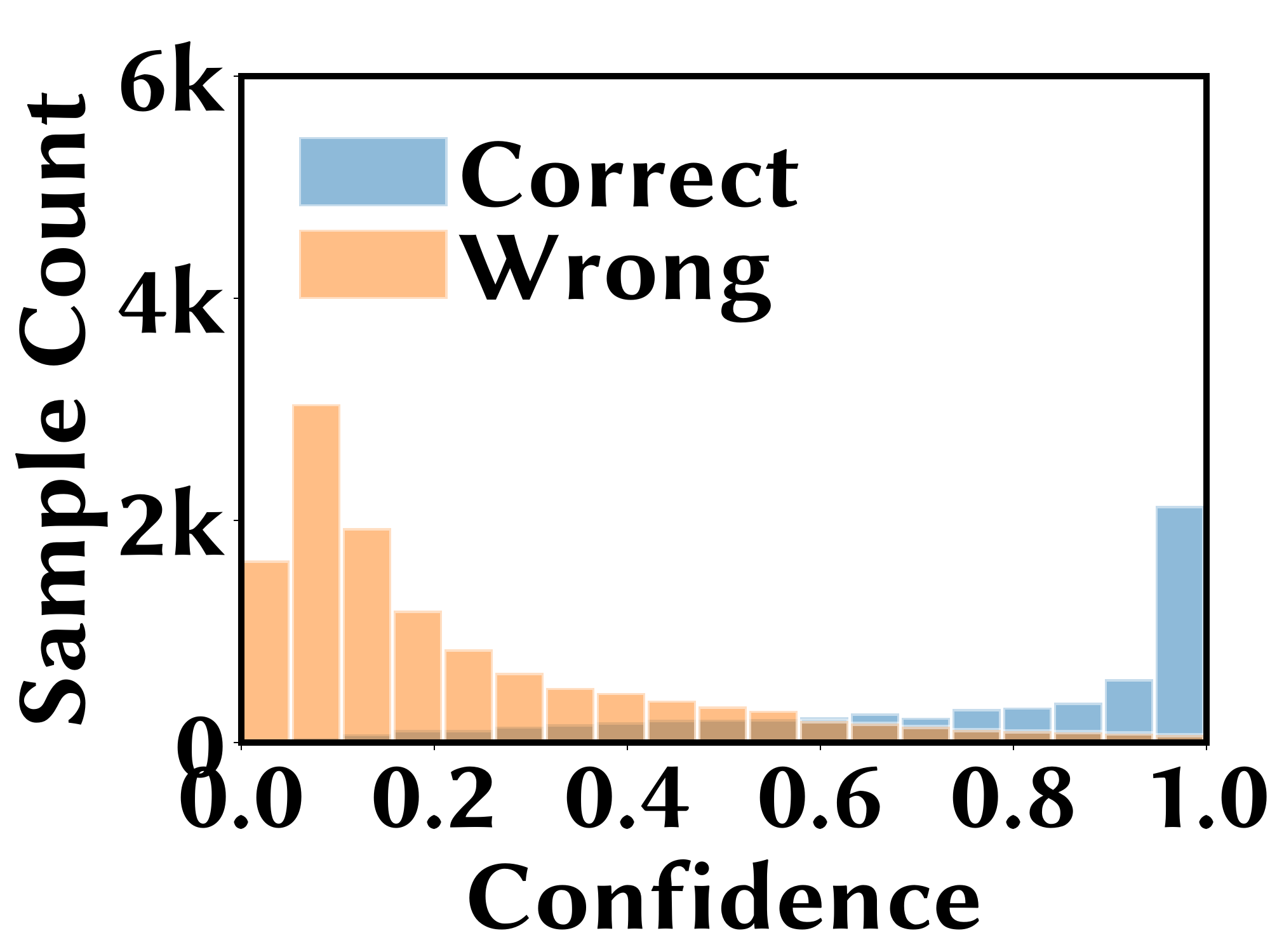}}
	\subfigure[]{\includegraphics[trim=0cm 0cm 0cm 0cm,clip,  width=0.24\textwidth]{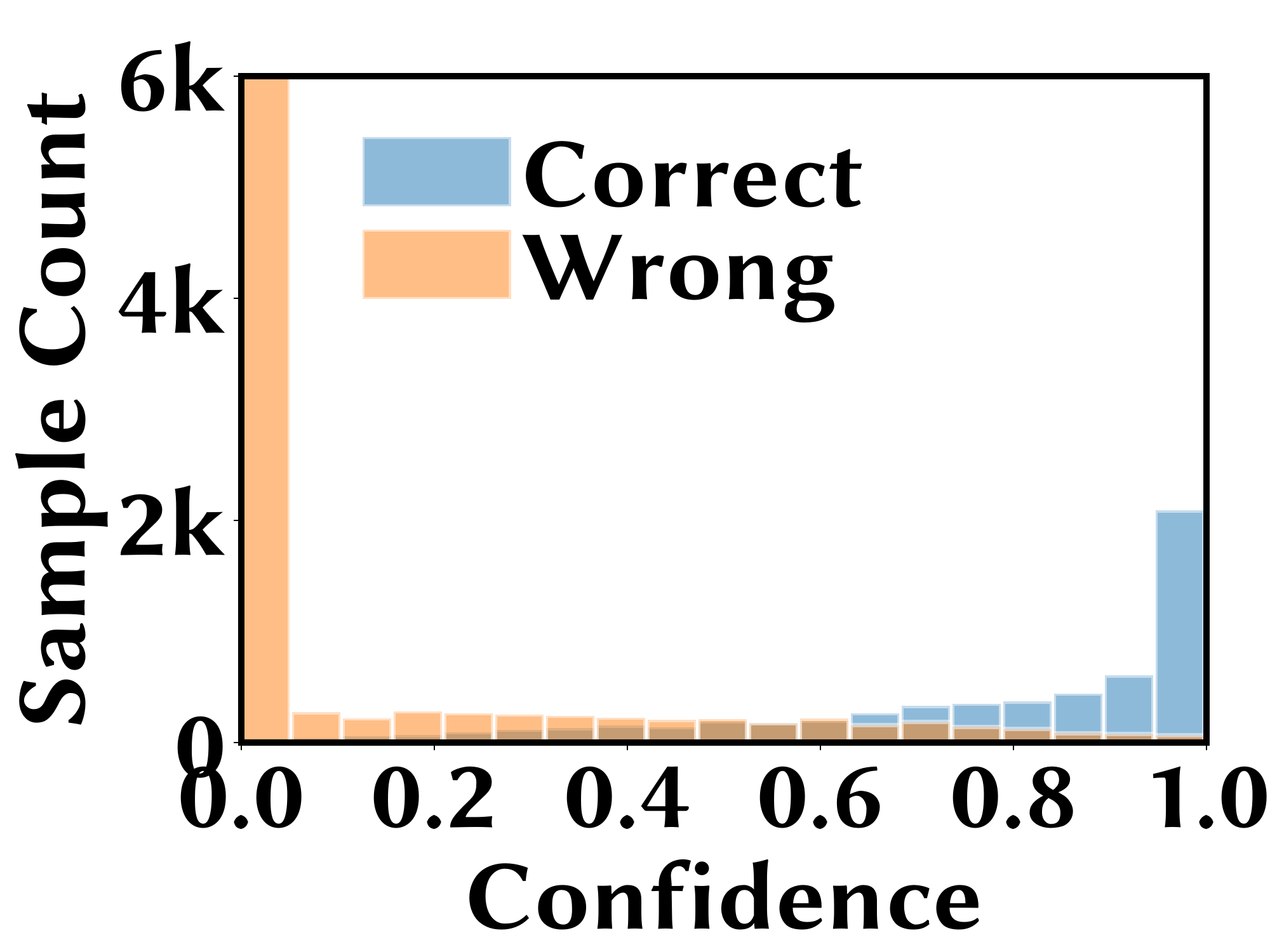}}
	\vspace{-0.3cm}
	\caption{Histograms of prediction confidences for VGG16 on ImageNet OOD dataset. (a) Un-calibrated. (b) Temperature scaling. (c) \advanceModel.}\label{fig:imagenet_OOD_histogram}
	\vspace{-0.3cm}
\end{figure*}

\begin{figure*}[!t]
	\centering
	\subfigure[]{\includegraphics[trim=0cm 0cm 0cm 0cm,clip,  width=0.24\textwidth]{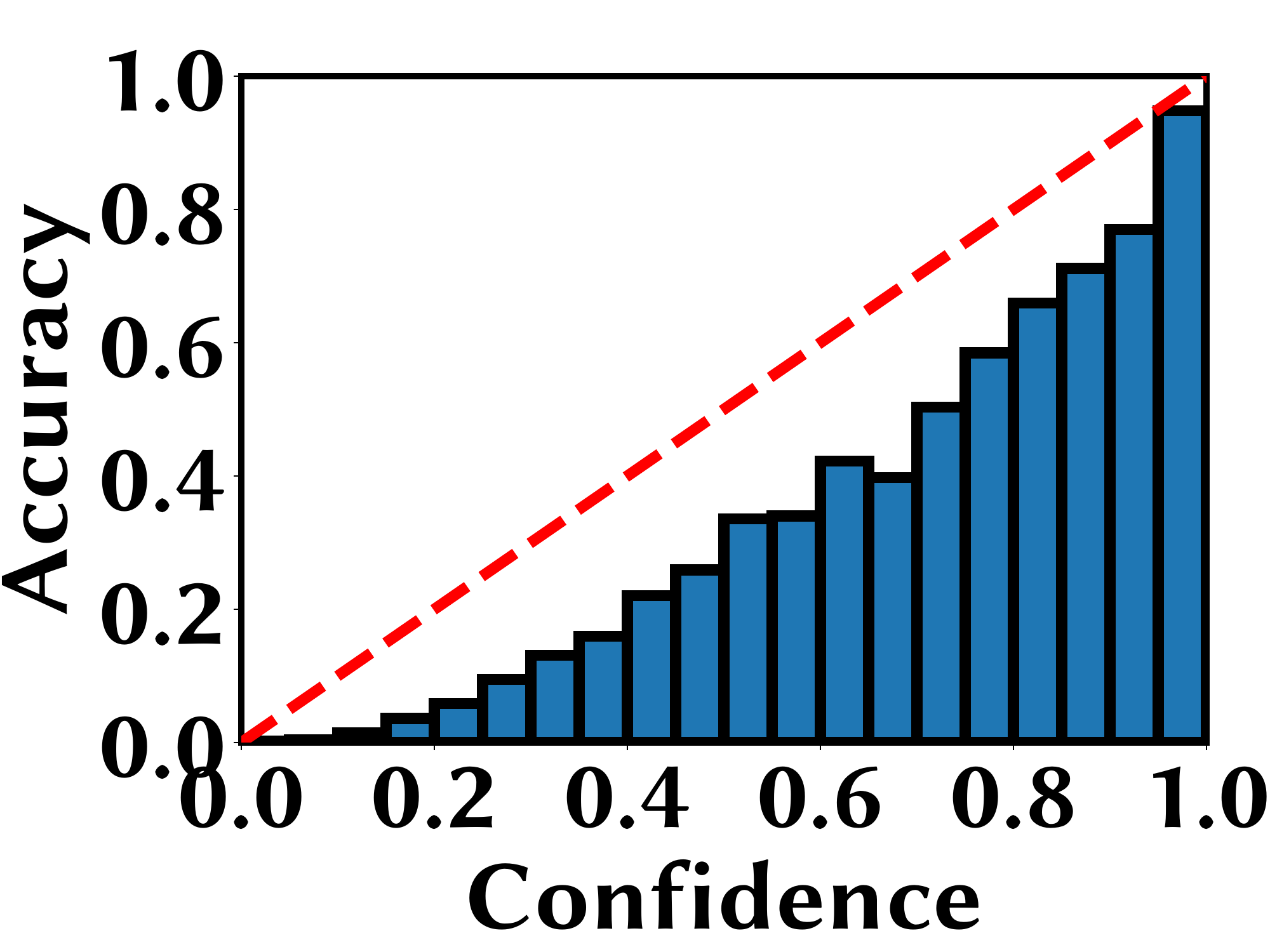}}
	\subfigure[]{\includegraphics[trim=0cm 0cm 0cm 0cm,clip,  width=0.24\textwidth]{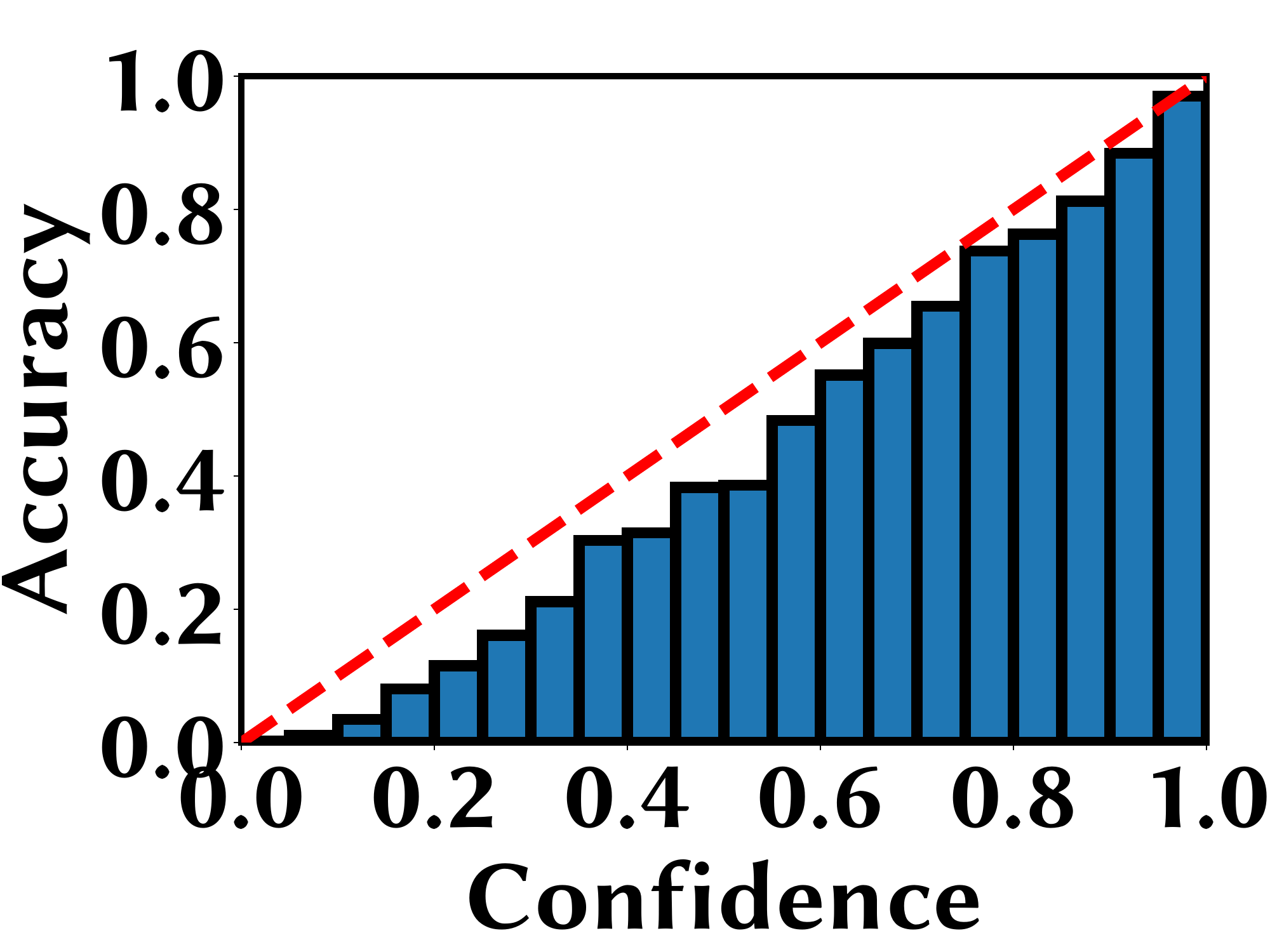}}
	\subfigure[]{\includegraphics[trim=0cm 0cm 0cm 0cm,clip,  width=0.24\textwidth]{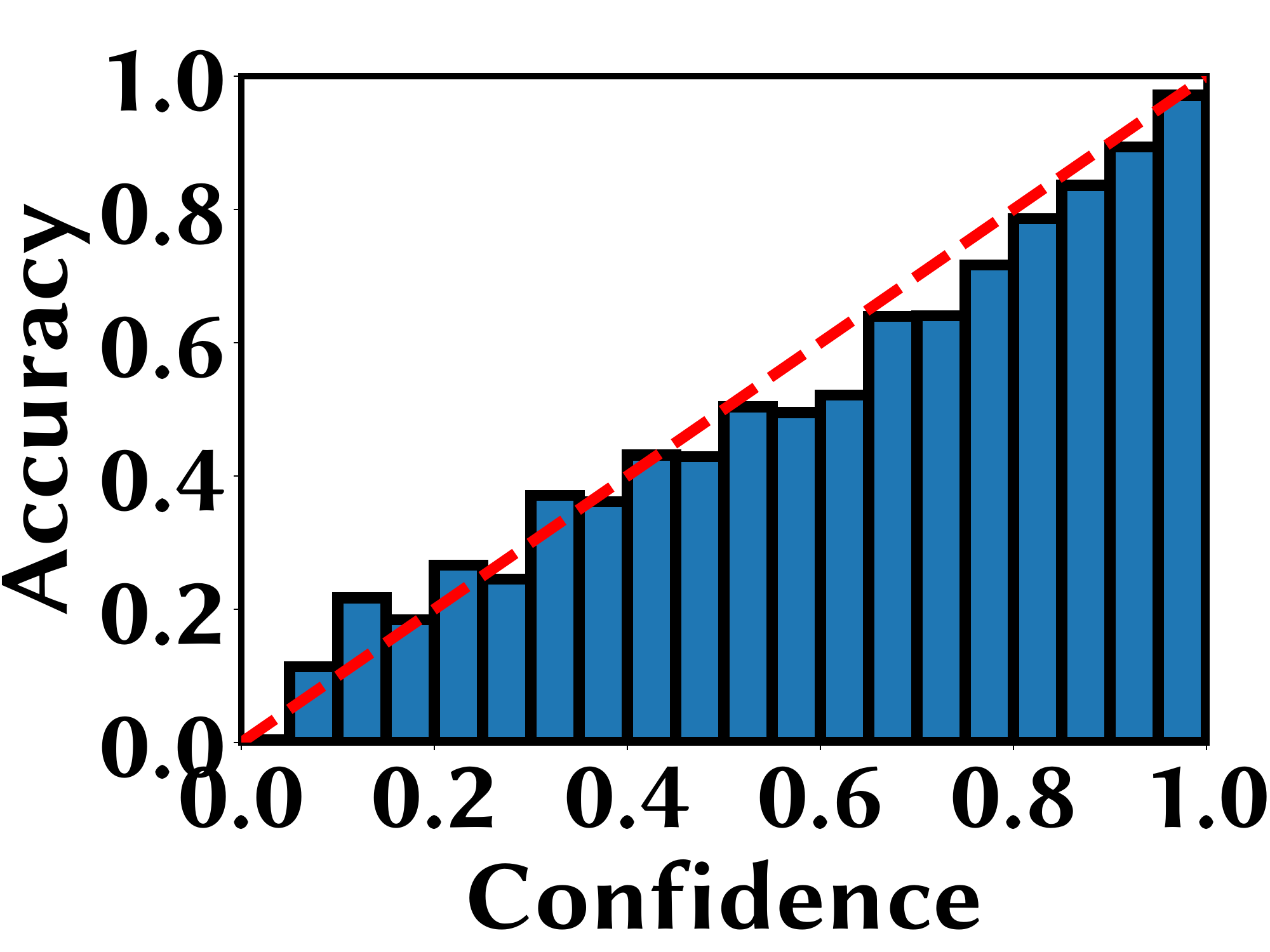}}
	\vspace{-0.3cm}
	\caption{Reliability diagrams of prediction confidences for VGG16 on ImageNet OOD dataset. (a) Un-calibrated. (b) Temperature scaling. (c) \advanceModel.}\label{fig:imagenet_OOD_ECE}
	\vspace{-0.3cm}
\end{figure*}

\textbf{Baselines. } The ImageNet dataset contains 1000 classes, significantly increasing the trainable weights and training dataset size required in  parametric calibration baselines of SB and Dirichlet. Thus, we only keep two baselines (MP and TS) for confidence calibration on ImageNet. For each dataset, the baseline TS is trained on the 10k training dataset. The 10k training dataset is the same as the one used for \ouralgtwo.

\textbf{Our method. } Given 1000 nodes in each layer, a fully-connected layer in \ouralg contains more than $10^6$ weights for 1000-class image classification.
Thus, due to limited size of D1 and OOD datasets, we only implement \ouralgtwo for 1000-class image classification. Specifically, the neural network of \ouralgtwo contains no hidden layers. The input layer contains 1000 nodes and the output layer contains 1001 nodes (including 1000 classes and one ``mis-classification'' class).  \ouralgtwo is trained over 1000 epochs using
the Adam optimizer (learning rate $10^{-4}$). The loss function hyperparameters ($\lambda_1$ and $\lambda_2$) and the confidence calculation method are selected with a minimal ECE on the validation dataset.

\textbf{Results. } The calibration results are presented in Table~\ref{table:imagenet_vgg_ECE_D1} and \ref{table:imagenet_vgg_ECE_OOD} for D1 and OOD datasets, respectively.
The results show that the proposed method (\advanceModel) outperforms the baseline in terms of confidence calibration, with significant improvement on the OOD dataset. In addition, the \advanceModel offers a better mis-classification detection performance on the OOD dataset in terms of AUROC/AUPR/p.9.
We further show the histograms of confidences for correct and wrong predictions in Fig.~\ref{fig:imagenet_OOD_histogram} on the OOD dataset. The corresponding reliability diagrams also are shown in Fig.~\ref{fig:imagenet_OOD_ECE} on the OOD dataset with different calibration methods.

\subsubsection{1000-class Image Classification with ResNet-50 DNN}

\begin{table}
	\centering
	\setlength{\tabcolsep}{1.5pt}
	\parbox{.48\linewidth}{
		\centering
		\caption{ResNet-50 on ImageNet D1}
		\label{table:imagenet_resnet_ECE_D1}
		\begin{tabular}{lccccc}
			\hline
			\textbf{Method} & \textbf{AUROC} & \textbf{AUPR} & \textbf{p.9} & \textbf{ECE} & \textbf{BS}\\
			\hline
			\textbf{MP} & \textbf{0.850} & \textbf{0.735} & 0.570 & 5.3\% & 0.154 \\
			\textbf{TS} & 0.849 & 0.733 & \textbf{0.570} & 2.8\% & \textbf{0.152} \\
			\textbf{\advanceModel} & 0.807 & 0.640 & 0.531 & \textbf{2.3\%} & 0.171 \\
			\hline
		\end{tabular}%
	}
	\hfill
	\parbox{.48\linewidth}{
		\centering
		\caption{ResNet-50 on ImageNet OOD}
		\label{table:imagenet_resnet_ECE_OOD}
		\begin{tabular}{lccccc}
			\hline
			\textbf{Method} & \textbf{AUROC} & \textbf{AUPR} & \textbf{p.9} & \textbf{ECE} & \textbf{BS}\\
			\hline
			\textbf{MP} & 0.886 & 0.931 & 0.843 & 25.0\% & 0.201 \\
			\textbf{TS} & 0.896 & 0.937 & 0.851 & 10.9\% & 0.134 \\
			\textbf{\advanceModel} & \textbf{0.957} & \textbf{0.979} & \textbf{0.916} & \textbf{2.8\%} & \textbf{0.081} \\
			\hline
		\end{tabular}%
		
	}
\end{table}
\begin{figure*}[!t]
	\centering
	\subfigure[]{\includegraphics[trim=0cm 0cm 0cm 0cm,clip,  width=0.24\textwidth]{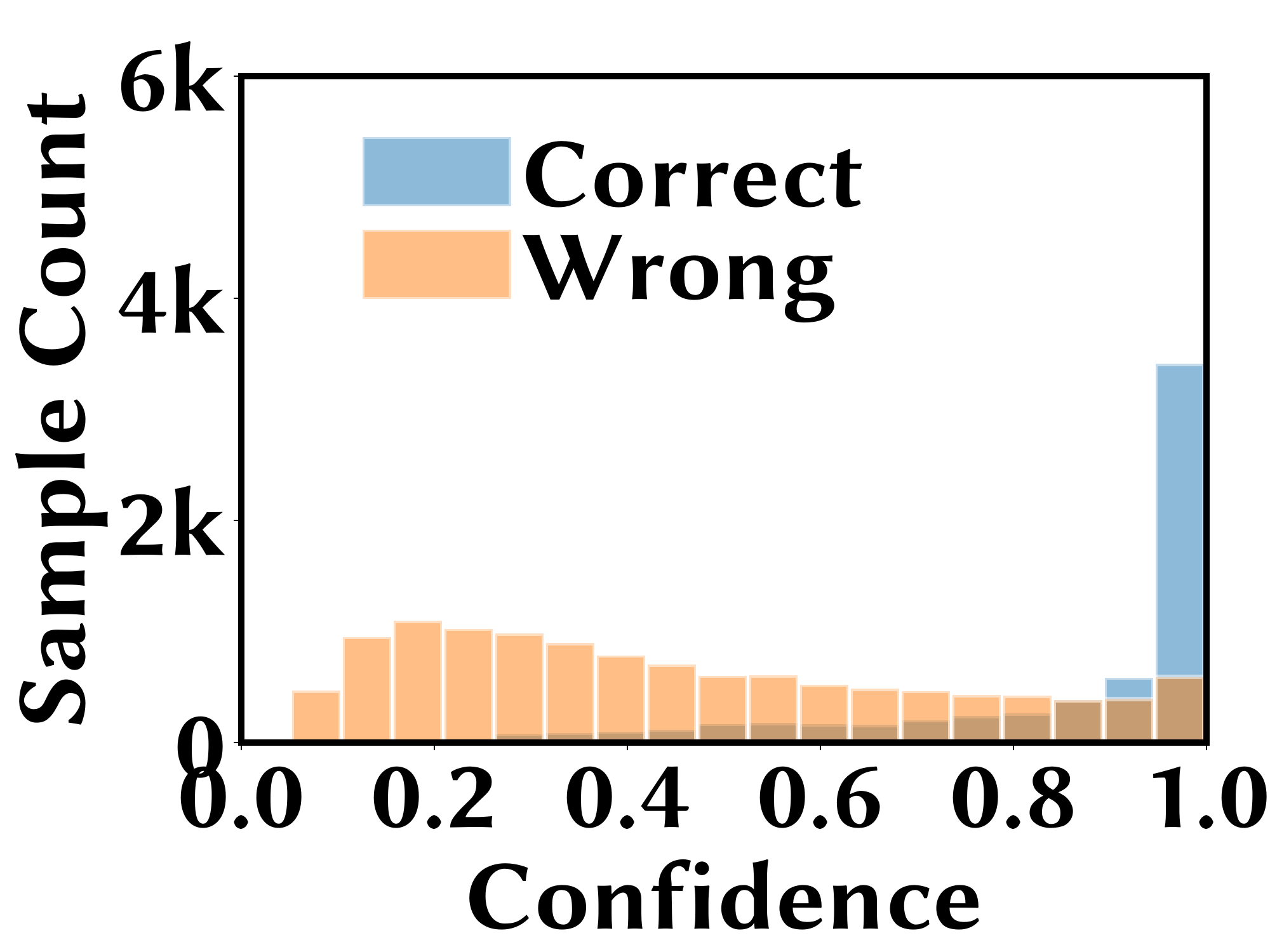}}
	\subfigure[]{\includegraphics[trim=0cm 0cm 0cm 0cm,clip,  width=0.24\textwidth]{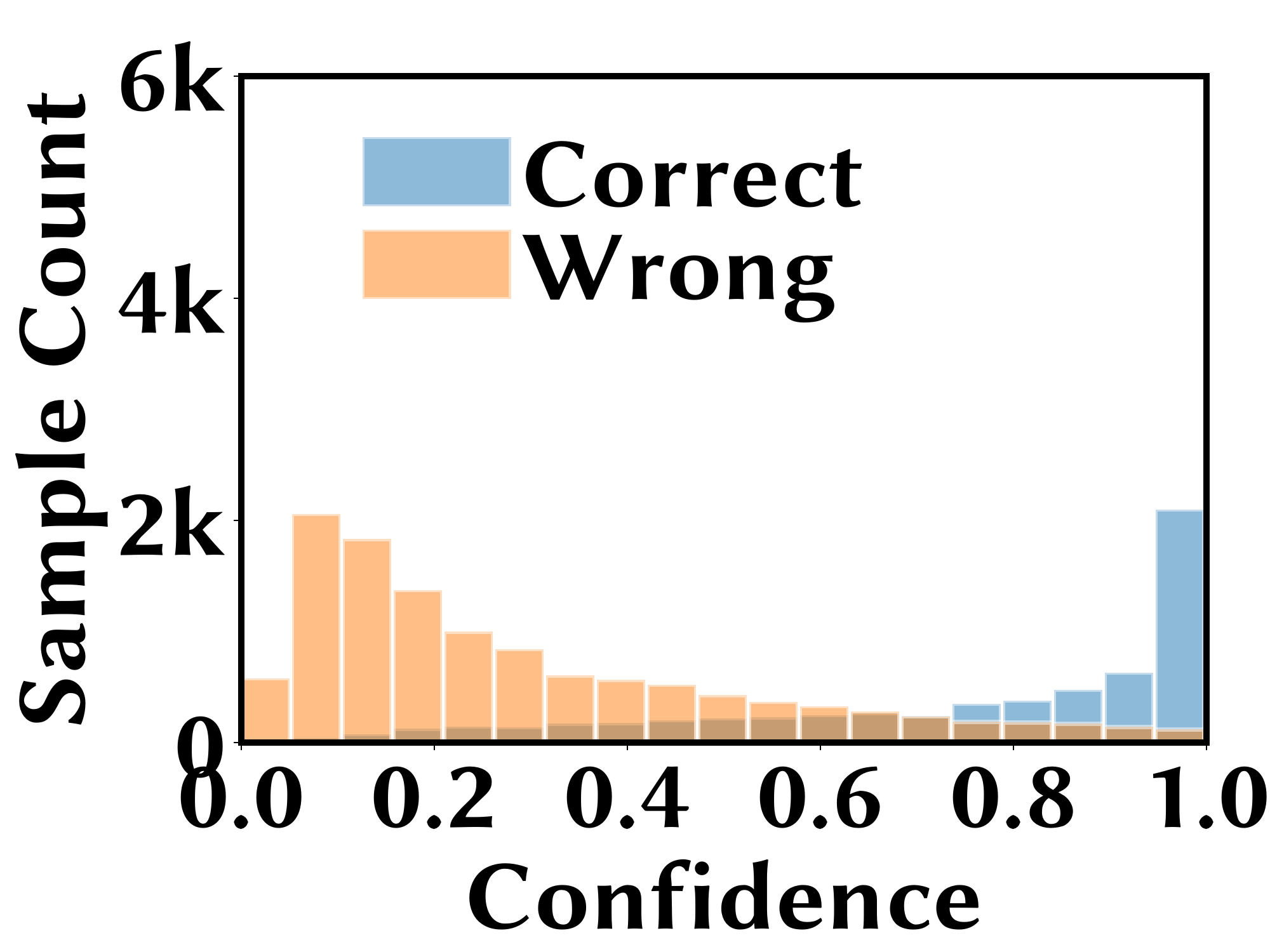}}
	\subfigure[]{\includegraphics[trim=0cm 0cm 0cm 0cm,clip,  width=0.24\textwidth]{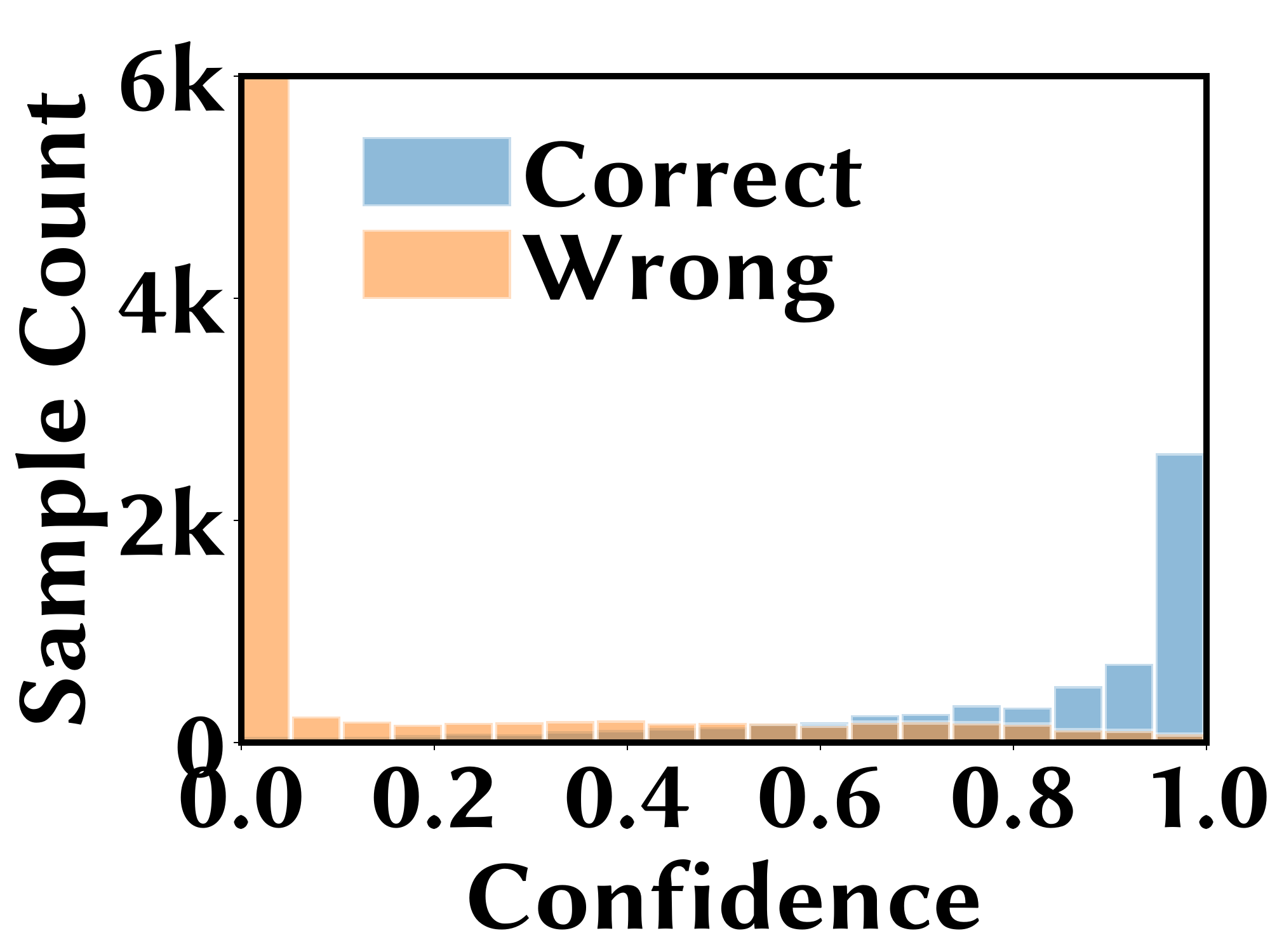}}
	\vspace{-0.3cm}
	\caption{Histograms of prediction confidences for ResNet-50 on ImageNet OOD dataset. (a) Un-calibrated. (b) Temperature scaling. (c) \advanceModel.}\label{fig:imagenet_resnet_OOD_histogram}
	\vspace{-0.3cm}
\end{figure*}

\begin{figure*}[!t]
	\centering
	\subfigure[]{\includegraphics[trim=0cm 0cm 0cm 0cm,clip,  width=0.24\textwidth]{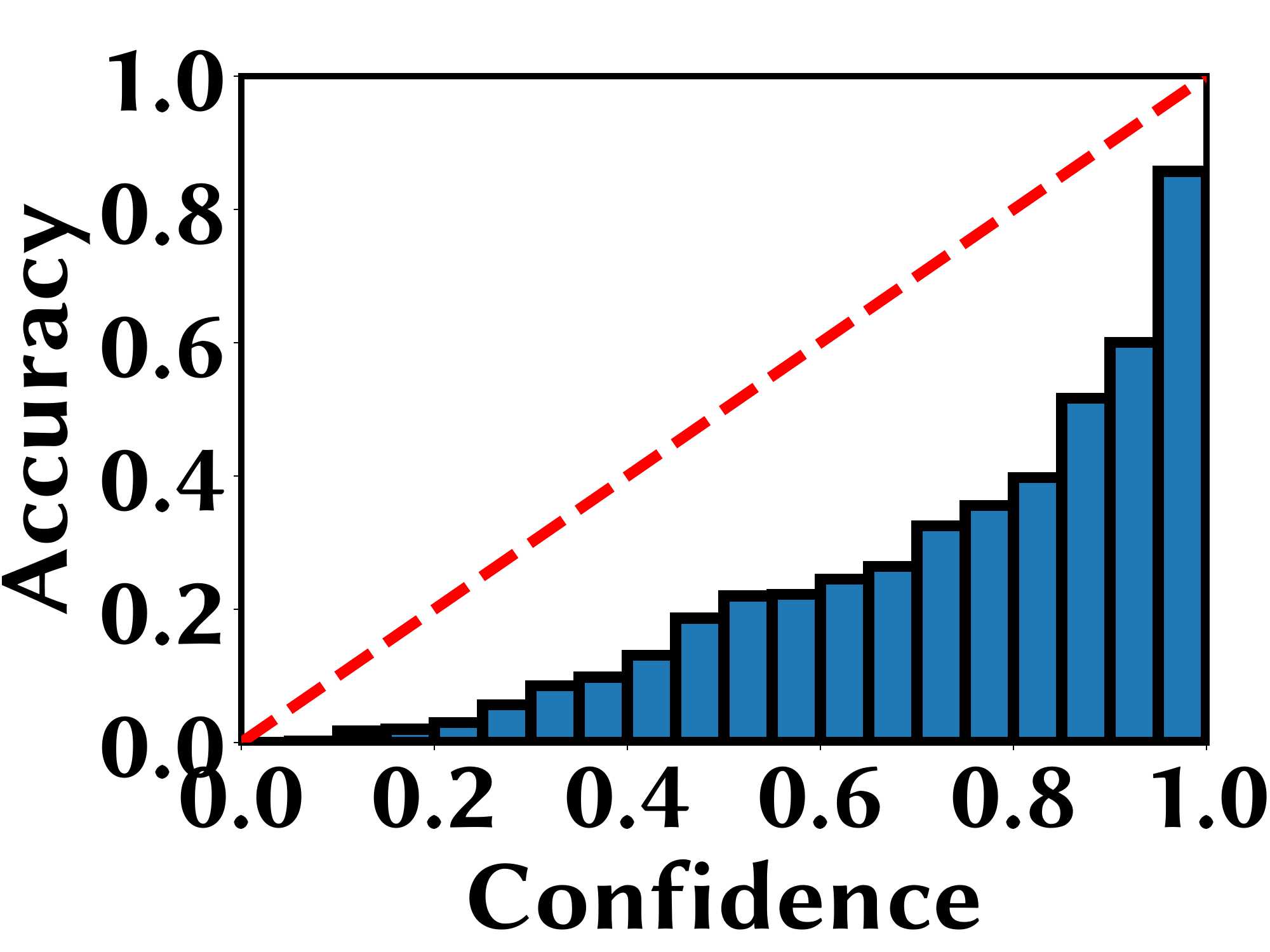}}
	\subfigure[]{\includegraphics[trim=0cm 0cm 0cm 0cm,clip,  width=0.24\textwidth]{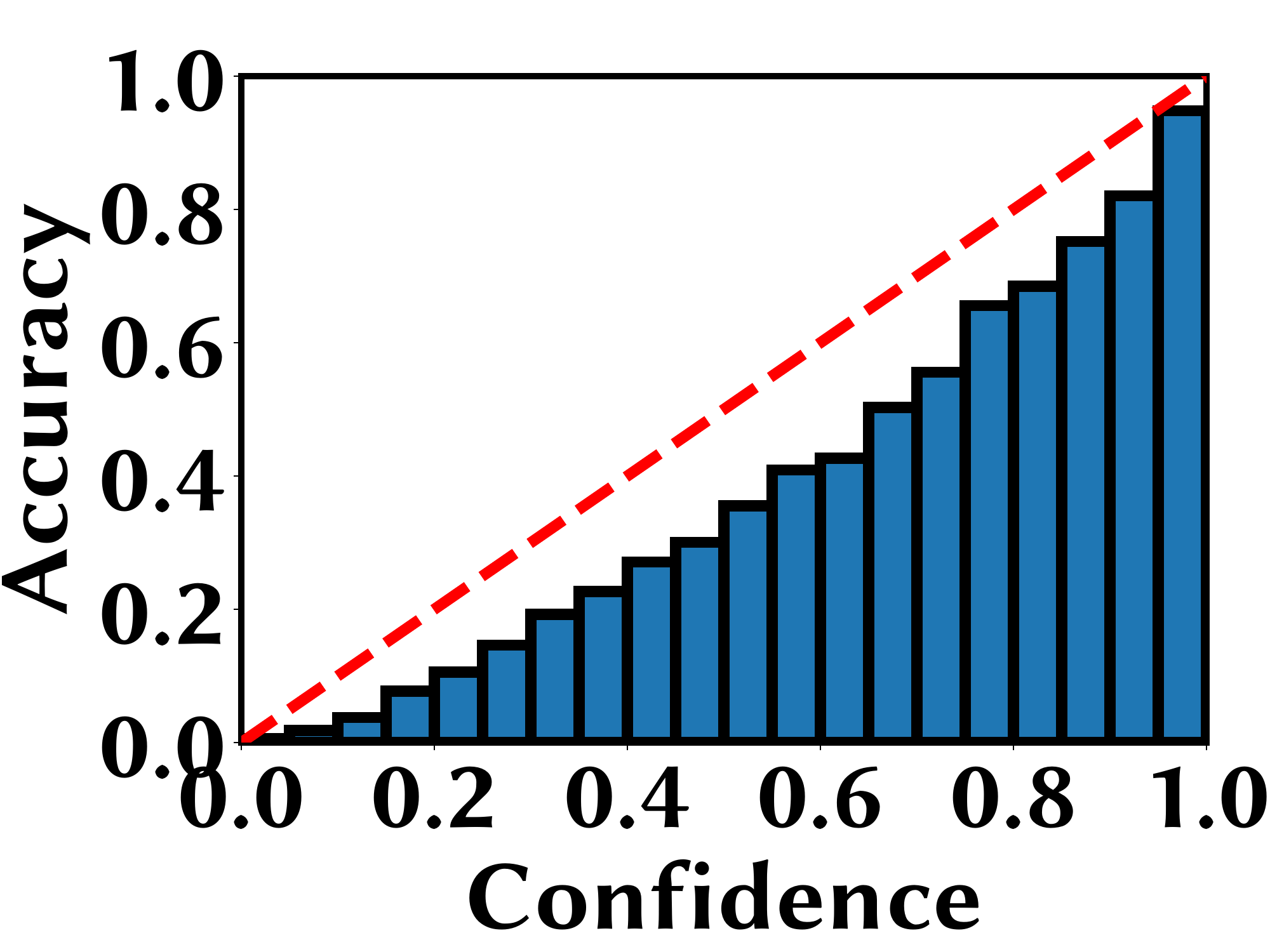}}
	\subfigure[]{\includegraphics[trim=0cm 0cm 0cm 0cm,clip,  width=0.24\textwidth]{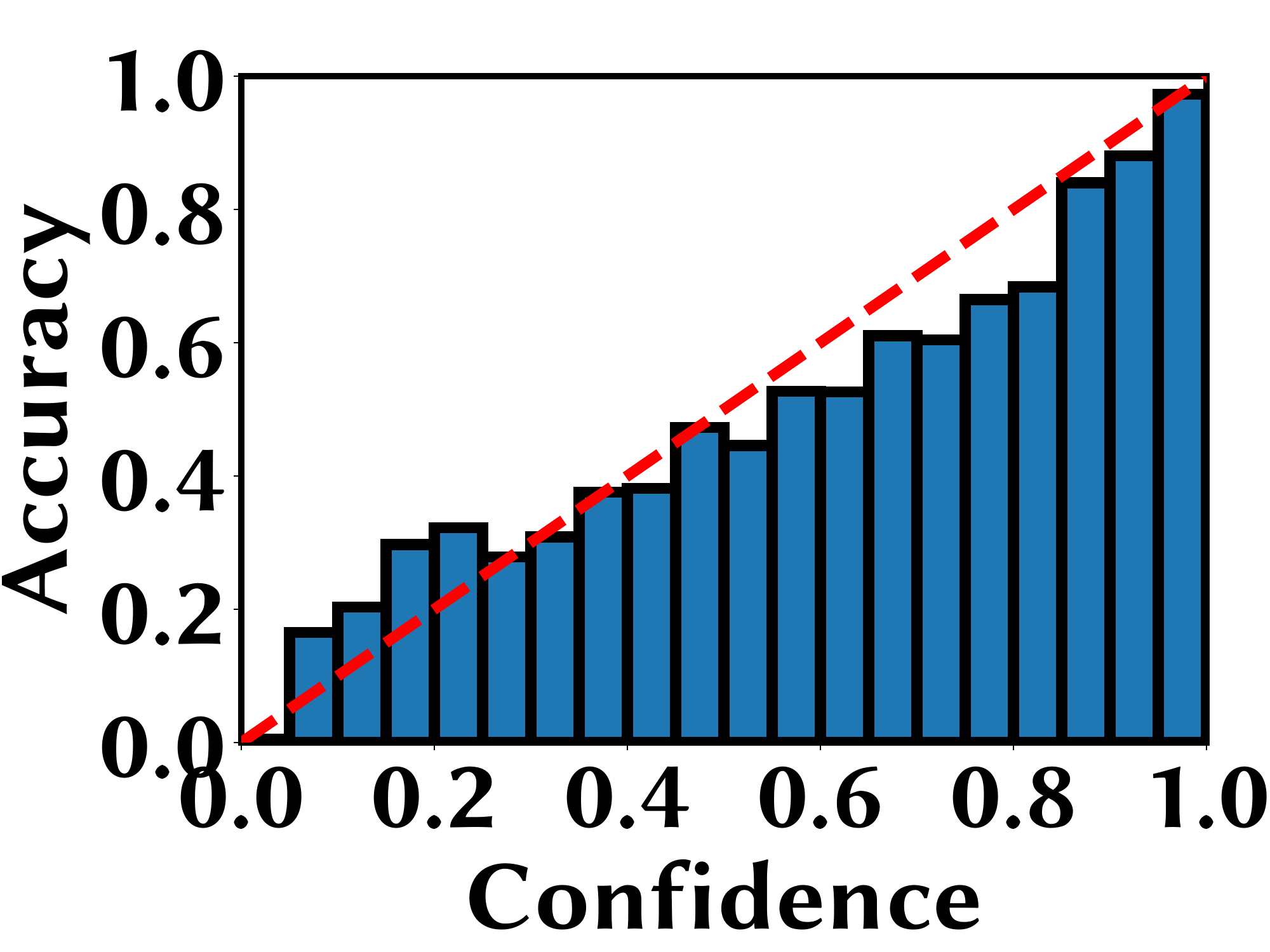}}
	\vspace{-0.3cm}
	\caption{Reliability diagrams of prediction confidences for ResNet-50 on ImageNet OOD dataset. (a) Un-calibrated. (b) Temperature scaling. (c) \advanceModel.}\label{fig:imagenet_resnet_OOD_ECE}
	\vspace{-0.3cm}
\end{figure*}

\textbf{Target DNN model. } The target DNN model is ResNet-50 for 1000-class image classification. The pre-trained weights of target DNN model are directly downloaded from Keras application package \cite{keras_2015}, which are trained on ImageNet 2012 training dataset \cite{imagenet_dataset_2012_IJCV}.

All the other settings, such as the datasets and architecture
of the neural network in \ouralgtwo, are the same as those for the 1000-class VGG16 DNN
described in Section~\ref{sec:1000class_vgg}. The inference accuracies on the two test datasets are 64\% (D1) and 34\% (OOD).

\textbf{Results. } The calibration results are presented in Table~\ref{table:imagenet_resnet_ECE_D1} and \ref{table:imagenet_resnet_ECE_OOD} for D1 and OOD datasets, respectively. The results show that the proposed method (\advanceModel) outperforms the baselines in terms of confidence calibration, with significant improvement on the OOD dataset. In addition, the \advanceModel offers a better mis-classification detection performance on OOD dataset in terms of AUROC/AUPR/p.9.
We further show the histograms of confidences for correct and wrong predictions in Fig.~\ref{fig:imagenet_resnet_OOD_histogram} on the OOD dataset. The corresponding reliability diagrams also are shown in Fig.~\ref{fig:imagenet_resnet_OOD_ECE} on the OOD dataset with different calibration methods.

\subsection{Results for Document Classification}

We consider two document classification applications: Reuters 8-topic classification \cite{reuters_lewis_1987} and 20 Newsgroups classification \cite{rennie_20_Newsgroups_2008}.

\subsubsection{8-topic Document Classification with DAN}

\textbf{Pre-processing.} For the Reuters dataset, news articles are partitioned into 8 categories (R8 dataset). In our experiments, the raw documents are pre-processed with the same \textit{tokenizer} and re-formatted into sequences each with 1000 words via \textit{pad\_sequences}, which
are then used as inputs to the target DNN model.

\textbf{Target DNN model. } The target DNN model is a Deep Average Network (DAN), trained in \textit{TensorFlow} on the R8 training dataset (5485 documents). The target DNN model includes 3 blocks with feed-forward layers, batch normalization layers, and dropout layers \cite{DAN_models_text_iyyer_2015}. The DAN model is trained via Adam optimizer with a learning rate of $10^{-3}$ over 50 epochs.

\begin{table}[!t]
	\centering
	
	\setlength{\tabcolsep}{1.5pt}
	\parbox{.48\linewidth}{
		\centering
		\caption{DAN on R8 D1}
		\label{table:r8_in_ECE_D1}
		\begin{tabular}{lccccc}
			\hline
			\textbf{Method} & \textbf{AUROC} & \textbf{AUPR} & \textbf{p.9} & \textbf{ECE} & \textbf{BS}\\
			\hline
			\textbf{MP} & 0.885 & 0.484 & 0.236 & 5.3\% & 0.065 \\
			\textbf{TS} & 0.884 & 0.470 & 0.238 & 2.9\% & 0.064 \\
			\textbf{SB} & 0.938 & 0.691 & 0.410 & 3.2\% & 0.049 \\
			\textbf{Dirichlet} & 0.816 & 0.338 & 0.150 & 2.8\% & 0.078 \\
			\textbf{\basicModel} & \textbf{0.954} & \textbf{0.729} & 0.437 & \textbf{1.9\%} & \textbf{0.044} \\
			\textbf{\advanceModel} & 0.947 & 0.722 & \textbf{0.452} & 2.2\% & 0.047 \\
			\textbf{\transferMethod} & -     & -     & -     & -     & - \\
			\hline
		\end{tabular}

	}
	\hfill
	\parbox{.48\linewidth}{
		\centering
		\caption{DAN on R8 OOD}
		\label{table:r8_in_ECE_OOD}
		\begin{tabular}{lccccc}
			\hline
			\textbf{Method} & \textbf{AUROC} & \textbf{AUPR} & \textbf{p.9} & \textbf{ECE} & \textbf{BS} \\
			\hline
			\textbf{MP} & 0.801 & 0.646 & 0.415 & 10.3\% & 0.171 \\
			\textbf{TS} & 0.799 & 0.643 & 0.413 & 5.6\% & 0.180 \\
			\textbf{SB} & 0.889 & 0.772 & 0.587 & 7.8\% & 0.129 \\
			\textbf{Dirichlet} & 0.717 & 0.512 & 0.419 & 5.0\% & 0.192 \\
			\textbf{\basicModel} & \textbf{0.929} & \textbf{0.860} & \textbf{0.672} & 3.5\% & \textbf{0.099} \\
			\textbf{\advanceModel} & 0.919 & 0.850 & 0.640 & \textbf{3.1\%} & 0.102 \\
			\textbf{\transferMethod} & 0.885 & 0.763 & 0.604 & 4.8\% & 0.126 \\
			\hline
		\end{tabular}
	}
\end{table}

\begin{figure*}[!t]
	\centering
	\subfigure[]{\includegraphics[trim=0cm 0cm 0cm 0cm,clip,  width=0.195\textwidth]{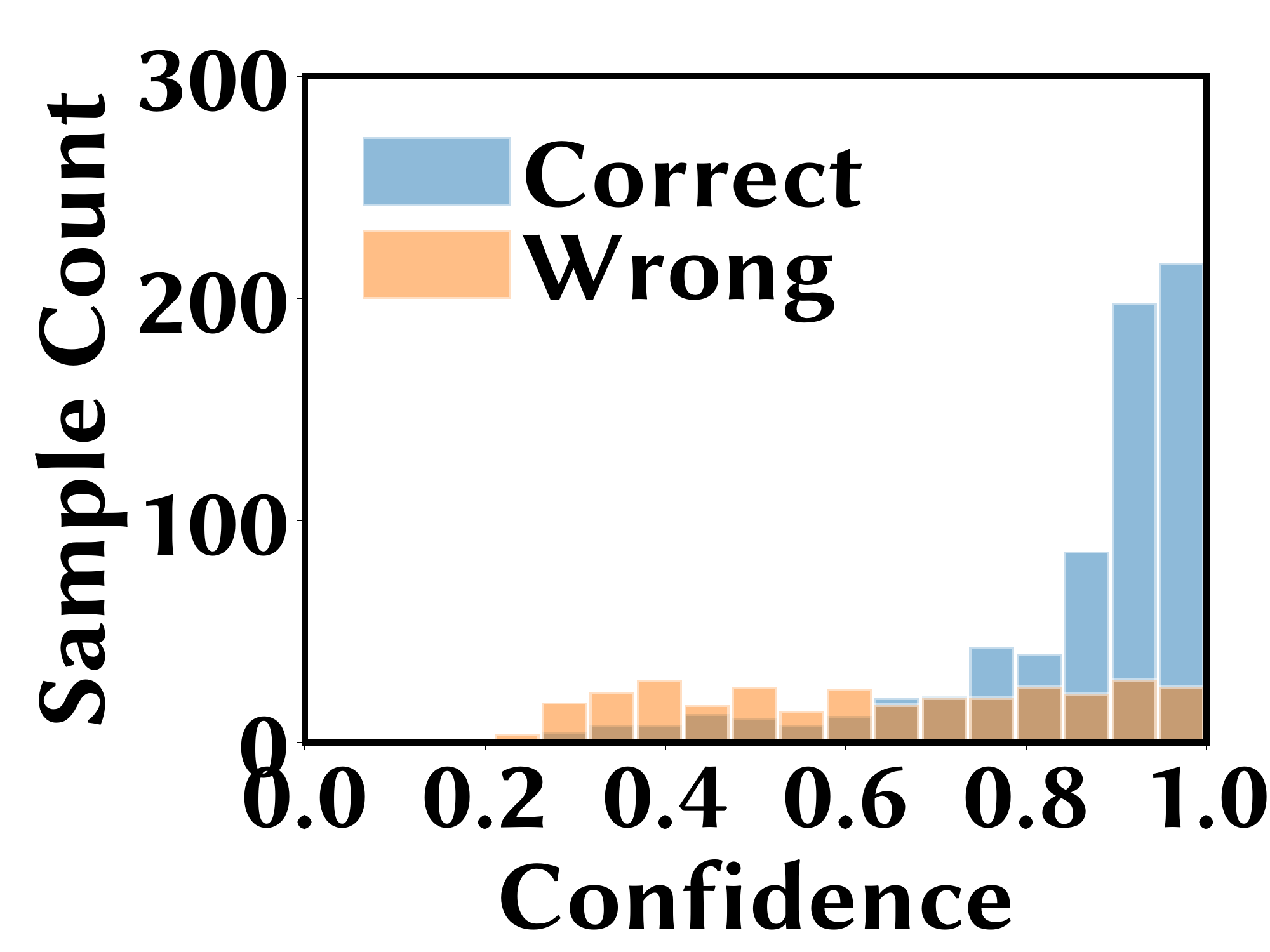}}
	\subfigure[]{\includegraphics[trim=0cm 0cm 0cm 0cm,clip,  width=0.195\textwidth]{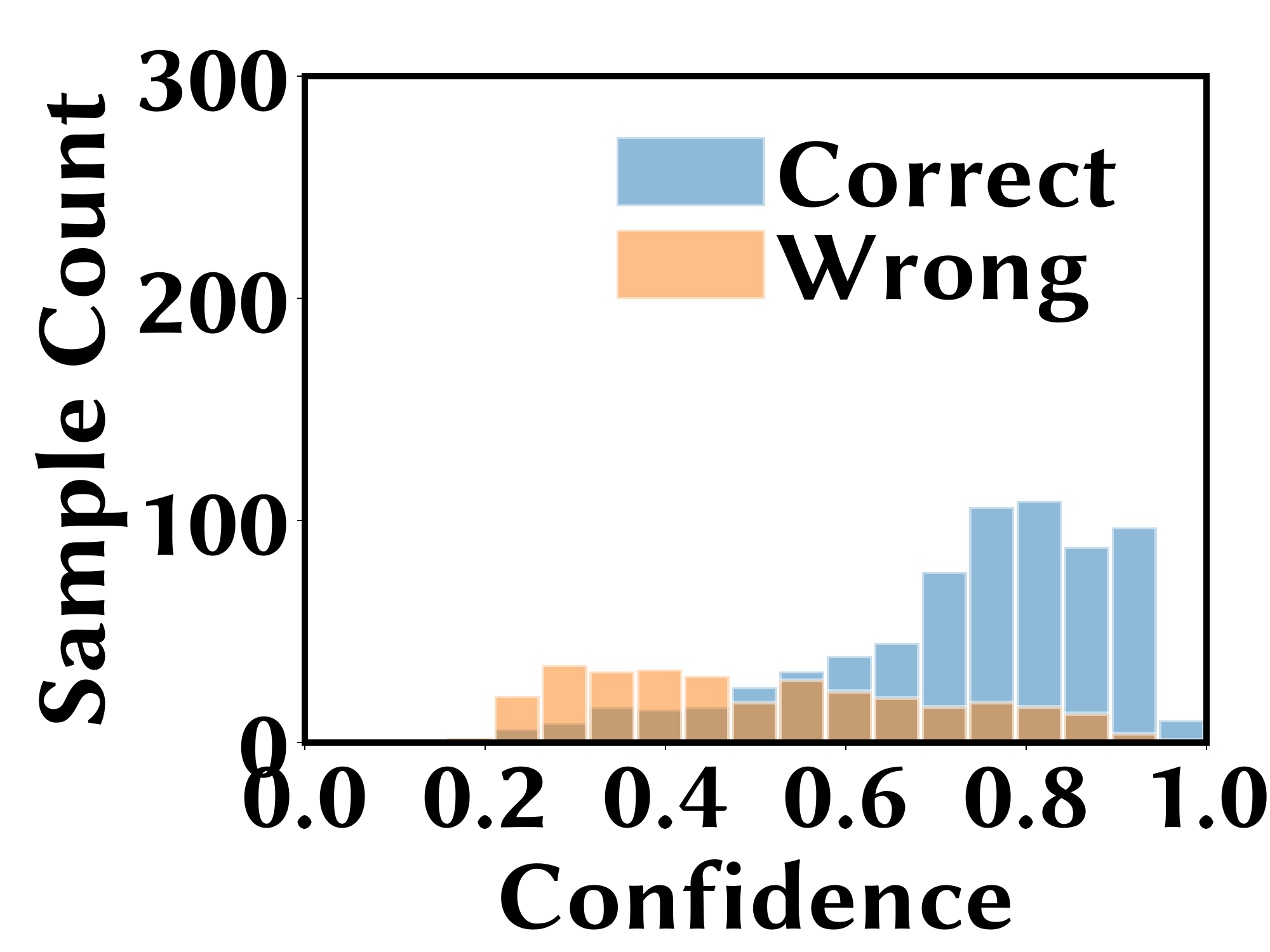}}
	\subfigure[]{\includegraphics[trim=0cm 0cm 0cm 0cm,clip,  width=0.195\textwidth]{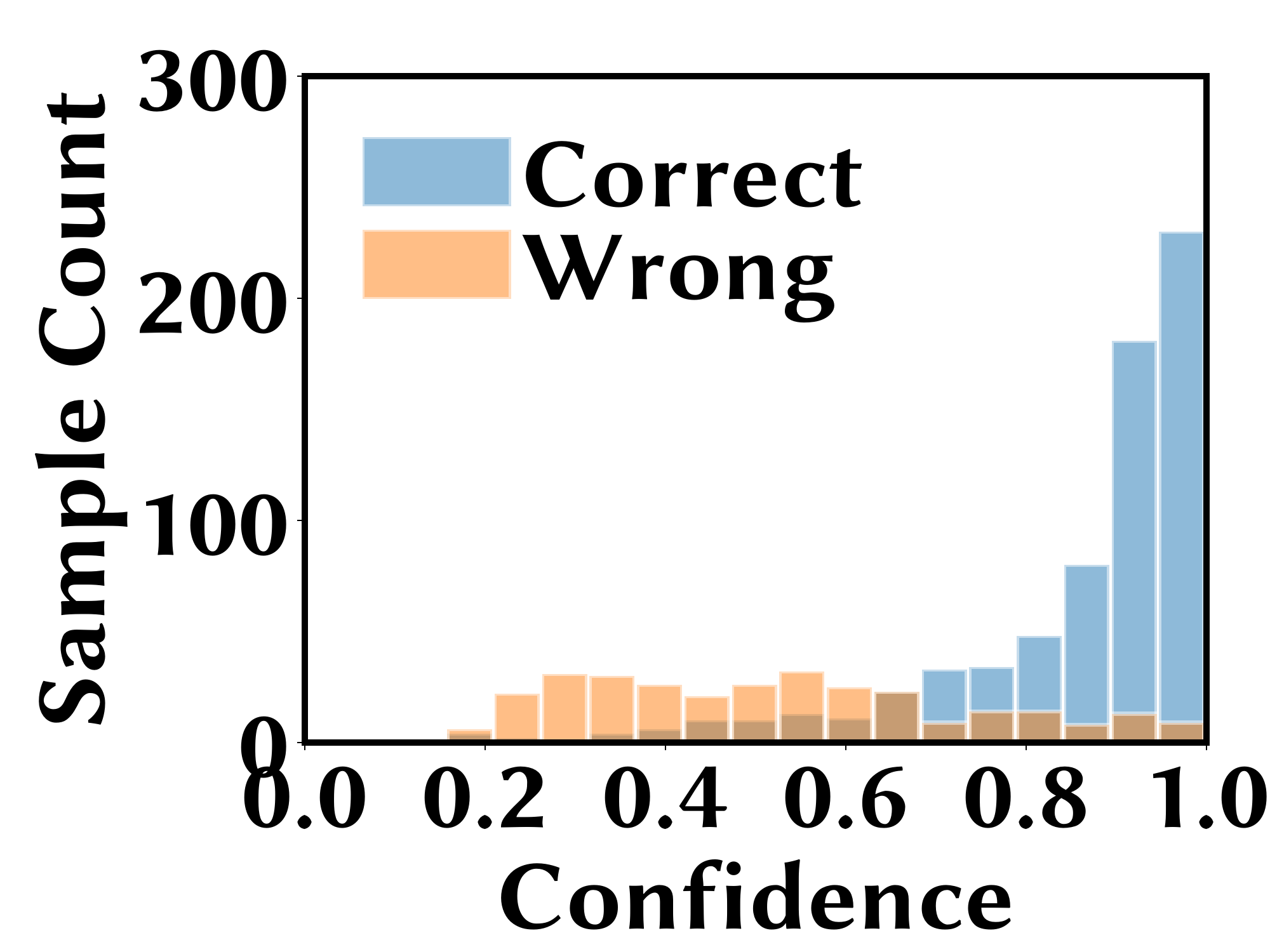}}
	\subfigure[]{\includegraphics[trim=0cm 0cm 0cm 0cm,clip,  width=0.195\textwidth]{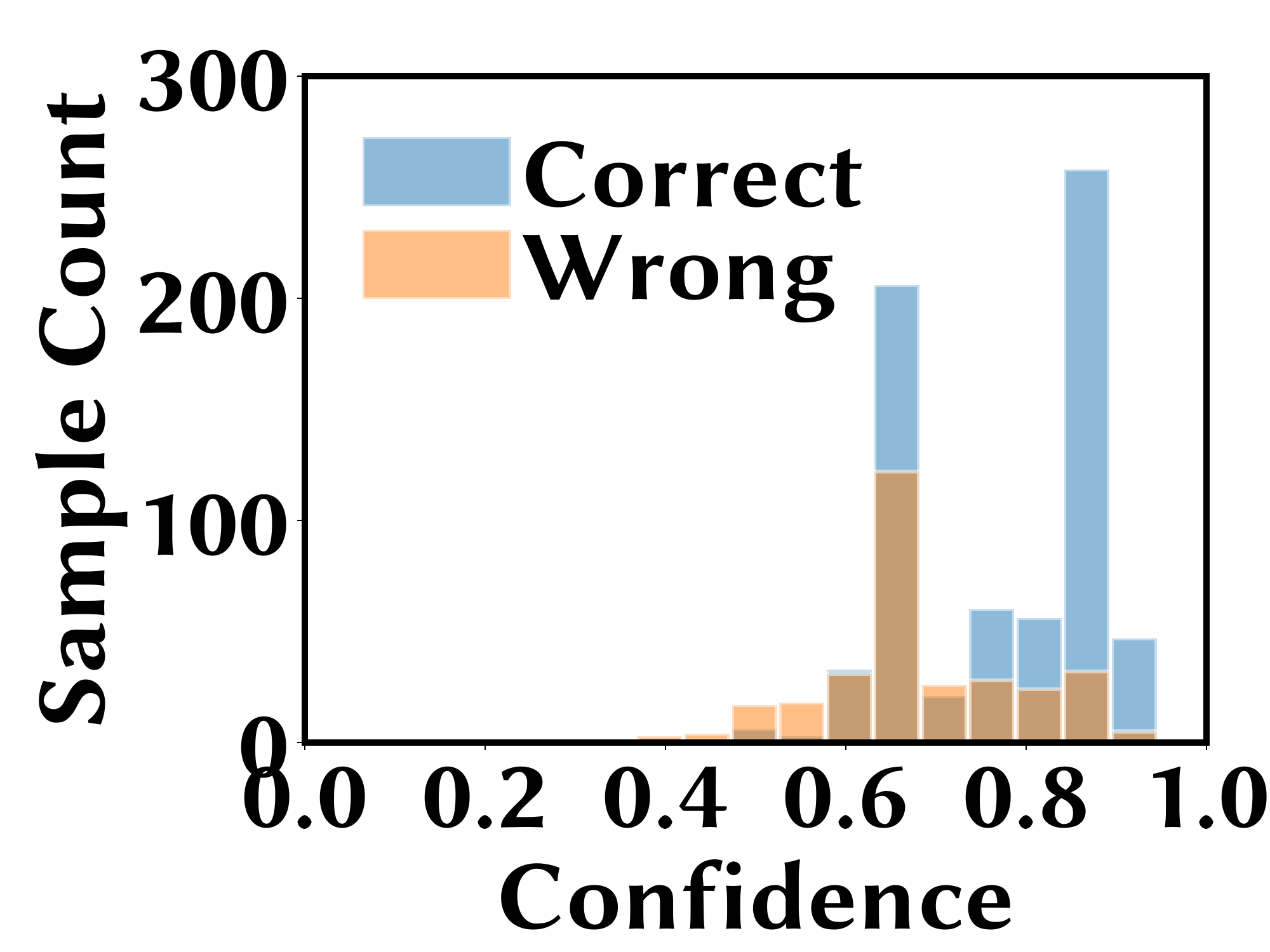}}
	\subfigure[]{\includegraphics[trim=0cm 0cm 0cm 0cm,clip,  width=0.195\textwidth]{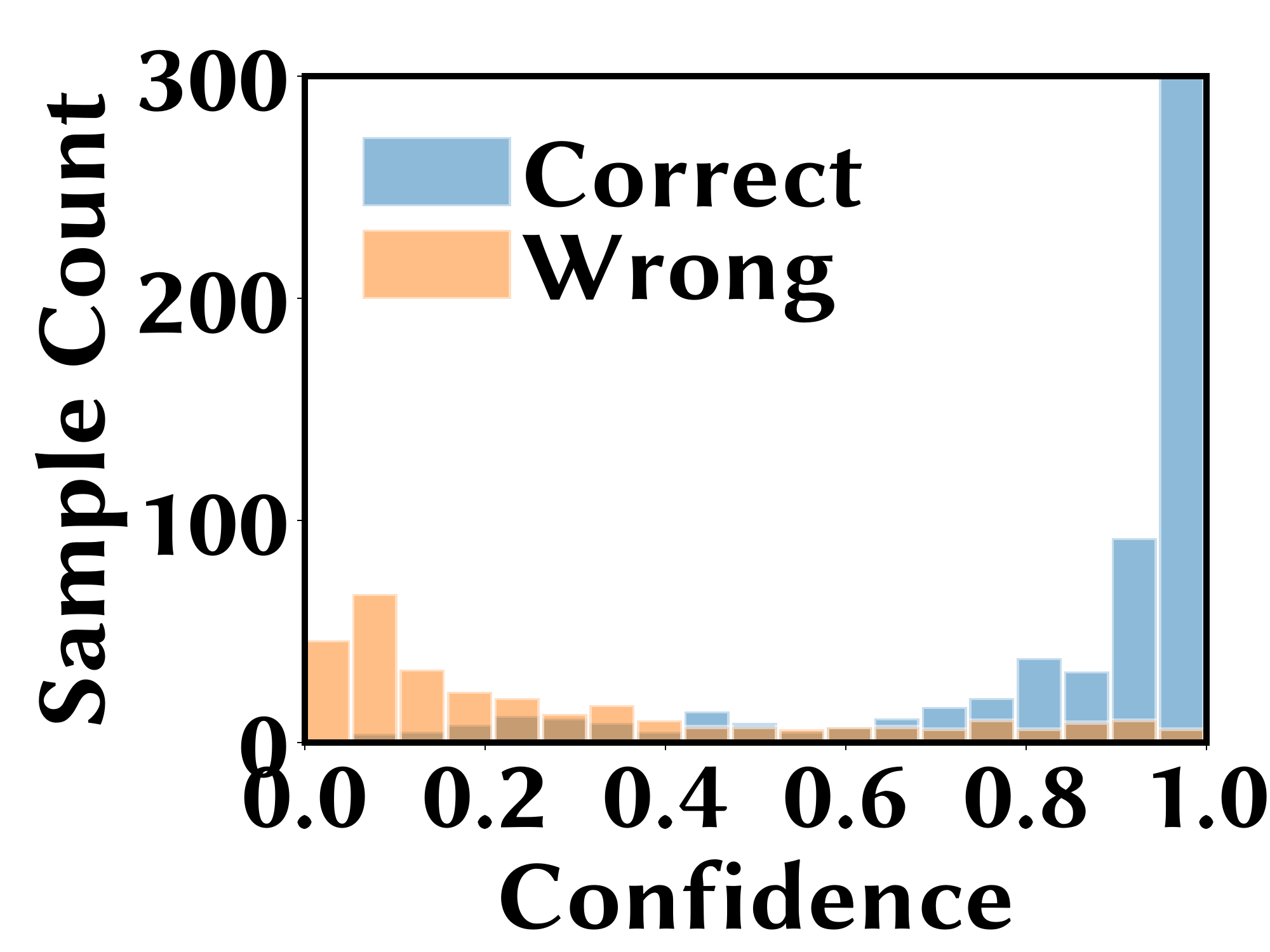}}
	\vspace{-0.3cm}
	\caption{Histograms of prediction confidences for DAN on R8 OOD dataset. (a) Un-calibrated. (b) Temperature scaling. (c) Scaling-binning. (d) Dirichlet calibration. (e) \ouralg.}\label{fig:r8_OOD_histogram}
	\vspace{-0.3cm}
\end{figure*}

\begin{figure*}[!t]
	\centering
	\subfigure[]{\includegraphics[trim=0cm 0cm 0cm 0cm,clip,  width=0.195\textwidth]{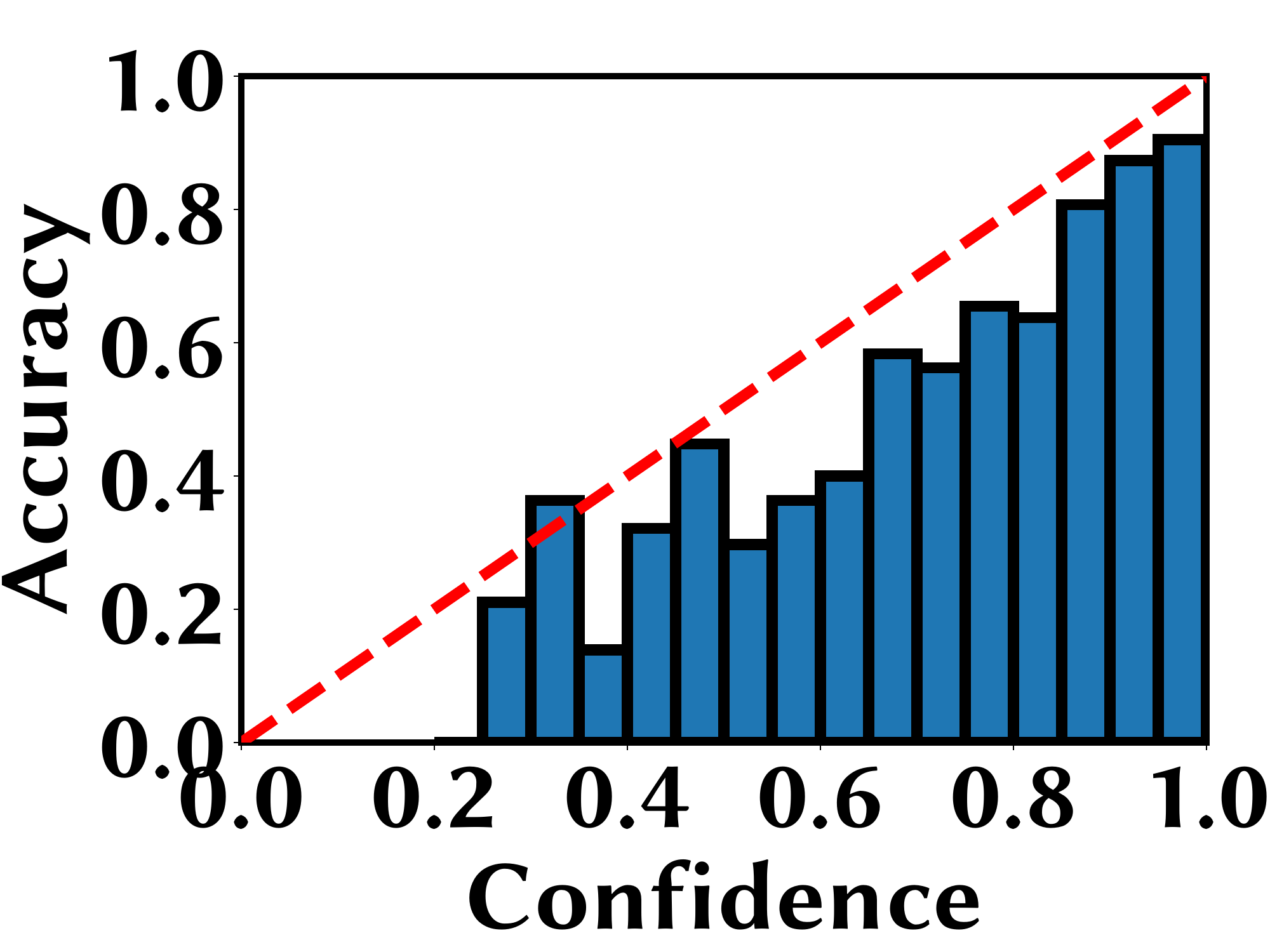}}
	\subfigure[]{\includegraphics[trim=0cm 0cm 0cm 0cm,clip,  width=0.195\textwidth]{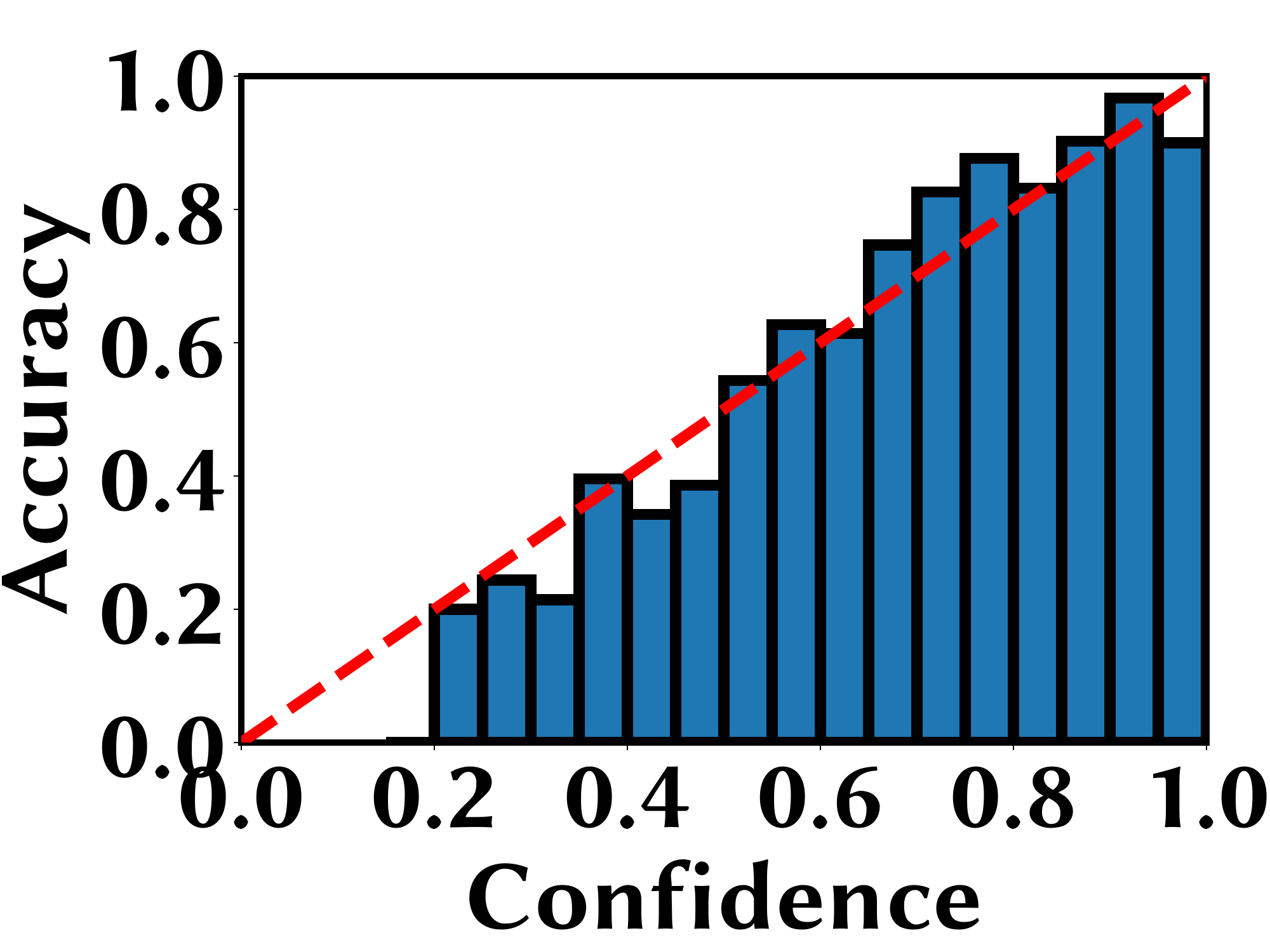}}
	\subfigure[]{\includegraphics[trim=0cm 0cm 0cm 0cm,clip,  width=0.195\textwidth]{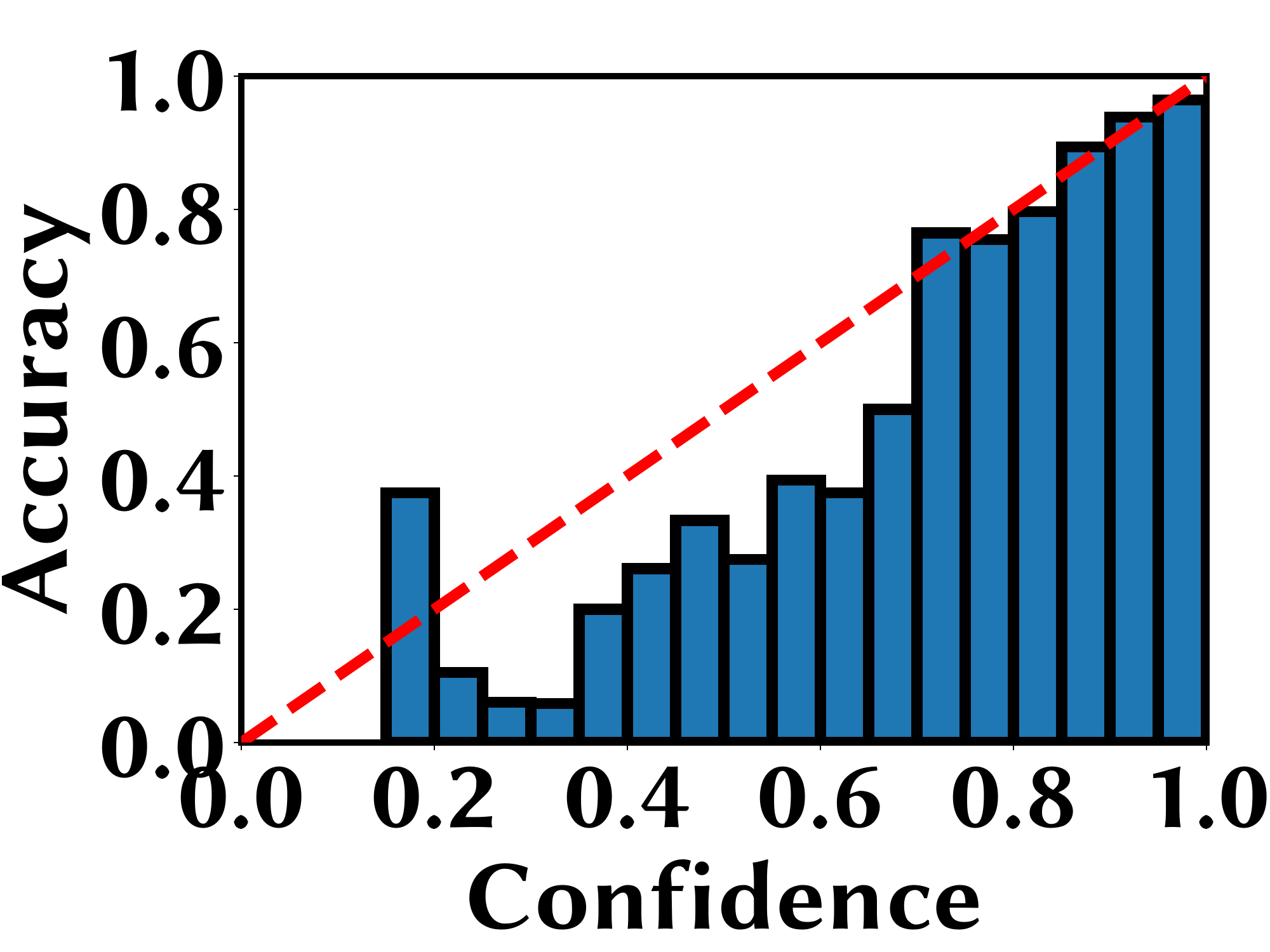}}
	\subfigure[]{\includegraphics[trim=0cm 0cm 0cm 0cm,clip,  width=0.195\textwidth]{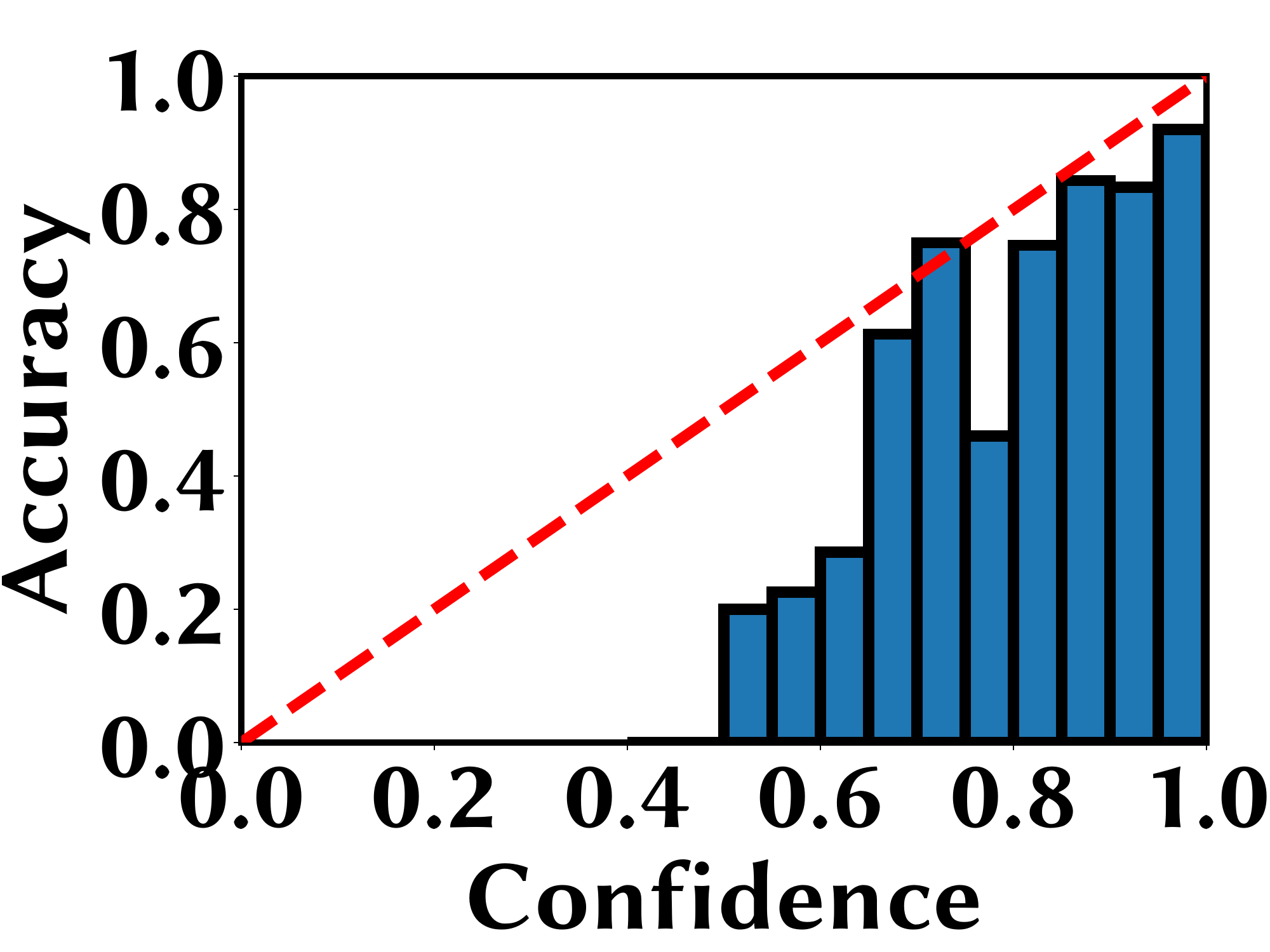}}
	\subfigure[]{\includegraphics[trim=0cm 0cm 0cm 0cm,clip,  width=0.195\textwidth]{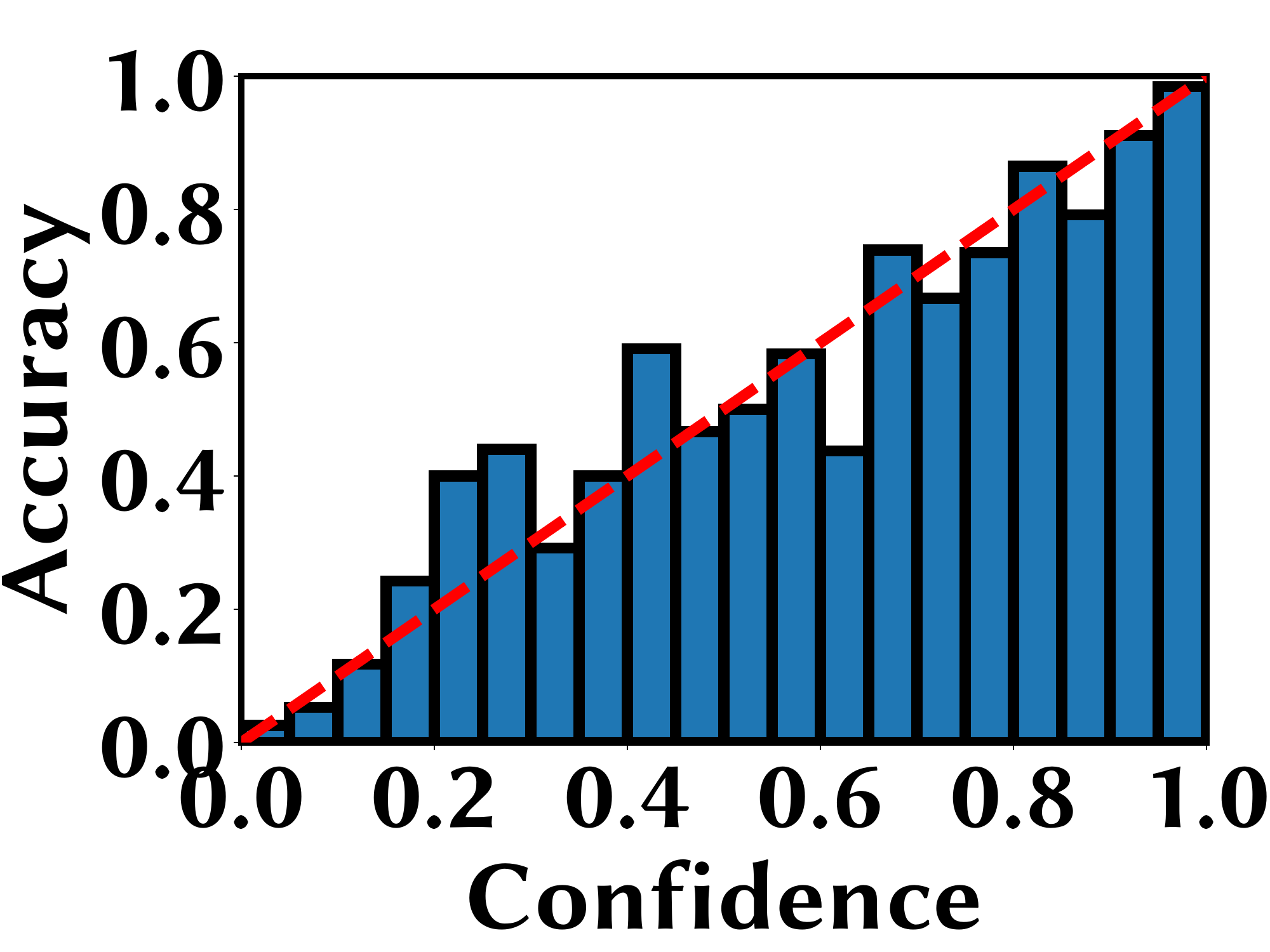}}
	\caption{Reliability diagrams of prediction confidences for DAN on R8 OOD dataset. (a) Un-calibrated. (b) Temperature scaling. (c) Scaling-binning. (d) Dirichlet calibration. (e) \ouralg.}\label{fig:RD_plot_r8_OOD}
	\vspace{-0.3cm}
\end{figure*}

\textbf{Datasets.} We evaluate the calibration performance on two datasets generated from the R8 dataset and Reuters-extension dataset (R52), including augmented dataset D1 and Out-of-Distribution dataset (OOD). The R52 dataset contains 1426 documents with different news topics. We first randomly select 2811 articles from R8 training dataset and 2189 articles from R8 testing dataset, and then perform augmentation operations on selected samples with the \textit{nlpaug} package \cite{nlp_augmentation_ma_2019}. The augmentation operations include the OCR engine error (OcrAug) operation and the randomly typo (RandomAug) operation with \textit{substitute}, \textit{swap} and \textit{delete}. The OOD dataset also includes 5k samples, with 1246 OOD samples from R52 dataset, 1385 samples from R8 training dataset, and 2189 samples from R8 testing datasest. As for the 1246 OOD samples from R52, the true labels are mapped to R8's class labels. Specifically, we treat the OOD samples with label ``income'' and ``jobs'' as the ``earn'' category in R8, and the OOD samples with label ``money-supply'' as the ``money-fx'' category in R8. The other OOD samples from R52 are treated with the ``NULL'' label, indicating not belonging to any of the 8 categories in R8. The inference accuracies on two test datasets are 90\% (D1) and 69\% (OOD). Additionally, the 5k samples (D1 and OOD) are randomly split into a 3k dataset for  training, a 0.5k validation dataset for hyperparameter tuning, and a 1.5k testing dataset for performance evaluation.

\textbf{Baselines.} For each dataset, the baselines of TS, SB, and Dirichlet calibration are trained on the respective 3k training dataset. For Dirichlet calibration, the regularization hyperparameters are tuned with a minimal ECE on the 0.5k validation dataset. The 3k training and 0.5k validation dataset are  the same as those used for training \ouralg and \ouralgtwo.

\textbf{Our method.} The neural network of \ouralg is implemented with 2 hidden layers, including 10 neurons in each hidden layer. The input layer contains 8 nodes and the output layer contains 9 nodes (including 8 classes and one ``mis-classification'' class). \ouralg is trained over 1500 epochs using the Adam optimizer (learning rate $0.002$). The loss function hyperparameters ($\lambda_1$ and $\lambda_2$) and the confidence calculation method are selected with a minimal ECE on the validation dataset. For \ouralgtwo, the neural network is implemented with one hidden layer, including 20 hidden nodes. The output layer of \ouralgtwo contains only one node, representing the ``mis-classification'' class. Also, the temperature scaling layer is implemented with a self-defined Keras layer with one learnable weight $T$. The \ouralgtwo is trained over 1500 epochs via Adam optimizer with a learning rate of $0.005$. For the transferred model \transferMethod, we first pre-train \advanceModel on dataset D1. Then, when applied to another target dataset, a set of 320 samples are randomly selected from the 3k training dataset for model transfer. In addition, we select another 200 samples from the 0.5k training  dataset to tune the corresponding hyperparameters and confidence calculation method in \ouralgthree.

\textbf{Results. } The calibration results are presented in Table~\ref{table:r8_in_ECE_D1}, \ref{table:r8_in_ECE_OOD} for D1 and OOD datasets, respectively. Like
for image classification applications, the results show that the proposed methods (\basicModel and \advanceModel) outperform the baselines in terms of both mis-classification detection and confidence calibration, with more significant improvement on the OOD dataset. In addition, even though the \transferMethod model is transferred to the OOD dataset with fewer training samples than the baselines, it still offers a better calibration performance than baselines in terms of ECE and BS.
We further show the histograms of confidences for correct and wrong predictions in Fig.~\ref{fig:r8_OOD_histogram} on the OOD dataset. The corresponding reliability diagrams also are shown in Fig.~\ref{fig:RD_plot_r8_OOD} on the OOD dataset with different calibration methods.

\subsubsection{20-topic Document Classification with DAN}

\textbf{Pre-processing.}
 The 20 Newsgroups dataset includes approximately 20k documents, which are partitioned into 20 different categories \cite{rennie_20_Newsgroups_2008}. We use the 20 Newsgroups dataset provided by \textit{scikit-learn} \cite{scikit-learn_2011}, including 11314 documents in the training dataset and 7532 documents in the testing dataset. In our experiment, we only use 18 categories from the 20 Newsgroups dataset, excluding category \textit{``comp.sys.mac.hardware''} and category \textit{``rec.sport.hockey''}. We denote
 the selected 18-category Newsgroups group dataset as NG18 and the remaining 2 categories as dataset NG2. The samples in NG2 will be treated as OOD samples for the target DNN model. The raw documents are pre-processed with  \textit{tokenizer} and re-formatted into sequences each with 1000 words via \textit{pad\_sequences}, which
 are then used inputs to the target DNN model.

\textbf{Target DNN model.} The target DNN model is a Deep Average Network (DAN), trained in \textit{TensorFlow} on the NG18 training dataset with 18 categories (10136 documents). The target DNN model includes a pre-trained embedding layer downloaded from \textit{glove.6B} \cite{glove_embedding_pennington_EMNLP_2014}, 3 convolution layers and 2 fully connected layers. The target DNN model is trained via Adam optimizer with a learning rate of $10^{-3}$ over 50 epochs.

\textbf{Datasets.} We evaluate the calibration performance on two datasets generated from
the 20 Newsgroups Dataset and Reuters-8 datasets (R8), including augmented dataset D1 and Out-of-Distribution dataset (OOD). We first randomly select 3252 documents from the NG18 training dataset and 6748 documents from NG18 testing dataset, and then perform augmentation operations on the selected samples with  \textit{nlpaug} \cite{nlp_augmentation_ma_2019}. The augmentation operations include the OCR engine error (OcrAug) operation and the randomly typo (RandomAug) operation with \textit{substitute},\textit{swap} and \textit{delete} actions. The OOD dataset also includes 10k samples, with 2553 documents randomly selected from the NG18 training dataset, 1962 documents from NG2 dataset, and 5485 documents from R8 dataset. Here, both documents in NG2 and R8 are considered as OOD samples. As for the OOD samples, the true labels are mapped to NG18's class labels. Specifically, we consider the OOD samples from NG2 with label
\textit{``comp.sys.mac.hardware''} as class \textit{``comp.sys.ibm.pc.hardware''} and \textit{``rec.sport.hockey''} as class \textit{``rec.sport.baseball''} in NG18 dataset. The other OOD samples from R8 are treated with the ``NULL'' label, indicating not belonging to any of the categories in NG18. The inference accuracies on the two test datasets are 75\% (D1) and 33\% (OOD). Additionally, the 10k samples (D1 and OOD) are randomly split into a 6k dataset for \method training, a 1k validation dataset for hyperparameter tuning, and a 3k testing dataset for performance evaluation.

\textbf{Our method.} The neural network of \ouralg is implemented with one hidden layer, including 20 hidden neurons. The input layer contains 18 nodes and the output layer contains 19 nodes (including 18 classes and one ``mis-classification'' class).  \ouralg is trained for 2000 epochs using the Adam optimizer (learning rate $10^{-4}$). The loss function hyperparameters ($\lambda_1$ and $\lambda_2$) and the confidence calculation method are selected with a minimal ECE on the validation dataset. For \ouralgtwo, the neural network is implemented with the same structure as \basicModel, but the output layer contains only one node, representing the ``mis-classification'' class. The temperature scaling layer is implemented with a self-defined customer Keras layer with one learnable weight $T$. \ouralgtwo is trained over 1000 epochs using the Adam optimizer (learning rate $10^{-3}$). For the transferred model \transferMethod, we first pre-train \advanceModel on dataset D1. Then, when applied to OOD dataset, a set of 320 samples is randomly selected from the 6k training dataset for model transfer. In addition, we select another 200 samples from the 1k validation dataset to tune the corresponding hyperparameters and confidence calculation method in \ouralgthree.

\begin{table}[!t]
	\centering
	
	\setlength{\tabcolsep}{1.5pt}
	\parbox{.48\linewidth}{
		\centering
		\caption{DAN on 20 Newsgroups D1}
		\label{table:ng20_in_ECE_D1}
		\begin{tabular}{lccccc}
			\hline
			\textbf{Method} & \textbf{AUROC} & \textbf{AUPR} & \textbf{p.9} & \textbf{ECE} & \textbf{BS}\\
			\hline
			\textbf{MP} & 0.826 & 0.580 & 0.403 & 17.4\% & 0.183 \\
			\textbf{TS} & 0.828 & 0.585 & 0.402 & 4.7\% & 0.139 \\
			\textbf{SB} & 0.825 & 0.578 & 0.398 & 4.3\% & 0.140 \\
			\textbf{Dirichlet} & 0.839 & 0.615 & 0.408 & 4.7\% & 0.136 \\
			\textbf{\basicModel} & \textbf{0.865} & \textbf{0.642} & 0.455 & \textbf{2.4\%} & \textbf{0.125} \\
			\textbf{\advanceModel} & 0.849 & 0.606 & \textbf{0.446} & 2.8\% & 0.133 \\
			\textbf{\transferMethod} & -     & -     & -     & -     & - \\
			\hline
		\end{tabular}

	}
	\hfill
	\parbox{.48\linewidth}{
		\centering
		\caption{DAN on 20 Newsgroups OOD}
		\label{table:ng20_in_ECE_OOD}
		\begin{tabular}{lccccc}
			\hline
			\textbf{Method} & \textbf{AUROC} & \textbf{AUPR} & \textbf{p.9} & \textbf{ECE} & \textbf{BS} \\
			\hline
			\textbf{MP} & 0.828 & 0.900 & 0.798 & 40.7\% & 0.349 \\
			\textbf{TS} & 0.855 & 0.917 & 0.815 & 7.5\% & 0.150 \\
			\textbf{SB} & 0.850 & 0.892 & 0.834 & 7.1\% & 0.140 \\
			\textbf{Dirichlet} & 0.704 & 0.833 & 0.695 & 8.6\% & 0.206 \\
			\textbf{\basicModel} & 0.861 & 0.894 & 0.875 & 3.9\% & 0.116 \\
			\textbf{\advanceModel} & \textbf{0.910} & \textbf{0.933} & \textbf{0.906} & \textbf{3.0\%} & \textbf{0.096} \\
			\textbf{\transferMethod} & 0.865 & 0.895 & 0.873 & 4.7\% & 0.118 \\
			\hline
		\end{tabular}
	}
\end{table}

\begin{figure*}[!t]
	\centering
	\subfigure[]{\includegraphics[trim=0cm 0cm 0cm 0cm,clip,  width=0.195\textwidth]{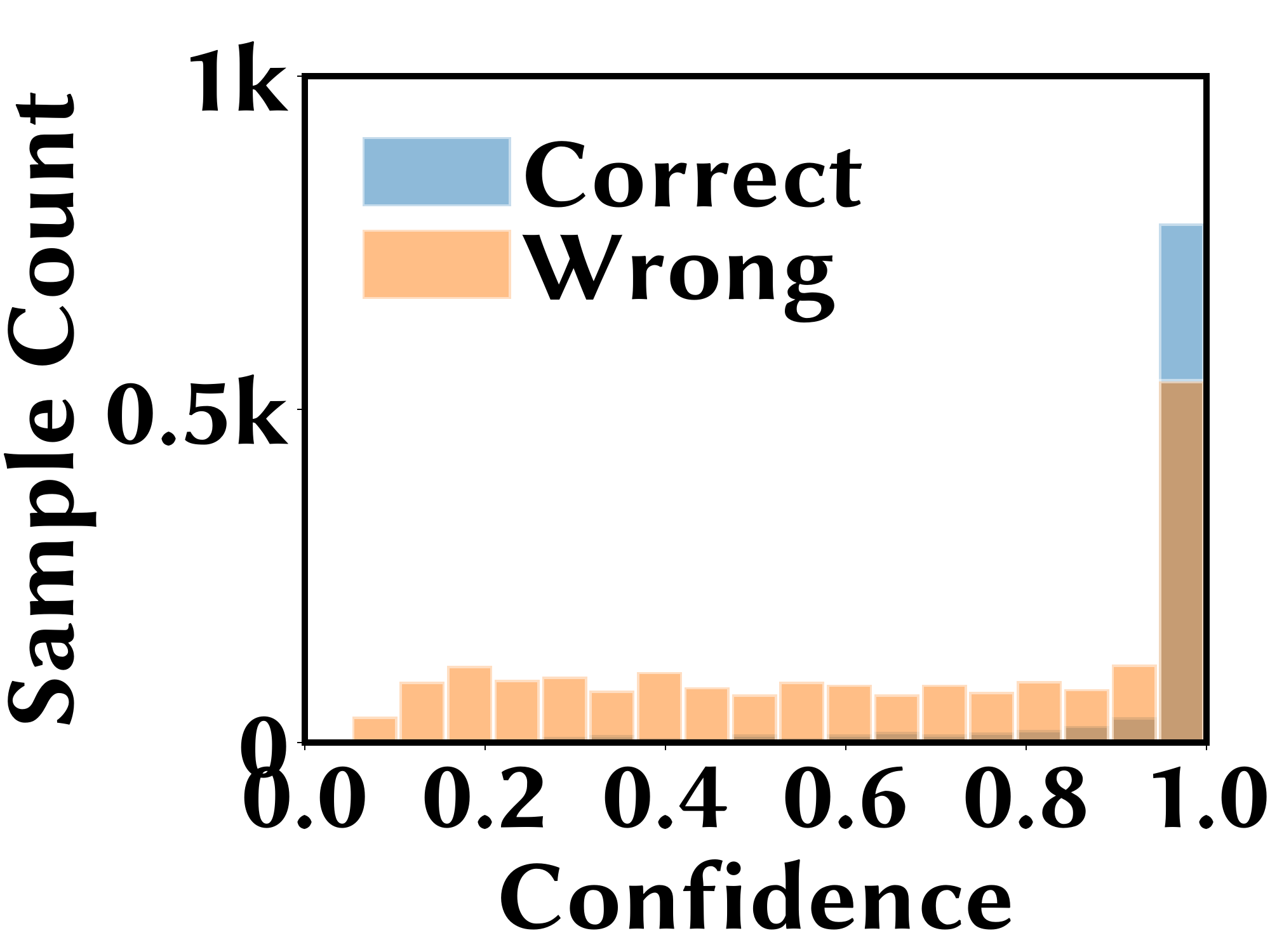}}
	\subfigure[]{\includegraphics[trim=0cm 0cm 0cm 0cm,clip,  width=0.195\textwidth]{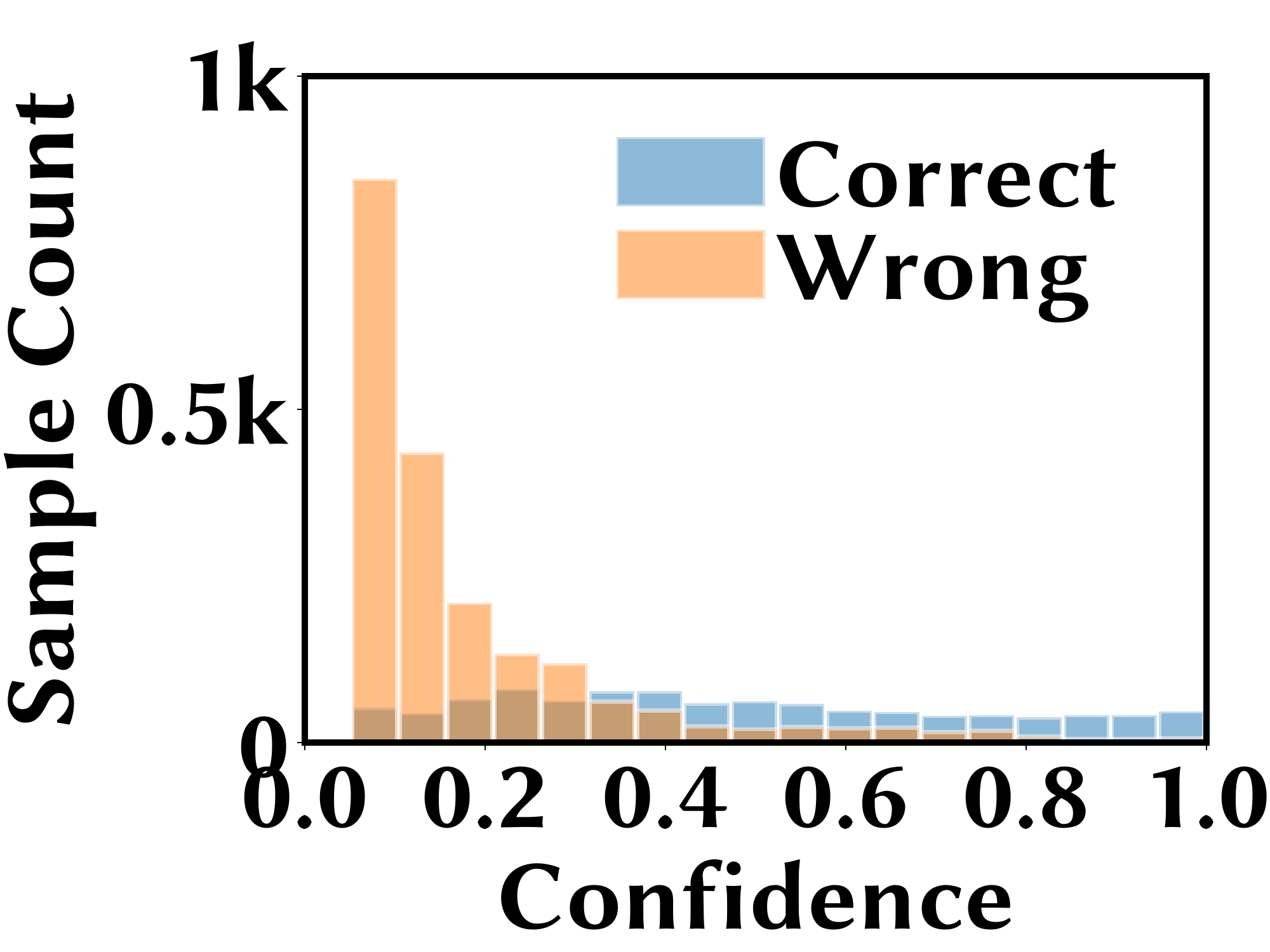}}
	\subfigure[]{\includegraphics[trim=0cm 0cm 0cm 0cm,clip,  width=0.195\textwidth]{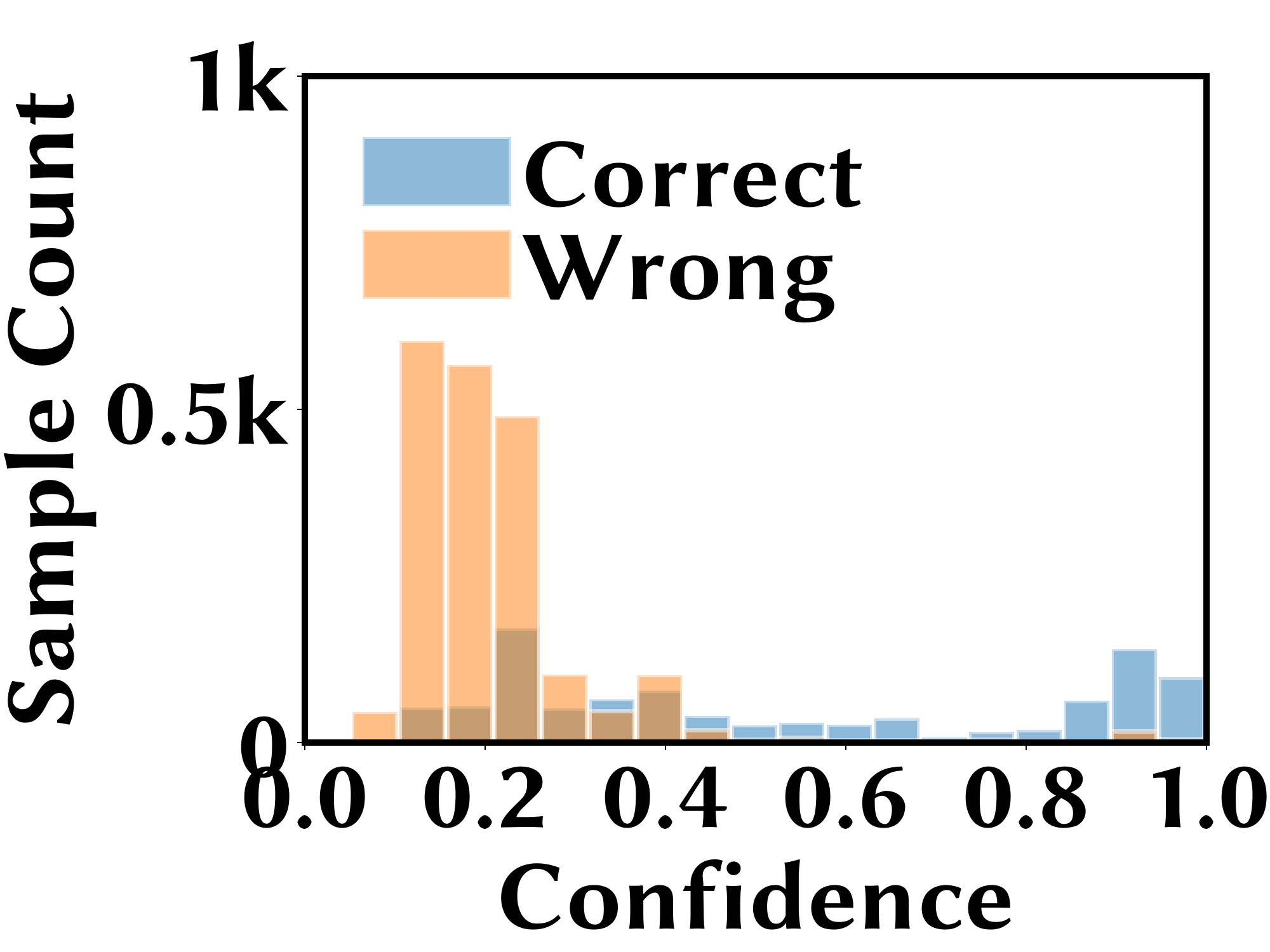}}
	\subfigure[]{\includegraphics[trim=0cm 0cm 0cm 0cm,clip,  width=0.195\textwidth]{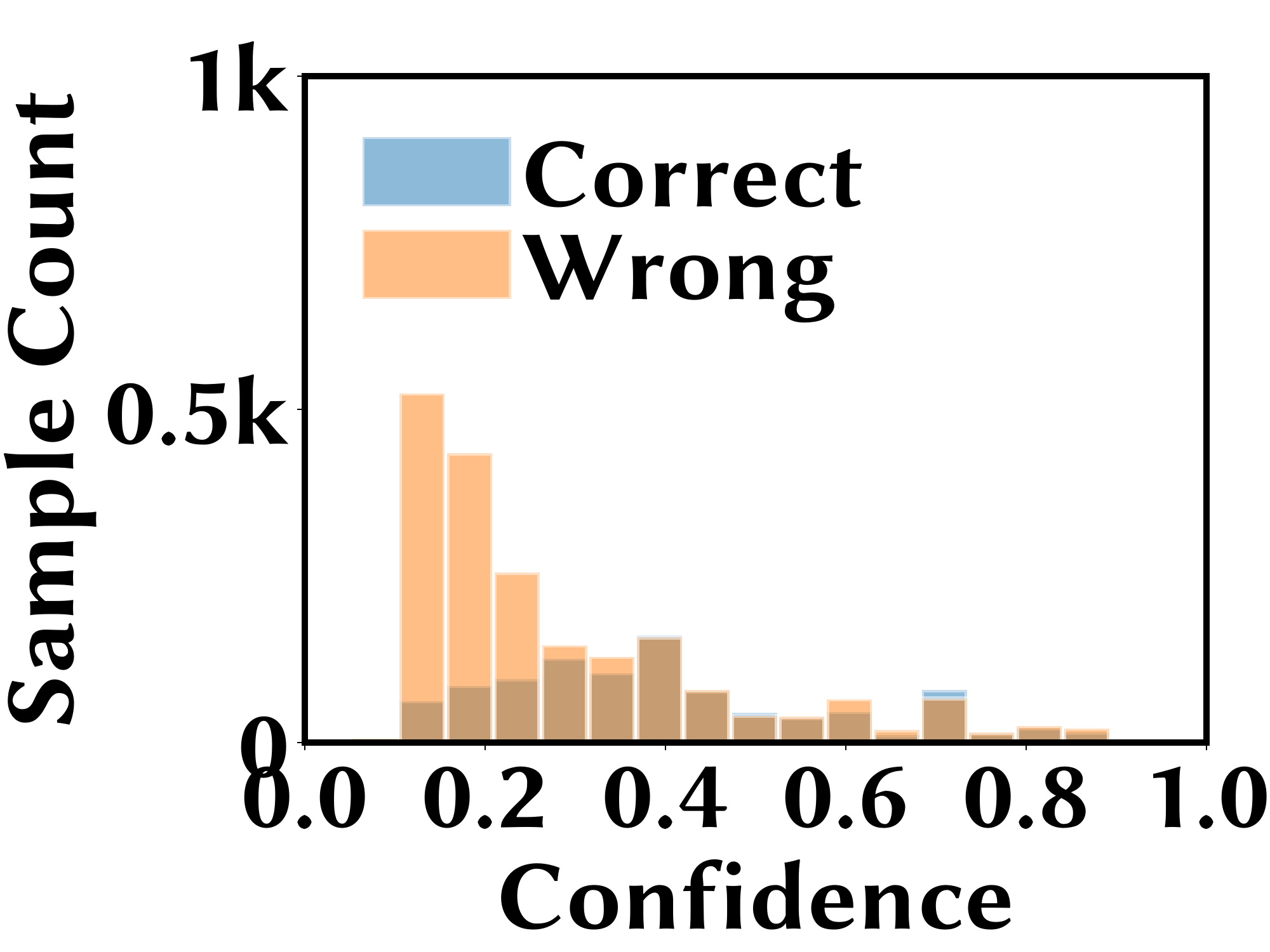}}
	\subfigure[]{\includegraphics[trim=0cm 0cm 0cm 0cm,clip,  width=0.195\textwidth]{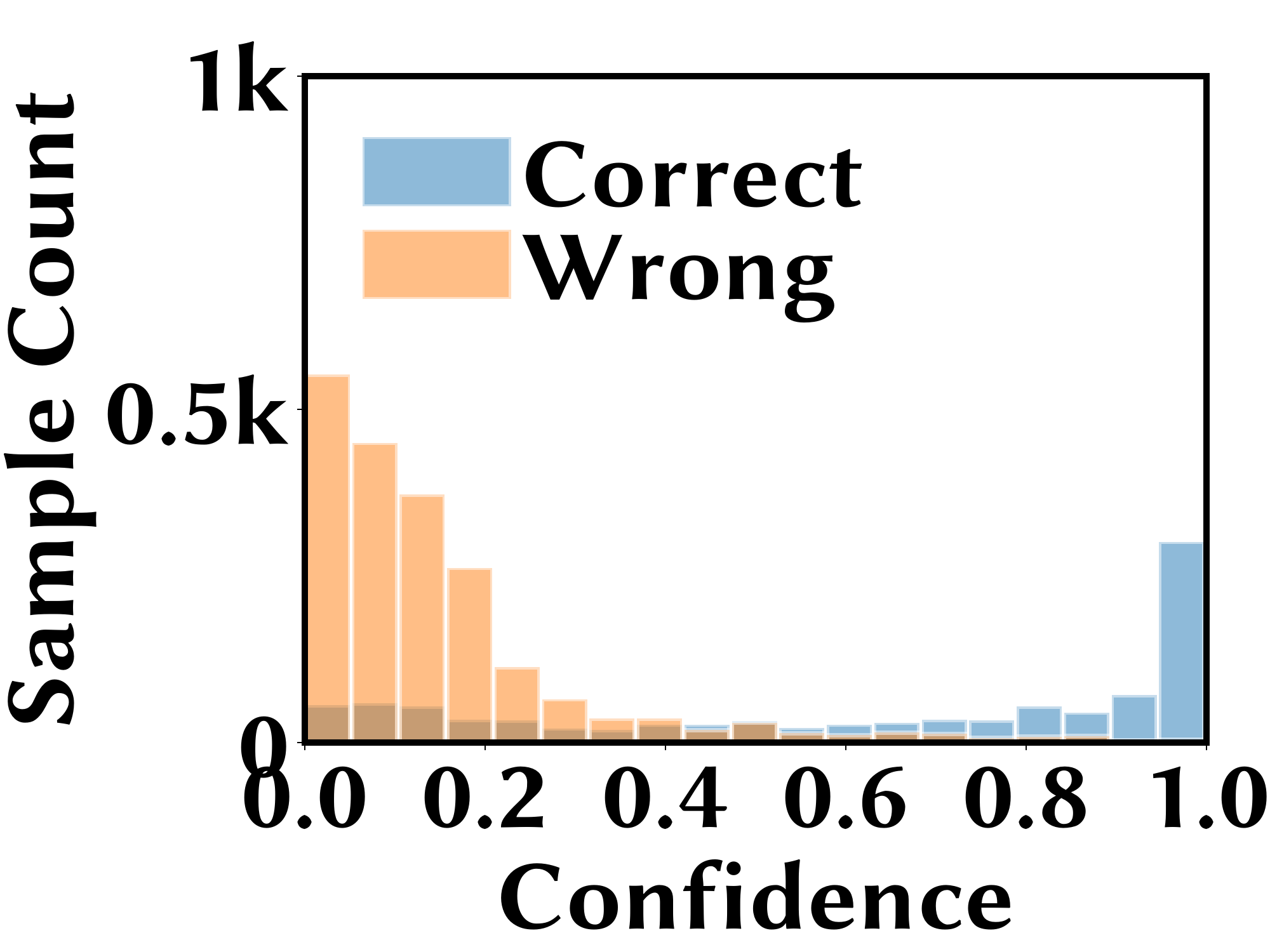}}
	\vspace{-0.3cm}
	\caption{Histograms of prediction confidences for DAN on 20 Newsgroups OOD dataset. (a) Un-calibrated. (b) Temperature scaling. (c) Scaling-binning. (d) Dirichlet calibration. (e) \ouralg.}\label{fig:ng20_OOD_histogram}
	\vspace{-0.3cm}
\end{figure*}

\begin{figure*}[!t]
	\centering
	\subfigure[]{\includegraphics[trim=0cm 0cm 0cm 0cm,clip,  width=0.195\textwidth]{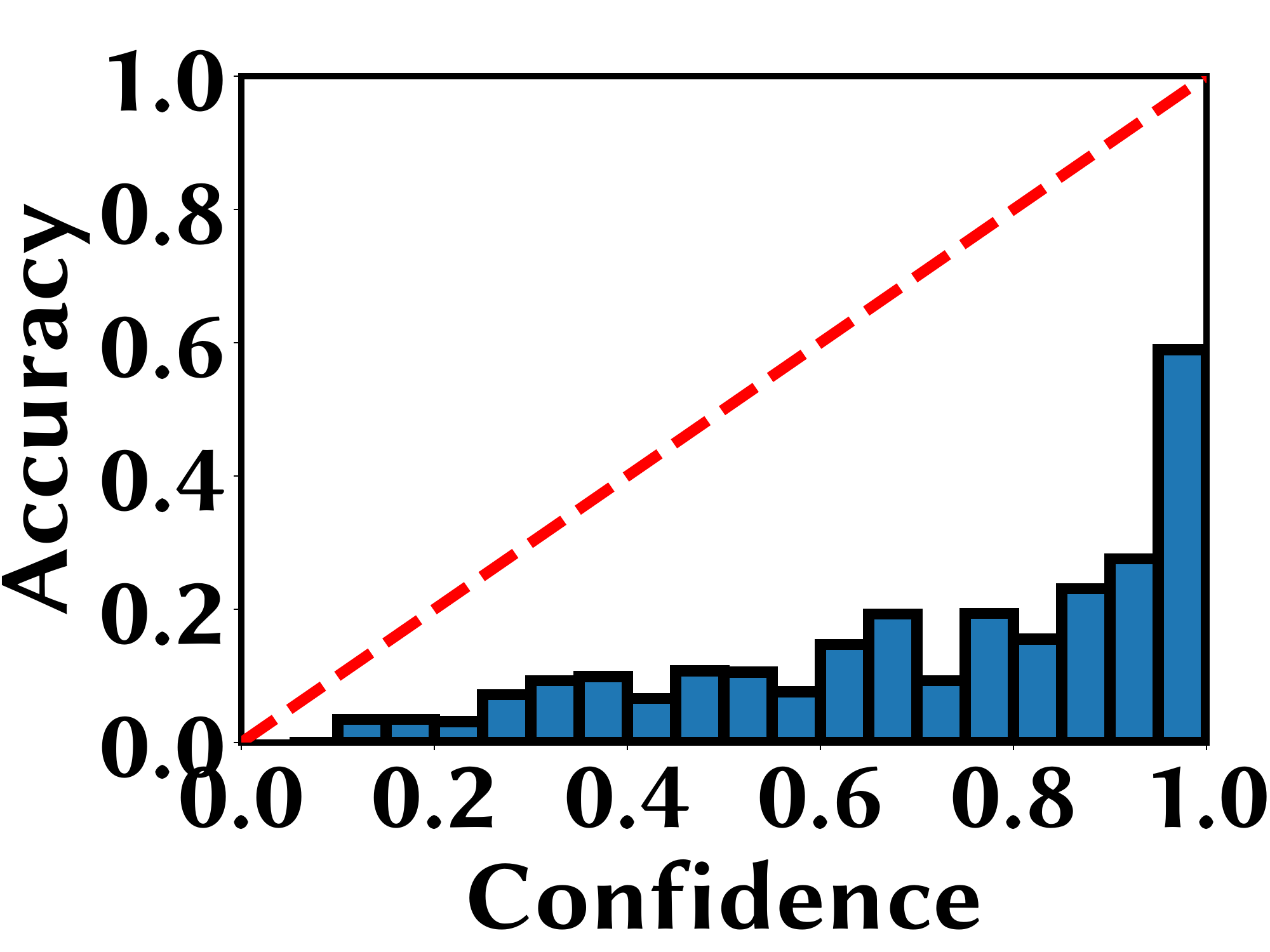}}
	\subfigure[]{\includegraphics[trim=0cm 0cm 0cm 0cm,clip,  width=0.195\textwidth]{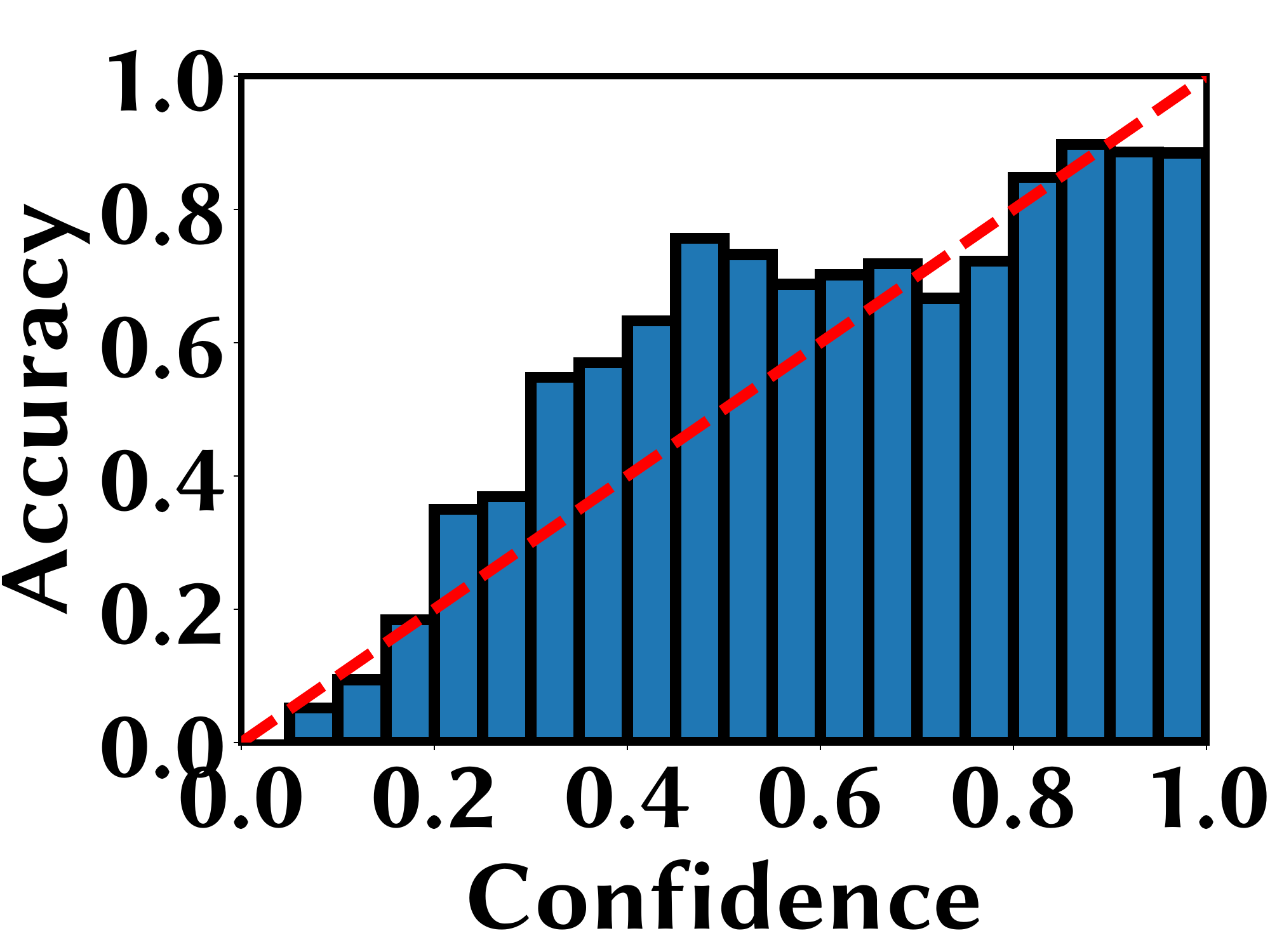}}
	\subfigure[]{\includegraphics[trim=0cm 0cm 0cm 0cm,clip,  width=0.195\textwidth]{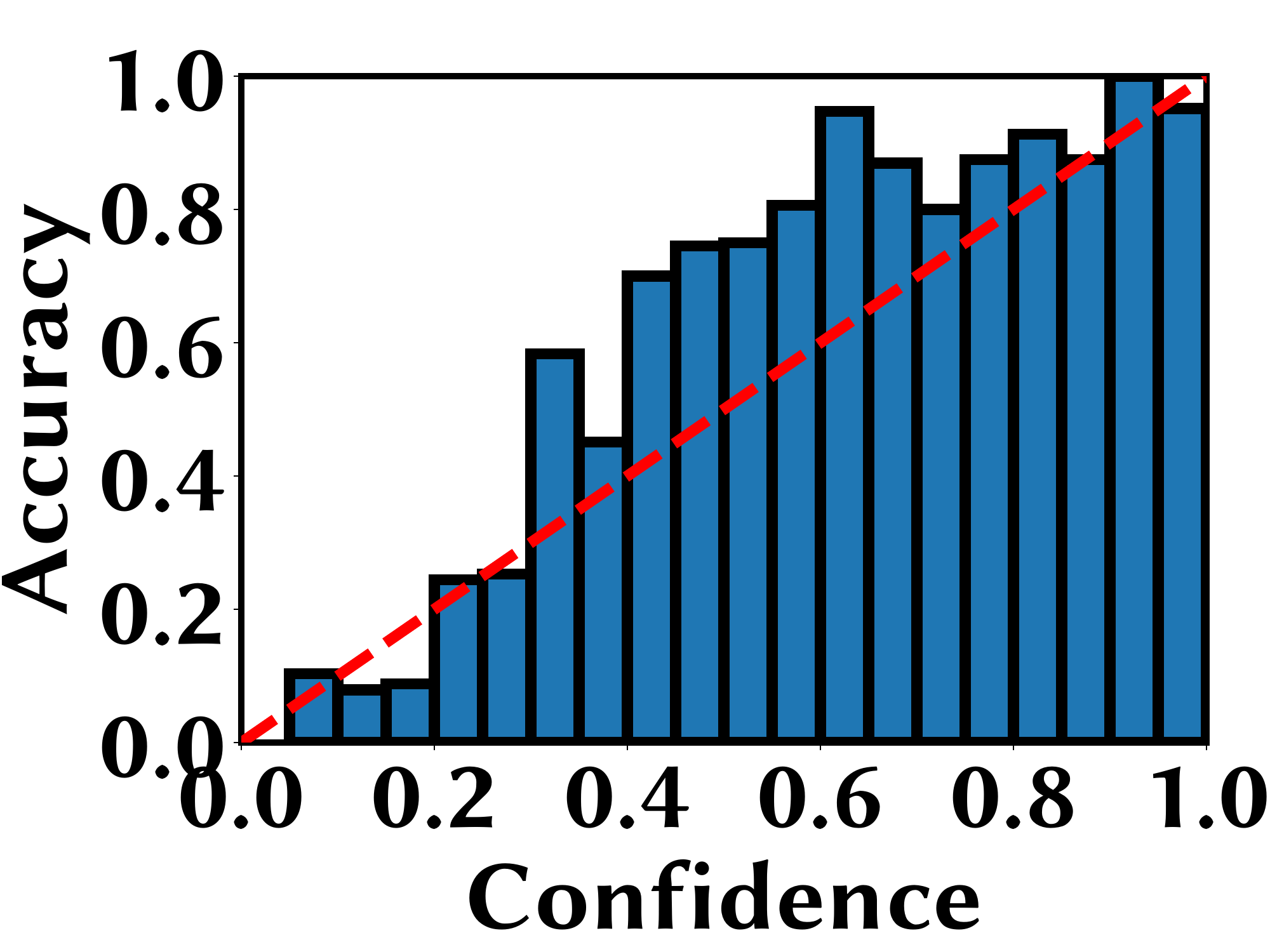}}
	\subfigure[]{\includegraphics[trim=0cm 0cm 0cm 0cm,clip,  width=0.195\textwidth]{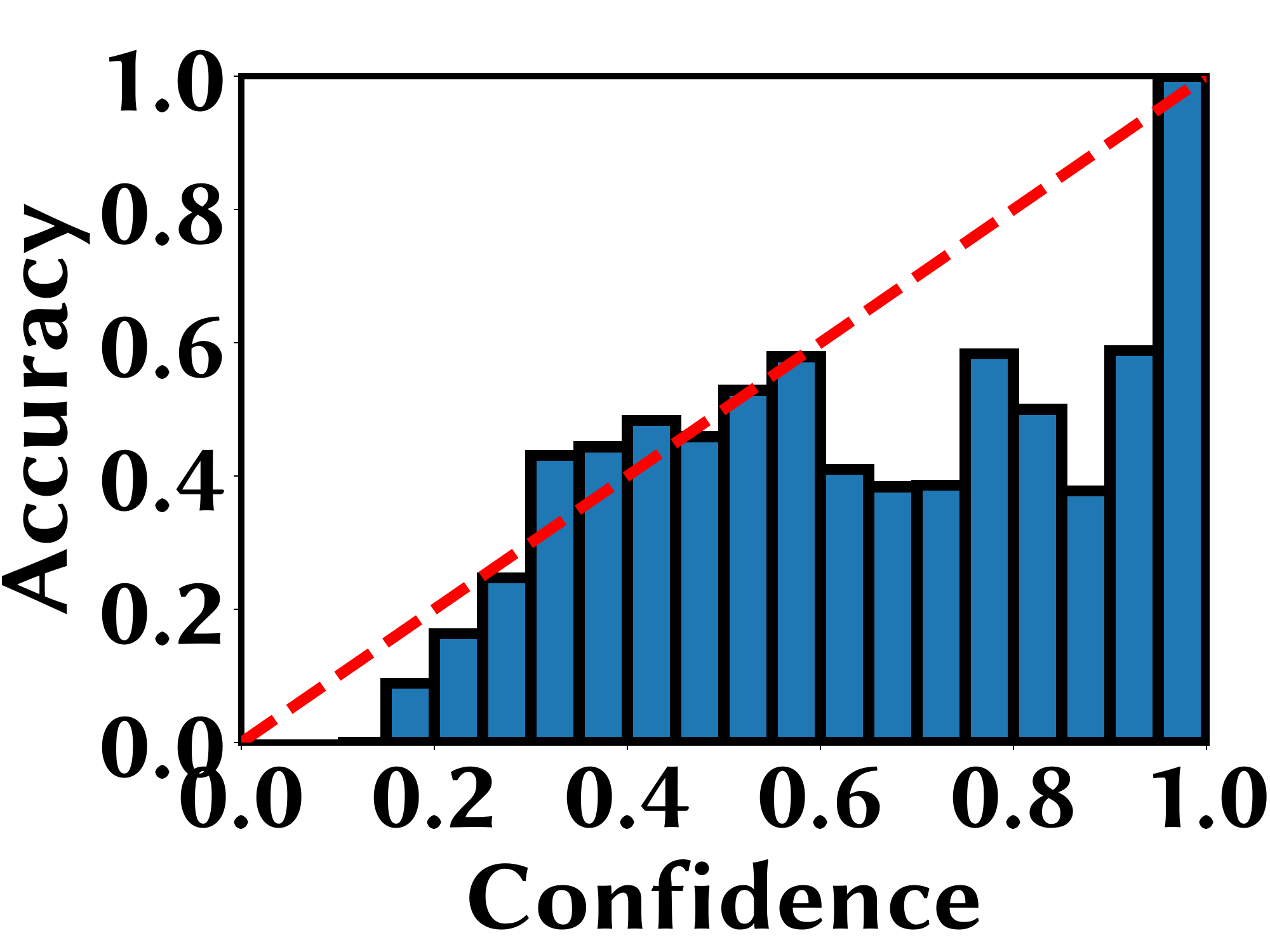}}
	\subfigure[]{\includegraphics[trim=0cm 0cm 0cm 0cm,clip,  width=0.195\textwidth]{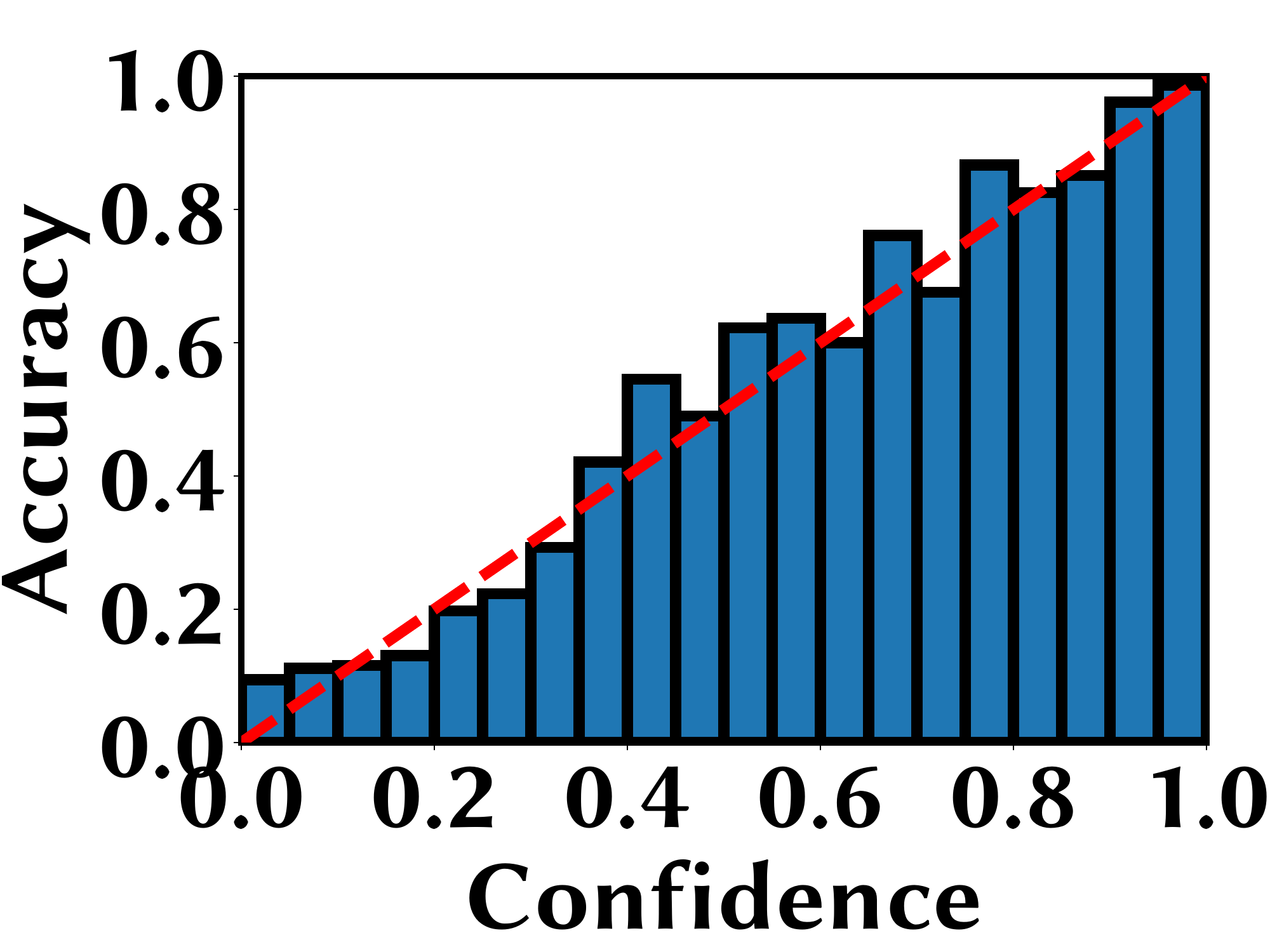}}
	\caption{Reliability diagrams of prediction confidences for DAN on 20 Newsgroups OOD dataset. (a) Un-calibrated. (b) Temperature scaling. (c) Scaling-binning. (d) Dirichlet calibration. (e) \ouralg.}\label{fig:RD_plot_ng20_OOD}
	\vspace{-0.3cm}
\end{figure*}

\textbf{Results. } The calibration results are presented in Tables~\ref{table:ng20_in_ECE_D1} and \ref{table:ng20_in_ECE_OOD} for D1 and AD datasets, respectively. The results show that the proposed methods (\basicModel, \advanceModel) outperform the baselines in terms of both mis-classification detection and confidence calibration, with more significant improvement on the OOD dataset. In addition, even though \transferMethod model is transferred to OOD dataset with fewer training samples than the baselines, it still offers a better calibration performance in terms of ECE and BS.
We further show the histograms of confidences for correct and wrong predictions in Fig.~\ref{fig:ng20_OOD_histogram} on OOD dataset. The corresponding reliability diagrams also are shown in Fig.~\ref{fig:RD_plot_ng20_OOD} on the OOD dataset with different calibration methods.

\section{Conclusion}
In this paper, we propose a new post-hoc confidence calibration method, called \method, for DNN classifiers on OOD datasets. Building on top of a lightweight neural network, \ouralg only needs the target DNN's logit layer as input and maps it to a new calibrated  confidence. The key novelty of \ouralg is  an auxiliary class in the calibration model to separate mis-classified samples from correctly classified ones, thus effectively mitigating the target DNN's being confidently wrong. We also propose a simplified version of \ouralg to reduce free parameters and facilitate transfer to a new unseen dataset. Our experiments on different DNN models, datasets and applications show that \ouralg can consistently outperform the prior post-hoc calibration methods.

\bibliographystyle{plain}


\begin{thebibliography}{10}

\bibitem{tensorflow_2016}
Mart{\'\i}n Abadi, Ashish Agarwal, Paul Barham, Eugene Brevdo, Zhifeng Chen,
  Craig Citro, Greg~S Corrado, Andy Davis, Jeffrey Dean, and Matthieu Devin.
\newblock Tensorflow: Large-scale machine learning on heterogeneous distributed
  systems.
\newblock {\em arXiv preprint arXiv:1603.04467}, 2016.

\bibitem{early_output_OOD_detection_2019}
Vahdat Abdelzad, Krzysztof Czarnecki, Rick Salay, Taylor Denounden, Sachin
  Vernekar, and Buu Phan.
\newblock Detecting out-of-distribution inputs in deep neural networks using an
  early-layer output.
\newblock {\em arXiv preprint arXiv:1910.10307}, 2019.

\bibitem{dnn_autonomous_driving_hayder_IV_2017}
Mohammed Al-Qizwini, Iman Barjasteh, Hothaifa Al-Qassab, and Hayder Radha.
\newblock Deep learning algorithm for autonomous driving using googlenet.
\newblock In {\em IVS}, 2017.

\bibitem{brier_score_books_1950}
Glenn~W Brier.
\newblock Verification of forecasts expressed in terms of probability.
\newblock {\em Monthly weather review}, 1950.

\bibitem{dnn_healthcare_caruana_2015_KDD}
Rich Caruana, Yin Lou, Johannes Gehrke, Paul Koch, Marc Sturm, and Noemie
  Elhadad.
\newblock Intelligible models for healthcare: Predicting pneumonia risk and
  hospital 30-day readmission.
\newblock In {\em KDD}, 2015.

\bibitem{deep_verifier_OOD_2019}
Tong Che, Xiaofeng Liu, Site Li, Yubin Ge, Ruixiang Zhang, Caiming Xiong, and
  Yoshua Bengio.
\newblock Deep verifier networks: Verification of deep discriminative models
  with deep generative models.
\newblock In {\em ICML}, 2020.

\bibitem{resnet_weights_github_2018}
Chenyaofo.
\newblock Pretrained models on {CIFAR}10/100 in pytorch.
\newblock \url{https://github.com/chenyaofo/CIFAR-pretrained-models}, 2018.

\bibitem{keras_2015}
Fran\c{c}ois Chollet.
\newblock Keras.
\newblock \url{https://keras.io}, 2015.

\bibitem{Learning_Boosting_with_Abstention_NIPS_2016_LearningNIPS2016_6336}
Corinna Cortes, Giulia DeSalvo, and Mehryar Mohri.
\newblock Boosting with abstention.
\newblock In {\em NIPS}. 2016.

\bibitem{Learning_with_Rejection_ALT_2016_10.1007/978-3-319-46379-7_5}
Corinna Cortes, Giulia DeSalvo, and Mehryar Mohri.
\newblock Learning with rejection.
\newblock In {\em Algorithmic Learning Theory}, 2016.

\bibitem{imagenet_lifeifei_2009_CVPR}
Jia Deng, Wei Dong, Richard Socher, Li-Jia Li, Kai Li, and Li~Fei-Fei.
\newblock Imagenet: A large-scale hierarchical image database.
\newblock In {\em CVPR}, 2009.

\bibitem{dnn_nlp_deng_2018}
Li~Deng and Yang Liu.
\newblock {\em Deep Learning in Natural Language Processing}.
\newblock Springer, 2018.

\bibitem{MCdropout_bayesian_gal_2016_ICML}
Yarin Gal and Zoubin Ghahramani.
\newblock Dropout as a bayesian approximation: Representing model uncertainty
  in deep learning.
\newblock In {\em ICML}, 2016.

\bibitem{DNN_SelectiveClassification_NIPS_2017_7073}
Yonatan Geifman and Ran El-Yaniv.
\newblock Selective classification for deep neural networks.
\newblock In {\em NIPS}. 2017.

\bibitem{vgg16_weights_github_2018}
Geifmany.
\newblock {VGG16} models for {CIFAR-10} and {CIFAR-100}.
\newblock \url{https://github.com/geifmany/cifar-vgg}, 2018.

\bibitem{DNN_Book_Goodfellow-et-al-2016}
Ian Goodfellow, Yoshua Bengio, and Aaron Courville.
\newblock {\em Deep Learning}.
\newblock MIT Press, 2016.
\newblock \url{http://www.deeplearningbook.org}.

\bibitem{explain_adversarial_goodfellow_2014}
Ian~J Goodfellow, Jonathon Shlens, and Christian Szegedy.
\newblock Explaining and harnessing adversarial examples.
\newblock 2015.

\bibitem{calibration_guo_2017_ICML}
Chuan Guo, Geoff Pleiss, Yu~Sun, and Kilian~Q Weinberger.
\newblock On calibration of modern neural networks.
\newblock In {\em ICML}, 2017.

\bibitem{dnn_speech_recognition_han_2014}
Awni Hannun, Carl Case, Jared Casper, Bryan Catanzaro, Greg Diamos, Erich
  Elsen, Ryan Prenger, Sanjeev Satheesh, Shubho Sengupta, and Adam Coates.
\newblock Deep speech: Scaling up end-to-end speech recognition.
\newblock {\em arXiv preprint arXiv:1412.5567}, 2014.

\bibitem{DNN_Uncertainty_Baseline_OOD_ICLR_2017}
Dan Hendrycks and Kevin Gimpel.
\newblock A baseline for detecting misclassified and out-of-distribution
  examples in neural networks.
\newblock In {\em ICLR}, 2017.

\bibitem{DAN_models_text_iyyer_2015}
Mohit Iyyer, Varun Manjunatha, Jordan Boyd-Graber, and Hal Daum{\'e}~III.
\newblock Deep unordered composition rivals syntactic methods for text
  classification.
\newblock In {\em AACL-IJCNLP}, 2015.

\bibitem{swa_average_wight_gordon_2018}
Pavel Izmailov, Dmitrii Podoprikhin, Timur Garipov, Dmitry Vetrov, and
  Andrew~Gordon Wilson.
\newblock Averaging weights leads to wider optima and better generalization.
\newblock In {\em UAI}, 2018.

\bibitem{DNN_Calibration_Classifier_Uncertainty_google_nips_2018}
Heinrich Jiang, Been Kim, Melody Guan, and Maya Gupta.
\newblock To trust or not to trust a classifier.
\newblock In {\em NIPS}, 2018.

\bibitem{DNN_Calibration_BayesianDNN_NIPS_2017_7141}
Alex Kendall and Yarin Gal.
\newblock What uncertainties do we need in bayesian deep learning for computer
  vision?
\newblock In {\em NIPS}. 2017.

\bibitem{cifar10_100_dataset_2009}
Alex Krizhevsky.
\newblock Learning multiple layers of features from tiny images.
\newblock 2009.

\bibitem{dirichlet_calibration_kull_2019_NIPS}
Meelis Kull, Miquel~Perello Nieto, Markus K{\"a}ngsepp, Telmo Silva~Filho, Hao
  Song, and Peter Flach.
\newblock Beyond temperature scaling: Obtaining well-calibrated multi-class
  probabilities with dirichlet calibration.
\newblock In {\em NIPS}, 2019.

\bibitem{beta_calibration_kull_2017}
Meelis Kull, Telmo~M Silva~Filho, and Peter Flach.
\newblock Beyond sigmoids: How to obtain well-calibrated probabilities from
  binary classifiers with beta calibration.
\newblock {\em Electronic Journal of Statistics}, 2017.

\bibitem{scaling_binning_calibration_kumar_2019_NIPS}
Ananya Kumar, Percy~S Liang, and Tengyu Ma.
\newblock Verified uncertainty calibration.
\newblock In {\em NIPS}, 2019.

\bibitem{deep_ensembles_charles_2017_NIPS}
Balaji Lakshminarayanan, Alexander Pritzel, and Charles Blundell.
\newblock Simple and scalable predictive uncertainty estimation using deep
  ensembles.
\newblock In {\em NIPS}, 2017.

\bibitem{Verification_SimpleFramework_OOD_Generative_Gaussian_NIPS_2018_7947}
Kimin Lee, Kibok Lee, Honglak Lee, and Jinwoo Shin.
\newblock A simple unified framework for detecting out-of-distribution samples
  and adversarial attacks.
\newblock In {\em NIPS}, 2018.

\bibitem{reuters_lewis_1987}
David Lewis.
\newblock Reuters-21578 text categorization collection data set.
\newblock
  \url{https://archive.ics.uci.edu/ml/datasets/Reuters-21578+Text+Categorization+Collection}.

\bibitem{Verification_OOD_Detection_ODIN_Srikant_UIUC_ICLR_2018}
Shiyu Liang, Yixuan Li, and Rayadurgam Srikant.
\newblock Enhancing the reliability of out-of-distribution image detection in
  neural networks.
\newblock In {\em ICLR}, 2018.

\bibitem{nlp_augmentation_ma_2019}
Edward Ma.
\newblock {NLP} augmentation.
\newblock \url{https://github.com/makcedward/nlpaug}, 2019.

\bibitem{tSNE_virtualization_geoffrey_2008}
Laurens van~der Maaten and Geoffrey Hinton.
\newblock Visualizing data using t-sne.
\newblock {\em Journal of Machine Learning Research}, 2008.

\bibitem{DNN_Uncertainty_PriorNetworks_NIPS_2018_malinin2018predictive}
Andrey Malinin and Mark Gales.
\newblock Predictive uncertainty estimation via prior networks.
\newblock In {\em NIPS}, 2018.

\bibitem{Verification_OOD_SelfSupervised_SeparateClass_TAMU_ZhengyangWang_AAAI_2020}
Sina Mohseni, Mandar Pitale, JBS Yadawa, and Zhangyang Wang.
\newblock A baseline for detecting misclassified and out-of-distribution
  examples in neural networks.
\newblock In {\em AAAI}, 2020.

\bibitem{bayesian_binning_milos_AAAI_2015}
Mahdi~Pakdaman Naeini, Gregory Cooper, and Milos Hauskrecht.
\newblock Obtaining well calibrated probabilities using bayesian binning.
\newblock In {\em AAAI}, 2015.

\bibitem{scikit-learn_2011}
F.~Pedregosa, G.~Varoquaux, A.~Gramfort, V.~Michel, B.~Thirion, O.~Grisel,
  M.~Blondel, P.~Prettenhofer, R.~Weiss, V.~Dubourg, J.~Vanderplas, A.~Passos,
  D.~Cournapeau, M.~Brucher, M.~Perrot, and E.~Duchesnay.
\newblock Scikit-learn: Machine learning in {P}ython.
\newblock {\em Journal of Machine Learning Research}, 2011.

\bibitem{glove_embedding_pennington_EMNLP_2014}
Jeffrey Pennington, Richard Socher, and Christopher~D Manning.
\newblock Glove: Global vectors for word representation.
\newblock In {\em EMNLP}, 2014.

\bibitem{platter_scaling_1999}
JC~Platt.
\newblock Probabilities for sv machines, advances in large margin classifiers,
  1999.

\bibitem{adversarial_toolbox_foolbox_metthias_ICML_2017}
Jonas Rauber, Wieland Brendel, and Matthias Bethge.
\newblock Foolbox: A python toolbox to benchmark the robustness of machine
  learning models.
\newblock In {\em ICML}, 2017.

\bibitem{rennie_20_Newsgroups_2008}
Jason Rennie and Ken Lang.
\newblock 20 {N}ews{g}roups {D}ataset.
\newblock \url{http://qwone.com/~jason/20Newsgroups}.

\bibitem{imagenet_dataset_2012_IJCV}
Olga Russakovsky, Jia Deng, Hao Su, Jonathan Krause, Sanjeev Satheesh, Sean Ma,
  Zhiheng Huang, Andrej Karpathy, Aditya Khosla, Michael Bernstein,
  Alexander~C. Berg, and Li~Fei-Fei.
\newblock {ImageNet Large Scale Visual Recognition Challenge}.
\newblock {\em International Journal of Computer Vision}, 2015.

\bibitem{dnn_medical_diagnose_MRI_sander_2019}
J{\"o}rg Sander, Bob~D de~Vos, Jelmer~M Wolterink, and Ivana I{\v{s}}gum.
\newblock Towards increased trustworthiness of deep learning segmentation
  methods on cardiac mri.
\newblock In {\em Medical Imaging 2019: Image Processing}, 2019.

\bibitem{model_uncertain_suchi_ICAIS_2019}
Peter Schulam and Suchi Saria.
\newblock Can you trust this prediction? auditing pointwise reliability after
  learning.
\newblock In {\em AISTATS}, 2019.

\bibitem{DNN_Uncertainty_Estimation_Summary_NIPS_2019_ovadia}
Jasper Snoek, Yaniv Ovadia, Emily Fertig, Ren, et~al.
\newblock Can you trust your model's uncertainty? {Evaluating} predictive
  uncertainty under dataset shift.
\newblock In {\em NeurIPS}, 2019.

\bibitem{bining_calibration_charles_ICML_2001}
Bianca Zadrozny and Charles Elkan.
\newblock Obtaining calibrated probability estimates from decision trees and
  naive bayesian classifiers.
\newblock In {\em ICML}, 2001.

\bibitem{isotonic_regression_charles_2012_KDD}
Bianca Zadrozny and Charles Elkan.
\newblock Transforming classifier scores into accurate multiclass probability
  estimates.
\newblock In {\em KDD}, 2002.

\end{thebibliography}

\end{document}